\documentclass[letterpaper]{article} % DO NOT CHANGE THIS
\usepackage{aaai2026}  % DO NOT CHANGE THIS
\usepackage{times}  % DO NOT CHANGE THIS
\usepackage{helvet}  % DO NOT CHANGE THIS
\usepackage{courier}  % DO NOT CHANGE THIS
\usepackage[hyphens]{url}  % DO NOT CHANGE THIS
\usepackage{graphicx} % DO NOT CHANGE THIS
\usepackage{natbib}  % DO NOT CHANGE THIS AND DO NOT ADD ANY OPTIONS TO IT
\usepackage{caption} % DO NOT CHANGE THIS AND DO NOT ADD ANY OPTIONS TO IT
\usepackage{algorithm}
\usepackage{algorithmic}
\usepackage{amsmath}
\usepackage{amssymb}
\usepackage{mathtools}
\usepackage{amsthm}
\usepackage{graphicx}
\usepackage{caption}
\usepackage{subcaption}
\usepackage{tabularx}
\usepackage{array}
\usepackage[most]{tcolorbox}
\usepackage{multirow}
\usepackage{algorithm}
\usepackage{algorithmic}
\usepackage{xcolor}         % colors
\usepackage{xspace}
\usepackage{enumitem}
\usepackage{booktabs}
\usepackage{pifont}
\usepackage[capitalize,noabbrev]{cleveref}

\newcommand{\blue}[1]{\textcolor{black}{#1}}

\newcommand{\eg}{\textit{e.g.}\xspace}
\newcommand{\ie}{\textit{i.e.}\xspace}

\DeclareMathOperator*{\argmax}{arg\,max}

\usepackage[capitalize,noabbrev]{cleveref}

\newcommand{\cB}{\mathcal{B}}

\theoremstyle{plain}

\theoremstyle{definition}

\theoremstyle{remark}

\usepackage{newfloat}
\usepackage{listings}
\DeclareCaptionStyle{ruled}{labelfont=normalfont,labelsep=colon,strut=off} % DO NOT CHANGE THIS
\floatstyle{ruled}
\newfloat{listing}{tb}{lst}{}
\floatname{listing}{Listing}
\title{An Information Theoretic Evaluation Metric for Strong Unlearning}

\author {
    Dongjae Jeon\textsuperscript{\rm 1}\equalcontrib,
    Wonje Jeung\textsuperscript{\rm 2}\equalcontrib,
    Taeheon Kim\textsuperscript{\rm 3},
    Albert No\textsuperscript{\rm 2}\thanks{Corresponding authors.},
    Jonghyun Choi\textsuperscript{\rm 3,4,5}\footnotemark[2]
}
\affiliations {
    \textsuperscript{\rm 1}Department of Computer Science, Yonsei University\\
    \textsuperscript{\rm 2}Department of Artificial Intelligence, Yonsei University\\
    \textsuperscript{\rm 3}ECE, \textsuperscript{\rm 4}IPAI and \textsuperscript{\rm 5}ASRI, Seoul National University\\
    \vskip 0.5em
    \{dongjae0324, specific0924, albertno\}@yonsei.ac.kr, 
    \{thkim0305, jonghyunchoi\}@snu.ac.kr
}

\begin{document}

\maketitle

\begin{abstract}
Machine unlearning (MU) aims to remove the influence of specific data from trained models, addressing privacy concerns and ensuring compliance with regulations such as the ``right to be forgotten.'' Evaluating strong unlearning, where the unlearned model is indistinguishable from one retrained without the forgetting data, remains a significant challenge in deep neural networks (DNNs). Common black-box metrics, such as variants of membership inference attacks and accuracy comparisons, primarily assess model outputs but often fail to capture residual information in intermediate layers. To bridge this gap, we introduce the Information Difference Index (IDI), a novel white-box metric inspired by information theory.
IDI quantifies retained information in intermediate features by measuring mutual information between those features and the labels to be forgotten, offering a more comprehensive assessment of unlearning efficacy. Our experiments demonstrate that IDI effectively measures the degree of unlearning across various datasets and architectures, providing a reliable tool for evaluating strong unlearning in DNNs.
\end{abstract}

%%%%%%%%%%%%%%%%%%%%%%%%%%%%%%%%%%%%%%%%%%%%%%%%%%%%%%%%%%%%
%%% Introduction
%%%%%%%%%%%%%%%%%%%%%%%%%%%%%%%%%%%%%%%%%%%%%%%%%%%%%%%%%%%%
\section{Introduction}
Machine unlearning (MU) seeks to remove the impact of specific data samples from a trained model, addressing privacy issues such as ``right to be forgotten''~\citep{voigt2017eu}.
In addition to privacy, MU is also emerging as a tool to eliminate the influence of corrupted or outdated data used during training~\citep{nguyen2022survey, kurmanji2024scrub}. 
The most straightforward approach to MU is \textit{exact unlearning}, where the model is retrained from scratch, excluding the data that need to be forgotten.
Although this method ensures complete data removal, it is computationally expensive and not scalable~\citep{aldaghri2021mu_exact_3, bourtoule2021mu_exact_1}.
Consequently, research has shifted towards \textit{approximate unlearning}, which aims to replicate the effects of retraining in a more efficient manner.

The goal of MU is to create an unlearned model that is indistinguishable from a model retrained from scratch, referred to as strong unlearning.
This objective has become particularly crucial with the rise of open-source models like Stable Diffusion~\citep{rombach_stable_opensource} and LLaMA~\citep{touvron2023llama_opensource}, which are widely used and fine-tuned by various users.
For unlearning algorithms to be practically useful, they must be capable of fully eliminating traces of private data and preventing potential exploitation.
While ($\epsilon$, $\delta$)-certified unlearning methods~\citep{zhang2024dnn_certified, mu2024dnn_certified2} provide theoretical guarantees, they are impractical for large-scale models. 
As a result, most approximate unlearning methods rely on heuristic approaches, lacking formal guarantees.
Thus, these methods must undergo empirical evaluation to demonstrate their effectiveness. 

However, current evaluations, primarily based on black-box approaches such as membership inference attacks (MIA)~\citep{shokri2017mia1} and accuracy comparisons, focus on output similarity rather than internal model changes. 
Although these metrics may capture weak unlearning~\citep{fan2023salun, liu2024l1sparse, chundawat2023badt}, they may not be sufficient for assessing strong unlearning. 
In this work, we investigate whether relying solely on outputs can truly reflect complete influence removal, considering that model outputs can be superficially adjusted~\citep{kirichenko2022head2}.

Surprisingly, our experiments reveal that even minimal changes to the model, such as modifying only the final layer while preserving all information in the intermediate layers, can still satisfy the black-box evaluation metrics, exposing their limitations in assessing strong unlearning.
This finding also raises critical concerns about whether current MU methods genuinely achieve information removal comparable to retraining from scratch.

Consequently, motivated by the Information Bottleneck principle~\citep{tishby2000ib_principle, tishby201ib_dpi1},
we introduce the \textbf{information difference index (IDI)}, a novel white-box metric designed to quantify residual information in intermediate layers after unlearning.
IDI measures the mutual information~\citep{shannon_mutual_information} between intermediate features and the forgetting labels, providing an interpretable value to assess the effectiveness of unlearning algorithms.
We observe that IDI remains stable under the stochasticity of the unlearning process and is compatible with diverse model architectures. 
Moreover, IDI can be reliably estimated from a data subset, making it practical for large-scale unlearning settings.

Through IDI, we observe that many recent MU methods, despite strong performance on black-box metrics, retain substantial information about the forgetting data in intermediate layers. 
To address this, we propose \textbf{COLapse-and-Align (COLA)}, a simple method that collapses representations associated with the forget set at the feature level to eliminate residual information, followed by re-aligning the retained features. 
Despite its simplicity, COLA consistently improves IDI scores while maintaining competitive performance across datasets such as CIFAR-10/100, and ImageNet-1K, and architectures (ResNet-18/50, ViT).

%%% Problem Statement and Preliminaries
%%%%%%%%%%%%%%%%%%%%%%%%%%%%%%%%%%%%%%%%%%%%%%%%%%%%%%%%%%%%
\section{Problem Statement and Preliminaries}

\subsection{Problem Statement}
Let \( D = \{(x_i, y_i)\}_{i=1}^N \) denote a training dataset comprising \( N \) image-label pairs \((x_i, y_i)\).
In a supervised learning setup, \( D \) is partitioned into two subsets: the {\it forget set} \( D_f \), containing the data points to be removed,
and the {\it retain set} \( D_r = D \setminus D_f \), containing the data points to be preserved.
The initial model \( \theta_o \), referred to as the \textbf{Original model}, is trained on the full dataset \( D \) using empirical risk minimization.
The \textbf{Retrain model} \( \theta_r \) is trained from scratch on only the retain set \( D_r \).
% [SEE} MU
The \textbf{unlearned model} \( \theta_u \) is obtained by applying a MU algorithm to the Original model \( \theta_o \), aiming to remove the influence of \( D_f \).
The goal of MU is for \( \theta_u \) to closely approximate \( \theta_r \), ensuring the unlearned model behaves as though \( D_f \) had never been used in training,
while preserving the training methodology across \( \theta_o \), \( \theta_r \), and \( \theta_u \).

Throughout the paper, within a given model \( \mathbf{\theta}\), we define the \textbf{head} as the last few layers responsible for classification; typically one to three linear layers. 
The \textbf{encoder}, on the contrary, encompasses the remainder of the network, which usually consists of convolutional layers or transformer encoders.
MU is often studied in the context of image classification~\citep{shaik2023mu_classification, nguyen2022survey}, where it is typically classified into two scenarios based on the nature of the forget set: \textbf{class-wise forgetting}, where all samples from a specific class are targeted, and \textbf{random data forgetting}, where samples are selected indiscriminately across all classes.
In this work, we tackle both scenarios.

\subsection{Preliminaries}
\paragraph{Machine Unlearning (MU).}

Exact unlearning, which involves creating Retrain, guarantees the information removal from the forget set but is computationally expensive~\citep{bourtoule2021mu_exact_1, yan2022mu_exact_2}.
To address this, approximate unlearning methods have been developed, focusing on efficiency rather than strict theoretical guarantees. 
Specifically, strong unlearning, where the unlearned model is indistinguishable from Retrain, has been explored through the application of differential privacy (DP)~\citep{journals_differential_privacy} inspired techniques, which aim to achieve parameter-level indistinguishability~\citep{journals_differential_privacy, neel2021dp_2, sekhari2021dp_3}.
However, applying such techniques to neural networks remains challenging due to their vast number of parameters and non-convex loss landscapes~\citep{qiao2024certified_hard}. 
As a result, recent studies typically assess the similarity of model outputs (i.e., predictions), using weak unlearning as a practical proxy for strong unlearning~\citep{Xu2023MachineSurveyStrong}.

Although empirically ensuring strong unlearning is challenging, it remains essential for deploying unlearning algorithms in compliance with legal requirements such as the GDPR~\citep{voigt2017eu} and the ``right to be forgotten.'' 
This need is further amplified by the widespread use of open-source models like CLIP~\citep{radford2021inconce_mm1_clip}, Stable Diffusion~\citep{rombach_stable_opensource}, and LLaMA~\citep{touvron2023llama_opensource}, where sensitive data may inadvertently persist and be exploited. 
Our work focuses on developing a robust empirical metric to evaluate unlearning algorithms, distinct from verification~\citep{zhang2024verifyone, sommer2020verifytwo}, which assesses effectiveness through real-world attack scenarios.

\paragraph{Evaluation Criteria in MU.}
As the goal of MU is to remove the influence of specific data while preserving the others, the unlearning algorithms are typically evaluated on three criteria: \textit{Efficacy}, \textit{Accuracy}, and \textit{Efficiency}~\citep{hayes2024inexact}.
Efficacy measures how closely the unlearned model approximates Retrain, which is key to unlearning quality. 
Accuracy ensures task performance remains intact after unlearning, while efficiency ensures the unlearning process is faster than retraining.

Accuracy and efficiency can be easily evaluated using existing metrics.
Accuracy consists of three categories: \textit{unlearning accuracy (UA)}, \textit{remaining accuracy (RA)}, and \textit{testing accuracy (TA)}.
UA measures performance on \(\mathcal{D}_f\) as \( UA(\mathbf{\theta_u}) = 1 - \text{Acc}_{\mathcal{D}_f}(\mathbf{\theta_u}) \), RA on \(\mathcal{D}_r\) as \( RA(\mathbf{\theta_u}) = \text{Acc}_{\mathcal{D}_r}(\mathbf{\theta_u}) \), and TA measures generalization to unseen data as \( TA(\mathbf{\theta_u}) = \text{Acc}_{\mathcal{D}_{test}}(\mathbf{\theta_u}) \).
Performance levels comparable to Retrain across these metrics indicate better unlearning.

In terms of efficiency, \textit{runtime efficiency (RTE)} measures the time an algorithm takes to complete unlearning, with lower RTE indicating more efficient unlearning~\citep{fan2023salun, liu2024l1sparse}.
However, assessing unlearning efficacy, or determining whether the unlearned model has fully removed the influence of specific data to the same extent as Retrain, remains a significant challenge in complex DNNs.
The efficacy metrics are divided into two categories: \textit{black-box} metrics, which focus solely on model outputs (i.e., predictions),
and \textit{white-box} metrics, which examine internal dynamics such as parameters, gradients, and features. 
While black-box metrics are typically used due to their convenience, no universally accepted standard exists, leaving room for more reliable assessment.

\begin{figure*}[!t]
    \centering
    \includegraphics[width=0.8\textwidth]{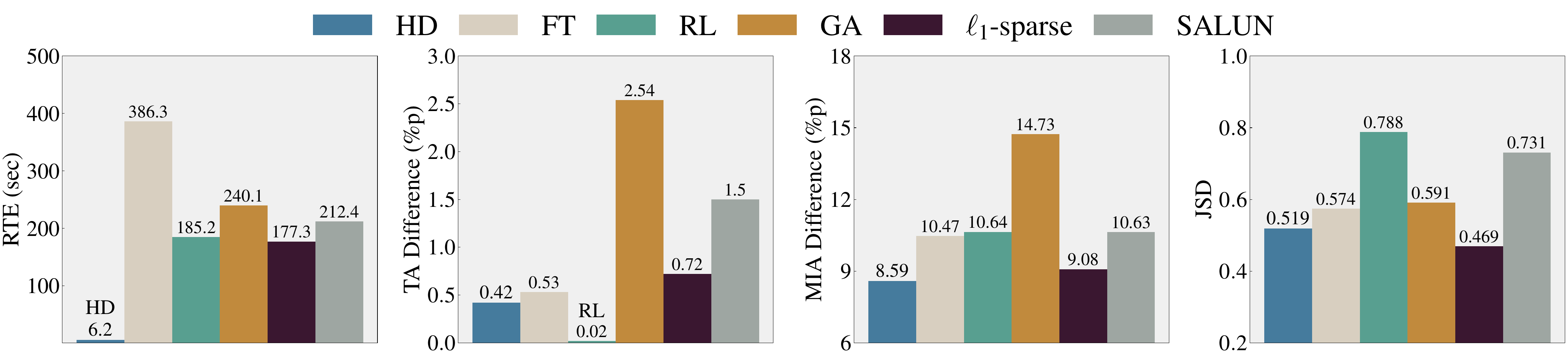}
    \centering
    \caption{
   Performance of six methods on (CIFAR-10, ResNet-18), evaluated in efficiency (RTE), accuracy (TA), and efficacy (MIA, JSD). 
   For TA, MIA, and JSD, lower differences from Retrain are preferred, indicating closer similarity to Retrain.
   }
    % \vspace{-1em}
    \label{fig:exp__hd_rte_ta}
\end{figure*}

\paragraph{Black-box Efficacy Metrics.}
Variants of membership inference attacks (MIA)~\citep{shokri2017mia1} are among the most widely used black-box metrics for evaluating unlearning~\citep{fan2023salun, liu2024l1sparse, foster2024ssd}. 
MIA trains an auxiliary classifier to determine whether a given sample was part of the training set, with attack success rates on the forget set close to those for Retrain being preferred in unlearning. 
Recent studies often combine MIA with UA, RA, TA, and RTE to assess unlearning across efficacy, accuracy, and efficiency~\citep{chen2023boundary, kim2024lau}, a practice referred to as the `full-stack' evaluation~\citep{liu2024l1sparse, fan2023salun}.

Other metrics, such as Jensen-Shannon divergence (JSD), and ZRF~\citep{chundawat2023badt, poppi2024multi} compare the output logits between the unlearned model and Retrain (or a random model for ZRF).
Additionally, time-based metrics like Anamnesis Index (AIN)~\citep{chundawat2023ain, tarun2023deep} and relearn time (RT)~\citep{tarun2023unsir} track how long the model takes to regain performance on the forget set.
While convenient, black-box metrics overlook internal behaviors and cannot verify strong unlearning by ensuring forgetting data's influence is fully removed. 
Their limitations are further discussed in~\cref{section_residualinformation}.

\paragraph{White-box Efficacy Metrics.}
In contrast, white-box metrics offer deeper insights by analyzing internal model dynamics. 
Previous studies have measured parameter-wise distances (e.g., \(\mathbf{\ell_2}\)-distance, KL-divergence) between the unlearned model and Retrain~\citep{golatkar2020eternal, wu2020param_dist1}.
However, this approach is computationally expensive and unreliable due to training randomness~\citep{hayes2024inexact, goel2022evaluating}.
\citet{becker2022evaluating} proposed a Fisher information based metric, but their results were inconsistent with theoretical intuition.
\citet{graves2021amnesiac} applied model inversion attacks to reconstruct images from the forget set, but their method relies on visual comparisons.

Although robust white-box metrics are currently lacking and challenging to develop, they are crucial for validating approximate methods that lack formal guarantees.
Without them, these algorithms cannot be trusted in privacy-sensitive applications that demand a high level of confidence in information removal.
To address this critical need, our work proposes a reliable and practical white-box metric.

\begin{figure*}[!t]
    \hspace{2.5em}
    \begin{subfigure}[b]{0.15\textwidth}
        \centering
        \resizebox{\linewidth}{!}{
            \includegraphics{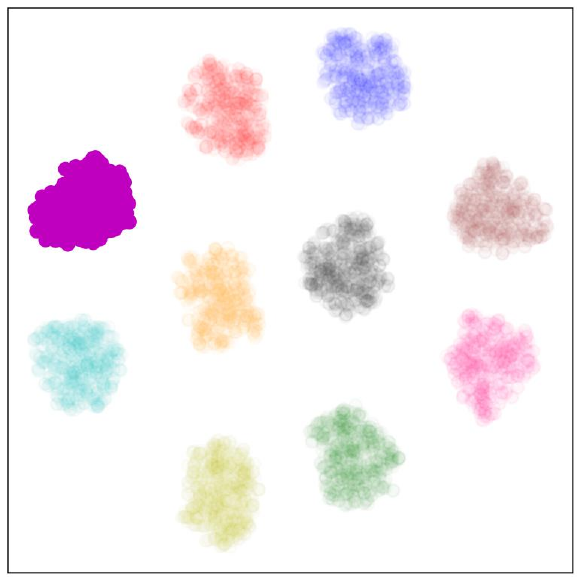}
        }
        % \vspace{-1.5em}
        \caption{Original / HD}
        \label{subfig:representation_tsne_a}
    \end{subfigure}
    \begin{subfigure}[b]{0.15\textwidth}
        \centering
        \resizebox{\linewidth}{!}{
            \includegraphics{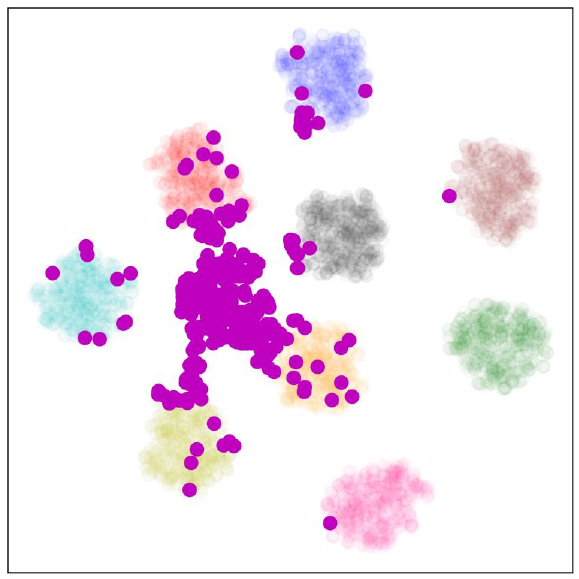}
        }
        % \vspace{-1.5em}
        \caption{Retrain}
        \label{subfig:representation_tsne_b}
    \end{subfigure}
    \hfill
    \begin{subfigure}[b]{0.15\textwidth}
        \centering
        \resizebox{\linewidth}{!}{
            \includegraphics{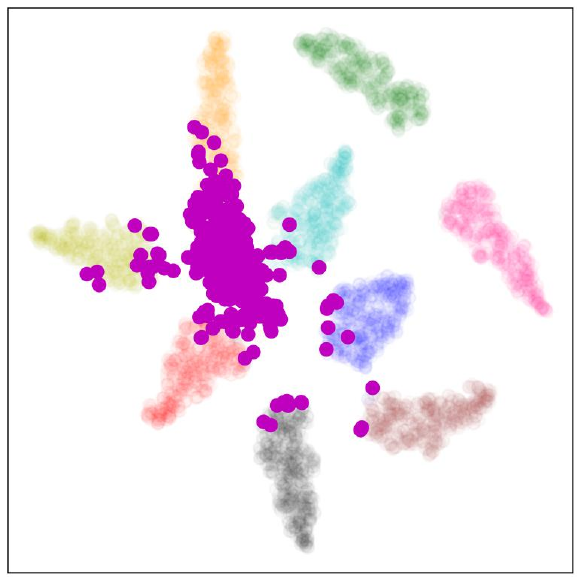}
        }
        
        % \vspace{-0.4em}
        \caption{GA}
        \label{subfig:representation_tsne_c}
    \end{subfigure}
    \begin{subfigure}[b]{0.15\textwidth}
        \centering
        \resizebox{\linewidth}{!}{
            \includegraphics{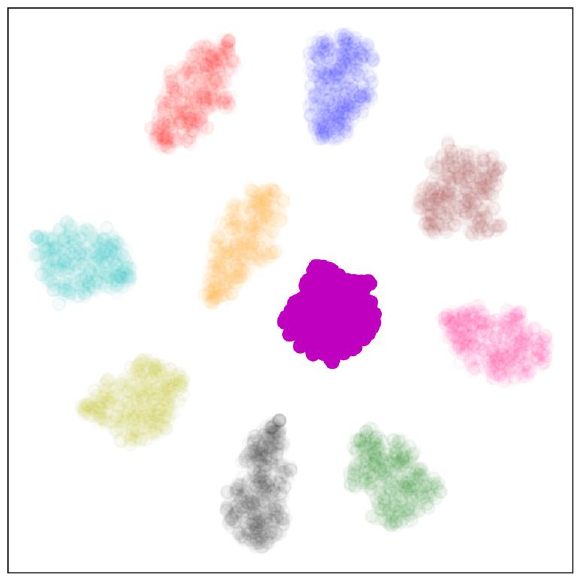}
        }
        % \vspace{-1.5em}
        \caption{RL}
        \label{subfig:representation_tsne_d}
    \end{subfigure} 
    \begin{subfigure}[b]{0.15\textwidth}
        \centering
        \resizebox{\linewidth}{!}{
            \includegraphics{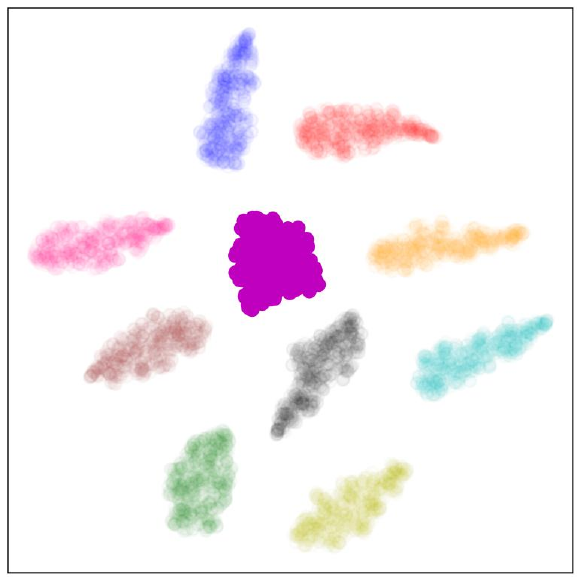}
        }
        % \vspace{-1.5em}
        \caption{SALUN}
        \label{subfig:representation_tsne_e}
    \end{subfigure}
    \hspace{2.5em}
    \caption{
    t-SNE visualizations of encoder outputs for Original, Retrain, and unlearned models from three MU methods (GA, RL, SALUN) on single-class forgetting with (CIFAR-10, ResNet-18). In each t-SNE plot, features of the forgetting class are represented in purple. 
    Original and HD have identical feature distribution as they share the same encoder.
   }
    \label{fig:representation_tsne} 
    % \vspace{-1em}
\end{figure*}
\section{Rethinking the Evaluation of Unlearning}
\label{section_residualinformation}

%%%%%%%%%%%%%%%%%%%%%%%%%%%%%%%%%%%%%%%%%%%%%%%%%%%%%%%%%%%%
%%% Challenging Black-box Metrics
%%%%%%%%%%%%%%%%%%%%%%%%%%%%%%%%%%%%%%%%%%%%%%%%%%%%%%%%%%%%
\subsection{Challenging Black-box Metrics}
\label{subsection_HD}

In this section, we reveal the limitations of commonly used black-box efficacy metrics by applying a simple unlearning technique to a single-class forgetting task. 
We show that these metrics can misrepresent unlearning efficacy, even when the model’s output closely resembles that of Retrain.

Inspired by the teacher-student framework, our strategy, termed \textbf{head distillation (HD)}, employs logit distillation from Original \(\mathbf{\theta_o}\).
The unlearned model \(\mathbf{\theta_u}\) is initialized from \(\mathbf{\theta_o}\) with the encoder frozen and only the head trainable.
During unlearning, the head is finetuned on training dataset \(\mathcal{D}\) using KL-divergence loss~\citep{hinton2015kldivergence} to match a masked version of \(\mathbf{\theta_o}\)'s output, where the logit for the forgetting class is set to negative infinity.
This approach enables \(\mathbf{\theta_u}\) to mimic a pseudo-retrained model, as the masked logits closely resemble those of Retrain. 
By aligning output behavior, HD approximates the intended unlearning effect.
Detailed explanation is available at~\Cref{app:detailed_head_distillation}.

% Setup
We evaluated HD on CIFAR-10~\citep{krizhevsky2009cifar} using ResNet-18~\citep{he2016resnet}, where the head is only a single linear layer.
For \textit{efficacy}, we used membership inference attack (MIA) and Jensen-Shannon divergence (JSD).
For \textit{accuracy} and \textit{efficiency}, we measured unlearning accuracy (UA), testing accuracy (TA), and run-time efficiency (RTE).
See~\cref{app:evaluation_metrics_detail} for metric details.
We compared HD with recent methods, including FT, RL~\citep{golatkar2020eternal}, GA~\citep{thudi2022unrolling}, $\ell_1$-sparse~\citep{liu2024l1sparse}, and SALUN~\citep{fan2023salun}. 
Details on the baselines can be found in~\cref{app:mu_baselines}.

\Cref{fig:exp__hd_rte_ta} shows the experimental results. 
Despite its simplicity, HD outperforms all other methods in MIA and ranking second in JSD. 
HD achieves this performance in just 6.2 seconds, approximately 30 to 60 times faster than competing methods. 
Additionally, HD maintains comparable testing accuracy (TA), effectively preserving task performance. 
All methods achieved perfect unlearning accuracy (100\% UA), which is omitted from~\cref{fig:exp__hd_rte_ta}. 
Notably, HD's strong performance generalize to multi-class and random data forgetting, as shown in~\cref{app:generalization_hd}.

The results indicate that HD performs exceptionally well across all black-box metrics. 
However, its validity as a MU algorithm requires scrutiny.
The primary issue is that HD closely resembles Original \(\mathbf{\theta_o}\), with changes limited to the single-layer head, while the encoder remains identical to \(\mathbf{\theta_o}\).
Consequently, all intermediate features related to the forget set are perfectly retained.
This raises a critical question:

\vspace{+0.5mm}
\begin{tcolorbox}[before skip=2mm, after skip=0.0cm, boxsep=0.0cm, middle=0.0cm, top=0.2cm, bottom=0.2cm]
%\vspace*{1mm}
\centering
\textit{
Do black-box metrics truly capture the unlearning quality,
or are they misled by superficial changes while deeper information persists?
}
\end{tcolorbox}
\vspace*{2mm}

%%%%%%%%%%%%%%%%%%%%%%%%%%%%%%%%%%%%%%%%%%%%%%%%%%%%%%%%%%%%
%%% Residual Information of Forgetting Data
%%%%%%%%%%%%%%%%%%%%%%%%%%%%%%%%%%%%%%%%%%%%%%%%%%%%%%%%%%%%
\subsection{Residual Information of Forgetting Data}
\label{subsection_residual_tsne_recovery}

To address the above question, we analyze recent unlearning methods to assess whether they internally remove information from the forget set, despite their strong performance on black-box metrics.
Note that analyses use the same experimental setup described in~\cref{subsection_HD}.

We begin with a qualitative analysis using t-SNE~\citep{tnse} visualizations of intermediate features from model encoders to compare Retrain and Original, and to examine internal behaviors of unlearning methods (\cref{fig:representation_tsne}). 
In \cref{subfig:representation_tsne_b}, the forgetting class (in purple) shows a scattered distribution in Retrain, indicating difficulty in forming coherent representations.
This scattering reflects an desirable outcome of strong unlearning, suggesting that the model has successfully `forgotten' how to encode meaningful semantic information from the forget set.

Notably, while the features from GA~\citep{thudi2022unrolling} appear scattered in a manner similar to Retrain, the features from RL~\citep{golatkar2020eternal} and SALUN~\citep{fan2023salun} closely resemble those of Original. 
In addition, HD, which shares the same encoder as Original, shows identical t-SNE results. 
These findings indicate that several unlearned models still retain a significant capacity to recognize the forgetting class, unlike Retrain.

\begin{figure}[htbp]
    \centering
    \includegraphics[width=.96\linewidth]{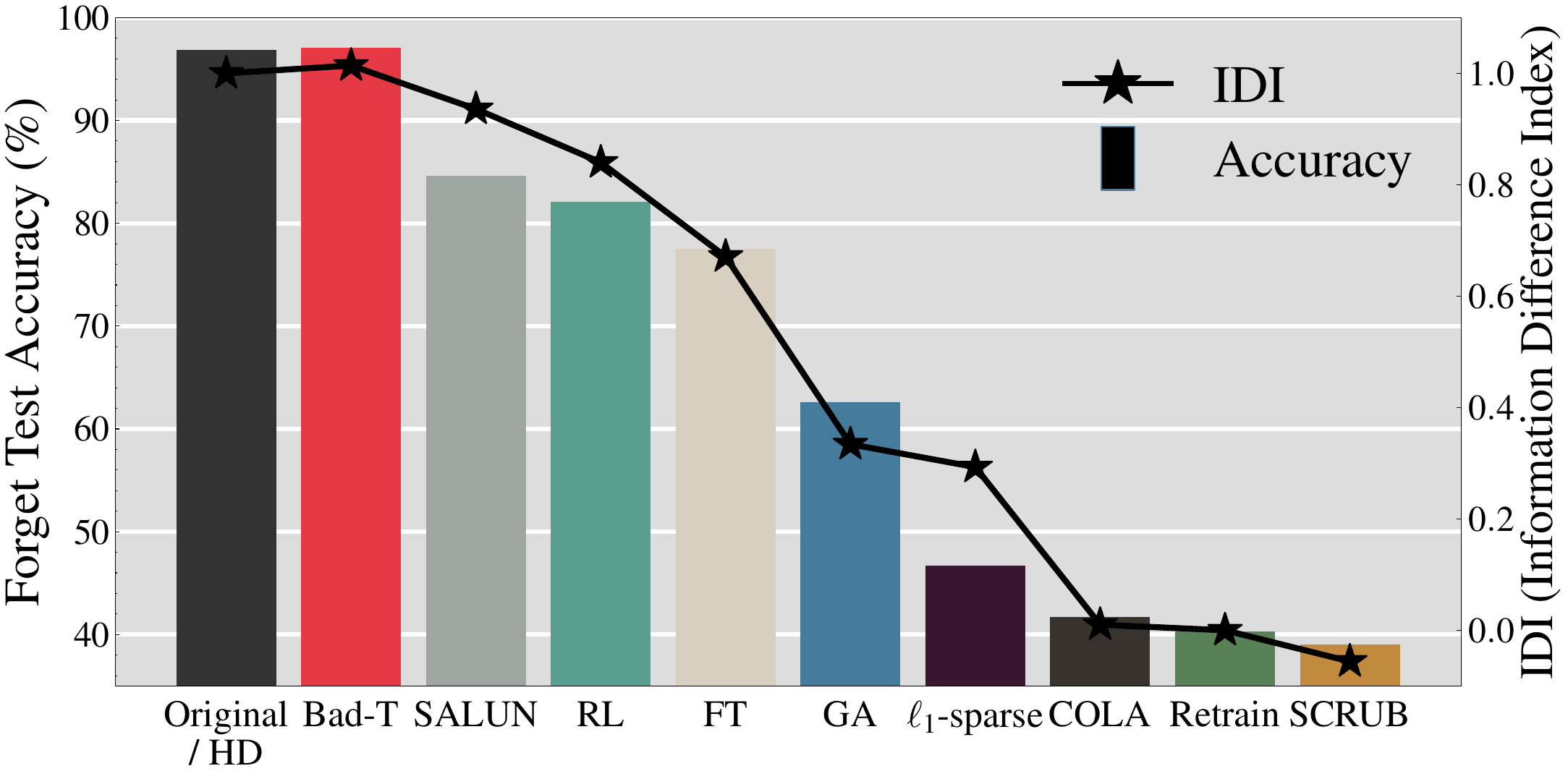}
    \centering
    \caption{
    Forget test accuracy and IDI (our metric in~\cref{section_idi}) for Original, Retrain, and MU methods (including COLA, our method in~\cref{section_cola}) after head retraining with fixed unlearned encoders using 2\% of $\mathcal{D}$ in (CIFAR-10, ResNet-18). IDI aligns with the recovered accuracy.}
    \label{fig:exp__recovery}
    % \vspace{-1em}
\end{figure}
To further examine the residual influence in unlearned models, we conducted a follow-up experiment inspired by time-based metrics (\eg,~\citet {chundawat2023ain}). 
We test whether unlearned encoders can recover forgotten information using minimal data.
Specifically, we replaced the heads of all models, including Retrain and Original, with randomly initialized ones. 
The encoders were then frozen, and new heads were trained on \( \mathcal{D'}\), a small subset (only 2\% of the total) of \( \mathcal{D}\) sampled at random.

After training, we evaluated the accuracy of the new models on the forget test data. 
Surprisingly, as shown in~\cref{fig:exp__recovery}, while the retrained head of Retrain achieves no more than 41\% accuracy, the heads from certain methods, like Bad-T, SALUN, and RL exhibit over 82\% accuracy.
Notably, the high accuracy observed in SALUN and RL aligns with their clustered t-SNE patterns in~\cref{fig:representation_tsne}.

The results from the above analyse demonstrate that unlearned models across various MU algorithms retain substantial residual influence from the forget set, indicating incomplete unlearning.
Critically, standard black-box metrics fail to capture these internal traces.
If such metrics cannot ensure strong unlearning, the reliability of approximate unlearning algorithms, which often lack theoretical guarantees, becomes questionable in real world applications. 
Therefore, developing practical white box approaches that consider internal model behaviors is essential to achieving the fundamental goal of unlearning.

%%%%%%%%%%%%%%%%%%%%%%%%%%%%%%%%%%%%%%%%%%%%%%%%%%%%%%%%%%%%
%%% An Information Theoretic Metric
%%%%%%%%%%%%%%%%%%%%%%%%%%%%%%%%%%%%%%%%%%%%%%%%%%%%%%%%%%%%
\section{An Information Theoretic Metric}

Black-box metrics often overlook residual information in intermediate layers, as shown in \cref{section_residualinformation}.
To address this, we measure residual information in the intermediate features of unlearned models using mutual information. 
Building on this,
we propose a novel white-box metric, IDI, that goes beyond output-based evaluations.

%%%%%%%%%%%%%%%%%%%%%%%%%%%%%%%%%%%%%%%%%%%%%%%%%%%%%%%%%%%%
%%% Quantifying Residual Information
%%%%%%%%%%%%%%%%%%%%%%%%%%%%%%%%%%%%%%%%%%%%%%%%%%%%%%%%%%%%%%%%%%%%%%%%
\subsection{Quantifying Residual Information}

\label{subsection_MI_estimation}
To quantify the relationship between intermediate features and data labels, we utilize Shannon's mutual information (MI), a robust measure that captures variable dependencies across dimensional complexities.
For an input \(\mathbf{X}\), let \(\mathbf{Z}^{(\mathbf{u})}_\ell\) and \(\mathbf{Z}^{(\mathbf{r})}_\ell\) denote the features from the \(\ell\)-th layer of the total \(L\)-layer encoder in the unlearned model and Retrain, respectively.
Let \(Y\) be a binary label indicating whether \(\mathbf{X}\) belongs to the forget set (\(Y=1\)) or not (\(Y=0\)). 
We compute MI \(I(\mathbf{Z}_\ell; Y)\) across each layer from 1 to \(L\),
to determine whether intermediate features retain information about the forget set.
For estimation, we adopt the InfoNCE loss~\citep{oord2018representation}, a robust method widely used in deep MI estimation~\citep{radford2021inconce_mm1_clip, jia2021infonce_mm2_align}.

Given a batch \(\cB = \{(U^{(k)}, V^{(k)}) : 1 \leq k \leq K\}\), sampled from a joint distribution \( P_{U,V} \), 
where \( U \in \mathcal{U} \) and \( V \in \mathcal{V} \) be random variables.
The InfoNCE loss~\citep{poole2019variational} is defined as:  
\[
% \mathcal{L}_{\text{NCE}}(\mathcal{B}, \nu, \eta) 
\mathcal{L}_{\text{NCE}}
= \frac{1}{K} \sum_{k=1}^K \log \frac{\exp(f_{\nu}(U^{(k)})^\top g_{\eta}(V^{(k)}))}{\frac{1}{K} \sum_{k'=1}^K \exp(f_{\nu}(U^{(k)})^\top g_{\eta}(V^{(k')}))},
\]
where \( f_\nu: \mathcal{U} \to \mathbb{R}^d \) and \( g_\eta: \mathcal{V} \to \mathbb{R}^d \) are critic functions,
with an output embedding dimension \( d \), parameterized by neural networks with parameters \(\nu\) and \(\eta\). 
This neural network parameterization, inspired by~\citet{radford2021inconce_mm1_clip},
effectively captures complex relationships in contrastive learning through flexible and expressive modeling of the joint distributions of \( U \) and \( V \).

The InfoNCE loss serves as a lower bound on the MI between \( U \) and \( V \).
In fact, the maximum value of the InfoNCE loss, when using the joint critic functions, equals the mutual information:
\[
I(U; V) = \max_{\nu, \eta} \mathcal{L}_{\text{NCE}}(\cB, \nu, \eta).
\]
By maximizing this loss over parameters \(\nu\) and \(\eta\) through neural networks, we effectively capture data structure and accurately quantify shared information between \( U \) and \( V \).

\begin{figure*}[t!]
    \centering
    \begin{subfigure}[b]{0.27\textwidth}
        \centering
        \includegraphics[width=\textwidth]{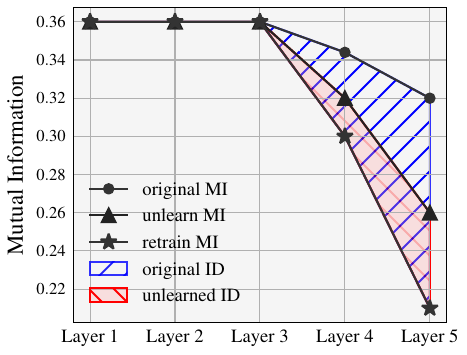}
        \caption{Illustraion of IDI}
        \label{fig:illust_idi}
    \end{subfigure}
    \hfill
    \begin{subfigure}[b]{0.7\textwidth}
        \centering
        \includegraphics[width=0.495\textwidth]{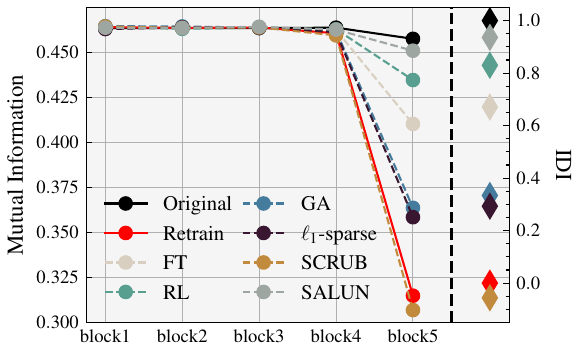}
        \includegraphics[width=0.495\textwidth]{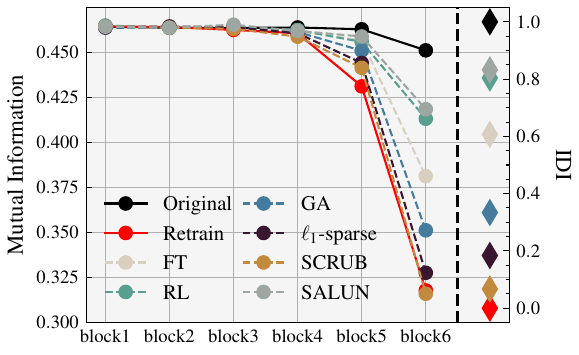}
        \caption{MI curves and IDI in single-class forgetting task}
        \label{fig:idi}
    \end{subfigure}
    \caption{
    (a) Conceptual illustration of IDI. Curves show estimated mutual information \(I(\mathbf{Z}_\ell; Y)\) for Original (\ding{108}), unlearned (\ding{115}), and Retrain (\ding{72}).
    IDI is the ratio \(\frac{\textcolor{red}{\textbf{ID}(\mathbf{\theta_u})}}{\textcolor{blue}{\textbf{ID}(\mathbf{\theta_o})}}\), corresponding to the red area divided by the blue area. % Best viewed in color.
    (b) MI curves and IDI values for Original, Retrain, and unlearned models (FT, RL, GA, \(\ell_1\)-sparse, SCRUB, SALUN) on CIFAR-10 across ResNet-18 (left) and ResNet-50 (right) blocks, averaged over five trials. See~\cref{app: C.2} for standard deviations.   
    }
    \label{fig:mi__resnet18_resnet50_cifar10_class}
    % \vspace{-0.5em}
\end{figure*}

To estimate mutual information (MI) at each layer, we define separate critic functions for every layer: \( f_{\nu_\ell} \) and \( g_{\eta_\ell} \), where \( \ell \in \{1, \dots, L\} \) denotes the layer index. 
The critic \( g_{\eta_\ell} \) models the binary variable \( Y \) as two trainable \( d \)-dimensional vectors, \( g_{\eta_\ell}(0) \) and \( g_{\eta_\ell}(1) \), selecting the appropriate one based on the value of \( Y \). 
In parallel, \( f_{\nu_\ell} \) maps intermediate features \( \mathbf{Z}_\ell \) from the \(\ell\)-th encoder layer to a shared \( d \)-dimensional embedding space. 
The parameters $\nu_\ell$ define the weights and biases of this neural network.

The complexity of $f_{\nu_\ell}$ depends on the layer depth: in earlier layers, \( f_{\nu_\ell} \) processes raw, less interpretable features, requiring a more intricate design
to capture the relationship between $\mathbf{Z}_\ell$ and $Y$.
In later layers, with more structured features, \( f_{\nu_\ell} \) can perform the mapping more directly.
This design enables the accurate estimation of \( I(\mathbf{Z}_\ell; Y) \), capturing the dependency between features and labels at different depths.
For details on \( f_{\nu_\ell} \) and \( g_{\eta_\ell} \), refer to~\cref{app:parameterization}.

To achieve a model-agnostic design, we construct \( f_{\nu_\ell} \) by reusing the network layers from \( \ell+1 \) to \( L \).
This approach allows us to approximate the mutual information between the output and intermediate features at layer \(\ell\) without requiring network redesign for each layer,
thus maintaining flexibility and scalability.
To ensure dimensional compatibility between \(f\) and \(g\),
we introduce an additional linear projection layer so that \(f_{\nu_\ell}(\mathbf{Z}_\ell)\) outputs a $d$-dimensional feature.

During optimization, we freeze the parameters of the model up to the \( \ell \)-th layer and reuse the subsequent layers, from \( \ell+1 \) to \( L \), as \( f_{\nu_\ell} \). 
These layers, together with the projection layer, are randomly initialized and trained to optimize the InfoNCE objective. 
Both the retain and forget sets are used to provide representations for \( Y = 0 \) and \( Y = 1 \), ensuring that information from both outcomes is captured for mutual information estimation.

This approach enables \(f_{\nu_\ell}\) to effectively exploit intermediate features \(\mathbf{Z}_\ell\) to classify \(Y\),
providing deeper insights into the model’s internal information processing at each layer.
It also reveals the model’s capacity to extract and utilize relevant information for distinguishing between output labels,
offering a clearer understanding of the information dynamics across the network.

\begin{table*}[t!]
    \centering

    \resizebox{\textwidth}{!}{%
    \begin{tabular}{lcccccccccccc}
        \toprule
        &\multicolumn{6}{c}{\textbf{CIFAR-10 (single class)}} &\multicolumn{6}{c}{\textbf{ImageNet-1K (five classes)}} \\
        \cmidrule(lr){1-1} \cmidrule(lr){2-7} \cmidrule(lr){8-13}
        Methods & UA & RA & TA & MIA & \textbf{IDI} & RTE (min) 
        & UA & RA & TA & MIA & \textbf{IDI} & RTE (min) \\
        \cmidrule(lr){1-1} \cmidrule(lr){2-7} \cmidrule(lr){8-13}
        Retrain
        & $100.0$ & $100.0$ & $95.64$ & $10.64$ & $0.0$ & $154.56$
        & $100.0$ & $88.80$ & $75.88$ & $9.41$ & $0.0$ & $2661.90$\\
        \cmidrule(lr){1-1} \cmidrule(lr){2-7} \cmidrule(lr){8-13}
        HD
        & {$\textbf{100.0}_{\pm 0.0}$} & {$\textbf{100.0}_{\pm 0.0}$} &{$95.22_{\pm 0.07}$} & {$2.05_{\pm 0.11}$} & {$1.000_{\pm 0.0}$} & {$\textbf{0.10}_{\pm 0.01}$} & {$\textbf{100.0}_{\pm 0.0}$} & {$87.94_{\pm 0.16}$} & {$\underline{75.60}_{\pm 0.07}$} & {$7.12_{\pm 0.12}$} & {$1.000_{\pm 0.0}$} & {$\textbf{4.75}_{\pm 0.03}$} \\
        FT
        & $\textbf{100.0}_{\pm 0.0}$ & $\textbf{100.0}_{\pm 0.0}$ & $95.12_{\pm 0.09}$
        & $0.17_{\pm 0.05}$ & $0.671_{\pm 0.008}$ & $6.44_{\pm 0.07}$
        & $\textbf{100.0}_{\pm 0.0}$ & $\textbf{88.52}_{\pm 0.0}$ & $\underline{76.16}_{\pm 0.01}$
        & $8.24_{\pm 1.23}$ & $0.102_{\pm 0.026}$ & $140.04_{\pm 1.42}$\\
        RL
        & $99.93_{\pm 0.01}$ & $\textbf{100.0}_{\pm 0.0}$ & $\textbf{95.66}_{\pm 0.05}$
        & $0.0_{\pm 0.0}$ & $0.830_{\pm 0.005}$ & $3.09_{\pm 0.03}$
        & $\underline{99.96}_{\pm 0.03}$ & $86.46_{\pm 0.07}$ & $75.23_{\pm 0.01}$
        & $0.23_{\pm 0.01}$ & $1.002_{\pm 0.007}$ & $200.73_{\pm 1.87}$\\
        GA
        & $\textbf{100.0}_{\pm 0.0}$ & $99.06_{\pm 0.25}$ & $93.10_{\pm 0.50}$
        & $25.37_{\pm 3.24}$ & $0.334_{\pm 0.014}$ & $4.00_{\pm 0.08}$
        & $\textbf{100.0}_{\pm 0.0}$ & $80.77_{\pm 0.22}$ & $71.49_{\pm 0.10}$
        & $4.20_{\pm 0.46}$ & $0.328_{\pm 0.023}$ & $212.14_{\pm 2.61}$\\ 
        Bad-T
        & $99.90_{\pm 0.14}$ & $\underline{99.99}_{\pm 0.0}$ & $94.99_{\pm 0.12}$
        & $68.17_{\pm 42.80}$ & $1.014_{\pm 0.004}$ & $4.64_{\pm 0.05}$
        & $98.01_{\pm 0.02}$ & $84.03_{\pm 0.03}$ & $73.42_{\pm 0.03}$
        & $69.13_{\pm 12.57}$ & $1.152_{\pm 0.011}$ & $211.52_{\pm 0.96}$\\
        % \blue{BoundaryExpand}
        % & \blue{$71.39_{\pm 0.31}$} & \blue{$99.20_{\pm 0.04}$} & \blue{$92.53_{\pm 0.02}$} &
        % \blue{$7.69_{\pm 0.33}$} & \blue{$0.892_{\pm 0.001}$} & \blue{$\underline{0.19}_{\pm 0.01}$}
        % & \blue{${77.22}_{\pm 0.11}$} & \blue{$82.79_{\pm 0.08}$} & \blue{$71.78_{\pm 0.09}$}
        % & \blue{$1.43_{\pm 0.51}$} & \blue{$0.628_{\pm 0.005}$} & \blue{${5.14}_{\pm 0.02}$} \\ 
        % \blue{BoundaryShrink}
        % & \blue{$85.16_{\pm 0.42}$} & \blue{$99.60_{\pm 0.17}$} & \blue{$93.48_{\pm 0.40}$}
        % & \blue{$0.25_{\pm 0.43}$} & \blue{$0.887_{\pm 0.009}$} & \blue{${0.59}_{\pm 0.02}$}
        % & \blue{${91.20}_{\pm 0.02}$} & \blue{$81.41_{\pm 0.17}$} & \blue{$70.55_{\pm 0.08}$}
        % & \blue{$1.45_{\pm 0.34}$} & \blue{$0.543_{\pm 0.011}$} & \blue{$\underline{4.81}_{\pm 0.03}$} \\ 
        EU-5
        & $\textbf{100.0}_{\pm 0.0}$ & $\textbf{100.0}_{\pm 0.0}$ & $95.25_{\pm 0.02}$
        & $0.06_{\pm 0.03}$ & $0.528_{\pm 0.005}$ & ${1.54}_{\pm 0.0}$
        & $\textbf{100.0}_{\pm 0.0}$ & $79.62_{\pm 0.0}$ & $71.22_{\pm 0.13}$
        & $13.33_{\pm 1.53}$ & $0.183_{\pm 0.028}$ & $193.38_{\pm 0.78}$\\ 
        CF-5
        & $98.13_{\pm 1.39}$ & $\textbf{100.0}_{\pm 0.0}$ & $\underline{95.54}_{\pm 0.09}$
        & $0.0_{\pm 0.0}$ & $0.675_{\pm 0.027}$ & ${1.57}_{\pm 0.03}$
        & $\textbf{100.0}_{\pm 0.0}$ & $84.31_{\pm 0.08}$ & $74.16_{\pm 0.06}$
        & $10.21_{\pm 5.33}$ & $0.701_{\pm 0.014}$ & ${81.53}_{\pm 0.56}$\\ 
        EU-10
        & $\textbf{100.0}_{\pm 0.0}$ & $99.50_{\pm 0.02}$ & $93.61_{\pm 0.08}$
        & $15.24_{\pm 1.08}$ & $-0.349_{\pm 0.019}$ & $2.42_{\pm 0.11}$
        & $\textbf{100.0}_{\pm 0.0}$ & $71.84_{\pm 0.03}$ & $65.78_{\pm 0.02}$
        & $16.65_{\pm 1.91}$ & $\underline{-0.051}_{\pm 0.021}$ & $193.79_{\pm 0.47}$\\ 
        CF-10
        & $\textbf{100.0}_{\pm 0.0}$ & $99.98_{\pm 0.0}$ & $94.95_{\pm 0.05}$
        & $\textbf{11.61}_{\pm 0.91}$ & $-0.060_{\pm 0.017}$ & $2.31_{\pm 0.03}$
        & $\textbf{100.0}_{\pm 0.0}$ & $80.87_{\pm 0.04}$ & $72.34_{\pm 0.08}$
        & $13.99_{\pm 5.41}$ & $0.608_{\pm 0.012}$ & ${82.29}_{\pm 0.34}$\\ 
        SCRUB
        & $\textbf{100.0}_{\pm 0.0}$ & $\textbf{100.0}_{\pm 0.0}$ & $95.37_{\pm 0.04}$
        & $19.73_{\pm 1.92}$ & $\underline{-0.056}_{\pm 0.008}$ & $3.49_{\pm 0.02}$
        & $99.28_{\pm 0.07}$ & $\underline{88.39}_{\pm 0.04}$ & $76.51_{\pm 0.03}$
        & $7.42_{\pm 0.51}$ & $0.517_{\pm 0.011}$ & $426.04_{\pm 2.98}$\\ 
        SALUN
        & $\underline{99.99}_{\pm 0.01}$ & $\textbf{100.0}_{\pm 0.0}$ & $95.42_{\pm 0.12}$
        & $0.01_{\pm 0.01}$ & $0.936_{\pm 0.012}$ & $3.54_{\pm 0.11}$
        & $89.67_{\pm 0.27}$ & $86.25_{\pm 0.15}$ & $75.54_{\pm 0.10}$
        & $0.50_{\pm 0.09}$ & $0.343_{\pm 0.017}$ & $793.82_{\pm 3.32}$\\ 
        \(\mathbf{\ell_1}\)-sparse
        & $\textbf{100.0}_{\pm 0.0}$ & $99.93_{\pm 0.02}$ & $94.90_{\pm 0.10}$
        & $1.56_{\pm 0.09}$ & $0.293_{\pm 0.012}$  & $2.96_{\pm 0.03}$
        & $97.57_{\pm 0.61}$ & $85.33_{\pm 0.07}$ & $74.77_{\pm 0.03}$
        & $\underline{8.84}_{\pm 1.39}$ & $0.239_{\pm 0.031}$ & $226.74_{\pm 1.35}$\\ 
        \cmidrule(lr){1-1} \cmidrule(lr){2-7} \cmidrule(lr){8-13}
        \textbf{COLA}
        & $\textbf{100.0}_{\pm 0.0}$ & $\textbf{100.0}_{\pm 0.0}$ & $95.36_{\pm 0.06}$
        & $\underline{12.64}_{\pm 0.92}$ & $\textbf{0.010}_{\pm 0.006}$  & $4.91_{\pm 0.04}$
        & $\textbf{100.0}_{\pm 0.0}$ & $87.93_{\pm 0.05}$ & $\textbf{76.15}_{\pm 0.04}$
        & $\textbf{9.95}_{\pm 1.21}$ & $\textbf{0.040}_{\pm 0.042}$ & $171.44_{\pm 0.75}$ \\ 
        \bottomrule
\end{tabular}
    }
\caption{Performance summary of MU methods (including COLA and \blue{14} other baselines) for class-wise forgetting task on (CIFAR-10, ResNet-18) and (ImageNet-1K, ResNet-50).
A better performance of an MU method corresponds to a smaller performance gap with Retrain (except RTE), with the top method in \textbf{bold} and the second best \underline{underlined}.
}
\label{tab:main_baseline}
\end{table*}

%%%%%%%%%%%%%%%%%%%%%%%%%%%%%%%%%%%%%%%%%%%%%%%%%%%%%%%%%%%%%%%%%%%%%%%%
%%% Residual Information in Unlearned Models
%%%%%%%%%%%%%%%%%%%%%%%%%%%%%%%%%%%%%%%%%%%%%%%%%%%%%%%%%%%%%%%%%%%%%%%%
\subsection{Residual Information in Unlearned Models} 
\label{subsection_analze_with_IDI}

We begin by plotting the estimated MI between the intermediate layers and the binary label indicating whether the data belong to the forget set, as shown in~\cref{fig:idi}.
As expected, MI decreases across layers, aligning with the Information Bottleneck principle~\citep{tishby2000ib_principle}.
This figure also reveals the internal behaviors of unlearned models that black-box assessments fail to capture.

In particular, SCRUB and \(\ell_1\)-sparse, which approximate the MI levels of Retrain,
are more likely to achieve the MU objective at the feature level across both ResNet architectures.
Their lower MI suggests that their encoders, like Retrain, struggle to differentiate between the forget set and the retain set. 
% reflects the difficulty that their encoders face in distinguishing between the forget set and the retain set, similar to Retrain.
Conversely, SALUN and RL show MI curves that are close to that of Original, indicating the opposite.
Note that HD produces the identical curve as Original, as its encoder remains unchanged.
We observe similar patterns in CIFAR-100 and ImageNet-1K, as well as in ViT.
Additionally, extending our experiment to multi-class forgetting tasks (\eg, 20 classes on CIFAR-100) reveals more pronounced MI differences between Retrain and Original.
See~\cref{app: D.1} for further results.

%%%%%%%%%%%%%%%%%%%%%%%%%%%%%%%%%%%%%%%%%%%%%%%%%%%%%%%%%%%%%%%%%%%%%%%%
%%% Information Difference Index (IDI)
%%%%%%%%%%%%%%%%%%%%%%%%%%%%%%%%%%%%%%%%%%%%%%%%%%%%%%%%%%%%%%%%%%%%%%%%
\subsection{Information Difference Index (IDI)}
\label{section_idi} 

\begin{figure}[h]%[htbp]
    \centering
    \includegraphics[width=0.7\linewidth]{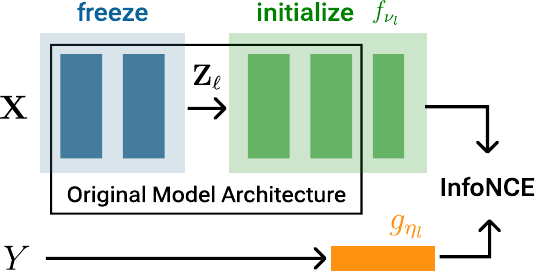}
    \caption{
    Illustration of estimating MI using InfoNCE. \(f_{\nu_\ell}\) represents a trainable network to capture features from \(\mathbf{Z}_\ell\), while \(g_{\eta_\ell}\) handles the binary input \(Y\).
    }
    \label{fig:illust_infonce}
% \vskip -.1in
\end{figure}
Motivated from the above experiment, we define the \textbf{information difference (ID)} of \(\mathbf{\theta_u}\) as the MI difference across intermediate layers between the unlearned model and Retrain, calculated as:
\begin{equation}
\mathbf{ID}(\mathbf{\theta_u}) = \sum_{\ell=1}^{L}
 \bigl(I(\mathbf{Z}^{(\mathbf{u})}_\ell; Y) - I(\mathbf{Z}^{(\mathbf{r})}_\ell; Y)\bigr).
\label{eq:id}
\end{equation}
ID of \(\mathbf{\theta_u}\) shows the extent of information retention through ensuing layers of the unlearned encoder. 
To provide a normalized measure, we introduce the \textbf{information difference index (IDI)}:
\begin{equation}
\mathbf{IDI}(\mathbf{\theta_u}) 
= \frac{\mathbf{ID}(\mathbf{\theta_u})}{\mathbf{ID}(\mathbf{\theta_0})}
= \frac{\sum_{\ell=1}^{L} \bigl(I(\mathbf{Z}^{(\mathbf{u})}_\ell; Y) - I(\mathbf{Z}^{(\mathbf{r})}_\ell; Y)\bigr)}{\sum_{\ell=1}^{L} \bigl(I(\mathbf{Z}^{(\mathbf{o})}_\ell; Y) - I(\mathbf{Z}^{(\mathbf{r})}_\ell; Y)\bigr)} 
,
\label{eq:idi}
\end{equation}
where \(\mathbf{Z}^{(\mathbf{o})}_\ell\) is the output of the \(\ell\)-th layer of Original encoder.
\cref{fig:illust_idi} illustrates IDI, which is conceptually the ratio of the areas between MI curves.

However, computing MI for all $L$ layers can be expensive.
In practice, we compute MI from the last $n$ layers (\ie, later blocks), where $n \ll L$,
as earlier layers show negligible differences between Retrain and Original~(\cref{fig:idi}).
To reduce overhead, we reuse the original network structure for MI estimation (see~\cref{fig:illust_infonce}).
With this setup, computing IDI on CIFAR-100 with ResNet-18 takes under 5 minutes.
Further details and cost analysis are provided in~\cref{Effect_of_number_of_Layers_for_IDI}.

IDI quantifies the information gap between the unlearned model and Retrain. 
An IDI of 0 denotes that the unlearned model has completely removed all information related to the forget set, achieving indistinguishability from Retrain.
In contrast, an IDI of 1 indicates that the encoder retains all the information found in Original.
Interestingly, a negative IDI value, termed \textit{over-unlearning}, occurs when the model removes more information than Retrain.
We also demonstrate IDI for random data forgetting in~\cref{app:idi_random}.

Note that as the denominator of IDI ($\mathbf{ID}(\theta_o)$) approaches zero, IDI may yield unexpected values. 
However, this case indicates that Original and Retrain are nearly identical, suggesting minimal unlearning utility. 
Thus, $\mathbf{ID}(\theta_o)$ can serve as an indicator for the necessity of unlearning in practice.

%%%%%%%%%%%%%%%%%%%%%%%%%%%%%%%%%%%%%%%%%%%%%%%%%%%%%%%%%%%%
%%% Experiments 
%%%%%%%%%%%%%%%%%%%%%%%%%%%%%%%%%%%%%%%%%%%%%%%%%%%%%%%%%%%%
\section{Experiments} 
\label{section_experiments}

\subsection{COLlapse and Align (COLA) Approach} 
\label{section_cola}
As discussed in both~\cref{section_residualinformation} and~\ref{subsection_analze_with_IDI}, we observed residual information in the intermediate layers of several unlearned models, despite their outputs being similar to those of Retrain.
To address this, we propose a robust two-step unlearning framework, \textbf{COLlapse and Align (COLA)}, consisting of a \textit{collapse phase} and an \textit{alignment phase} to directly remove residual information at the feature level.

During the \textit{collapse phase}, COLA eliminates feature-level information by applying supervised contrastive loss~\citep{khosla2020supervised} to encoder outputs.
Rather than dispersing features from the forget set, which could harm model performance, COLA applies the loss to the retain set, promoting tight intra-class clustering.
As these clusters shrink, features from the forget set are forced to collapse into the clusters of the retain set, achieving catastrophic forgetting.
After feature collapsing, the \textit{alignment phase} optimizes the entire model using cross-entropy loss on the retain set to align the encoder and head.
For an illustration of COLA, as well as COLA+, a method tailored for random data forgetting, and detailed loss formulations, refer to~\cref{app:cola_pseudocode}.

\subsection{Evaluation of Unlearning Methods with IDI}
\label{section_evaluation}

We demonstrate the utility of IDI as a valuable efficacy metric and highlight the strong performance of COLA and its variant COLA+ through extensive experiments.
Our experiments cover three datasets: CIFAR-10, CIFAR-100~\citep{krizhevsky2009cifar}, and ImageNet-1K~\citep{DenDon09Imagenet}, 
and three model architectures: ResNet-18, ResNet-50~\citep{he2016resnet}, and ViT~\citep{dosovitskiy2020vit}.
For simplicity, we approximate IDI using the features from blocks rather than every layer in ResNet and ViT. 
See~\cref{app:experiment_detail} for experimental details.

%%%%%%%%%%%%%%%%%%%%%%%%%%%%%%%%%%%%%%%%%%%
\begin{table}[h]
 % \vskip -.12in
% \captionsetup{
%   labelfont={color=blue}, % "표 1" 등의 레이블을 파란색으로
%   textfont={color=blue}   % 캡션 텍스트를 파란색으로
% }
    \centering
    
    % \vskip -.08in
    \resizebox{\linewidth}{!}{%
    \begin{tabular}{lcccccccccccc}
        \toprule
        \multicolumn{8}{c}{\textbf{CIFAR-10 (500 samples per class)}} \\
        \midrule
        Methods & UA & RA & TA & MIA & JSD & \textbf{IDI} & RTE (min)\\
        \midrule
        % Original
        % & $0.0$
        % & $100.0$
        % & $95.54$
        % & $92.90$
        % & $0.09$
        % & $1.000$
        % & $170.32$
        % \\ 
        Retrain
        & $3.94$
        & $100.0$
        & $95.26$
        & $75.12$
        & $0.0$
        & $0.0$
        & $152.87$
        \\ \midrule
        \blue{HD} 
        & $\underline{3.64}_{\pm 1.66}$
        & $97.93_{\pm 1.38}$
        & $92.80_{\pm 1.18}$
        & $\underline{77.47}_{\pm 4.09}$
        & $0.08_{\pm 0.04}$
        & $1.000_{\pm 0.0}$
        & $\textbf{0.30}_{\pm 0.05}$
        \\ 
        FT
        & $5.03_{\pm 0.40}$
        & $98.95_{\pm 0.21}$
        & $92.94_{\pm 0.26}$
        & $83.52_{\pm 0.58}$
        & $0.07_{\pm 0.11}$
        & $\underline{-0.069}_{\pm 0.013}$
        & $8.11_{\pm 0.03}$\\
        RL
        & $4.77_{\pm 0.27}$
        & $\textbf{99.92}_{\pm 0.0}$
        & $\textbf{93.54}_{\pm 0.04}$
        & $22.47_{\pm 1.19}$
        & $0.38_{\pm 0.02}$
        & $0.084_{\pm 0.030}$
        & $2.75_{\pm 0.01}$\\
        GA
        & $2.86_{\pm 0.76}$
        & $98.37_{\pm 0.71}$
        & $91.90_{\pm 0.70}$
        & $85.49_{\pm 2.17}$
        & $0.09_{\pm 0.01}$
        & $0.924_{\pm 0.028}$
        & $4.31_{\pm 0.03}$\\
        Bad-T
        & $5.47_{\pm 1.05}$
        & $\underline{99.87}_{\pm 0.05}$
        & $91.51_{\pm 0.61}$
        & $39.53_{\pm 3.43}$
        & $0.27_{\pm 0.03}$
        & $0.939_{\pm 0.053}$
        & $4.78_{\pm 0.09}$\\
        EU-10
        & $3.16_{\pm 0.19}$
        & $98.68_{\pm 0.08}$
        & $93.07_{\pm 0.12}$
        & ${83.40}_{\pm 0.21}$
        & $\underline{0.06}_{\pm 0.01}$
        & $-0.110_{\pm 0.013}$
        & ${2.13}_{\pm 0.05}$\\ 
        CF-10
        & $2.71_{\pm 0.24}$
        & $99.11_{\pm 0.06}$
        & $\underline{93.47}_{\pm 0.15}$
        & $84.33_{\pm 0.05}$
        & $\textbf{0.05}_{\pm 0.01}$
        & $0.219_{\pm 0.029}$
        & $\underline{2.10}_{\pm 0.06}$\\ 
        SCRUB
        & ${4.31}_{\pm 1.50}$
        & $96.21_{\pm 1.70}$
        & $88.83_{\pm 1.86}$
        & $37.88_{\pm 7.65}$
        & $0.56_{\pm 0.09}$
        & $0.322_{\pm 0.016}$
        & $3.37_{\pm 0.05}$\\ 
        SALUN
        & $2.74_{\pm 0.30}$
        & $97.77_{\pm 0.04}$
        & $91.68_{\pm 0.44}$
        & $83.52_{\pm 2.20}$
        & $0.10_{\pm 0.03}$
        & $0.861_{\pm 0.012}$
        & $5.69_{\pm 0.04}$\\
        \(\mathbf{\ell_1}\)-sparse
        & $5.47_{\pm 0.22}$
        & $96.66_{\pm 0.07}$
        & $91.31_{\pm 0.25}$
        & $\textbf{77.12}_{\pm 0.21}$
        & $0.09_{\pm 0.01}$
        & $-0.157_{\pm 0.026}$
        & $3.03_{\pm 0.04}$\\ 
        \midrule
        % \textbf{COLA}
        % & $1.22_{\pm 0.07}$
        % & $\textbf{99.99}_{\pm 0.01}$
        % & $94.71_{\pm 0.05}$
        % & $88.95_{\pm 0.31}$
        % & $0.07_{\pm 0.01}$
        % & $0.849_{\pm 0.076}$
        % & $6.95_{\pm 0.03}$\\ 
        \textbf{COLA+}
        & $\textbf{3.90}_{\pm 0.08}$
        & $99.24_{\pm 0.17}$
        & $93.23_{\pm 0.09}$
        & $83.48_{\pm 0.10}$
        & $\underline{0.06}_{\pm 0.01}$
        & $\textbf{0.024}_{\pm 0.010}$
        & $7.80_{\pm 0.02}$\\
        \bottomrule
%%%%%%%%%%%%%%%%%%%%%%%%%%%%%%%%%%%%%%%%%%%
\end{tabular}
}
\caption{
    Performance summary for random data forgetting on (CIFAR-10, ResNet-18), with the top method in \textbf{bold} and the second best \underline{underlined}. }
\label{tab:main_supcon_random}
\end{table}
\Cref{tab:main_baseline} shows the experimental results on CIFAR-10 and ImageNet-1K in class-forgetting tasks.
At first glance, excluding the IDI column, several methods show similar accuracy (UA, RA, TA) but greater deviations in efficacy (MIA) and efficiency (RTE). 
This suggests that previous unlearning studies likely ranked MU methods based on MIA and RTE. 
However, as discussed earlier, relying solely on black-box metrics can be misleading, as they fail to account for residual information.
Indeed, some methods show strong MIA performance but fail to remove forget data from intermediate layers, as reflected by high IDI values.
For instance, CF-5 on ImageNet-1K achieves a favorable MIA value (10.21) close to Retrain (9.41) in the shortest time (81.53 min), yet its IDI (0.701) shows significant retention of forget data.
Similarly, EU-5 on CIFAR-10, which appears highly efficient (1.54 min), presents a high IDI (0.528), suggesting that its efficiency stems from incomplete unlearning.
The discrepancy between black-box metrics (MIA, JSD) and IDI is similarly observed in random data forgetting, as shown in \cref{tab:main_supcon_random}, particularly for methods like SALUN.
By incorporating IDI alongside existing metrics, we gain a more comprehensive and insightful evaluation of MU methods.

%%%%%%%%%%%%%%%%%%%%%%%%%%%%%%%%%%%%%%%%%%%%%%%%%%%%%%%%%%%%
%%% Discussions
%%%%%%%%%%%%%%%%%%%%%%%%%%%%%%%%%%%%%%%%%%%%%%%%%%%%%%%%%%%%
\subsection{Discussions}
\label{section_idi_property}

\paragraph{IDI as a Real-World Efficacy Metric.}
Accuracy metrics (UA, RA, TA) and efficacy metrics (MIA, JSD), commonly used in recent unlearning studies, require the presence of Retrain as a gold standard to compare model outputs.
While this approach is crucial for advancing MU methods in controlled experimental settings, where the field of unlearning for DNNs is still in its infancy, it becomes impractical in real-world applications where Retrain is unavailable.
Similar to current black-box metrics, the original formulation of IDI (see Equations~\ref{eq:id} and~\ref{eq:idi}) uses Retrain as a reference to assess unlearning efficacy. 
However, IDI allows for flexibility by using any available unlearned model as the reference.
Although the absence of Retrain changes the interpretation of IDI (\ie, an IDI of zero means complete unlearning as Retrain), 
it still provides valuable insights relative to the chosen reference.
This adaptability enhances the IDI's practicality, making it useful for evaluating unlearned models even in real-world scenarios. 
A detailed explanation and examples are provided in~\cref{app: D.2}.

\paragraph{IDI compare to White-Box MIA.}
\begin{table}[h]
    \centering
    
    \resizebox{\linewidth}{!}{%
        \begin{tabular}{lcccccccc}
            \toprule
            &\multicolumn{3}{c}{\textbf{CIFAR-10}} &\multicolumn{3}{c}{\textbf{CIFAR-100}} \\
            \cmidrule(lr){1-1} \cmidrule(lr){2-4} \cmidrule(lr){5-7}
            Methods & Activation & Gradient & \textbf{IDI}
            & Activation & Gradient & \textbf{IDI}\\
            \cmidrule(lr){1-1} \cmidrule(lr){2-4} \cmidrule(lr){5-7}
            Original & $99.98_{\pm 0.03}$ & $100.0_{\pm 0.0}$ & $1.000$ 
                     & $53.13_{\pm 2.88}$ & $61.34_{\pm 3.23}$ & $1.000$ \\
            Retrain  & $94.89_{\pm 1.07}$ & $95.13_{\pm 1.12}$ & $0.000$ 
                     & $52.87_{\pm 6.15}$ & $59.12_{\pm 4.12}$ & $0.000$ \\
            Random & $52.89_{\pm 41.03}$ & $45.23_{\pm 23.04}$ & $-1.281_{\pm 0.018}$ 
                        & $53.20_{\pm 5.15}$ & $47.12_{\pm 7.21}$ & $-2.955_{\pm 0.046}$ \\
            \cmidrule(lr){1-1} \cmidrule(lr){2-4} \cmidrule(lr){5-7}
            RL & $100.0_{\pm 0.0}$ & $99.98_{\pm 0.01}$ & $0.830_{\pm 0.005}$ 
               & $93.20_{\pm 3.53}$ & $95.30_{\pm 0.82}$ & $0.467_{\pm 0.010}$\\
            GA & $97.07_{\pm 0.35}$ & $96.01_{\pm 0.13}$ & $0.334_{\pm 0.014}$ 
               & $97.44_{\pm 2.12}$ & $82.44_{\pm 0.95}$ & $0.392_{\pm 0.021}$ \\
            EU-10 & $86.13_{\pm 4.78}$ & $89.42_{\pm 2.32}$ & $-0.349_{\pm 0.019}$ 
                  & $64.41_{\pm 1.65}$ & $72.13_{\pm 4.13}$ & $-0.221_{\pm 0.009}$\\
            CF-10 & $97.99_{\pm 0.38}$ & $98.33_{\pm 0.23}$ & $-0.060_{\pm 0.017}$ 
                  & $21.62_{\pm 0.61}$ & $23.15_{\pm 1.23}$ & $0.175_{\pm 0.040}$\\
            SCRUB & $99.43_{\pm 0.09}$ & $99.15_{\pm 0.05}$ & $-0.056_{\pm 0.008}$ 
                  & $46.44_{\pm 1.28}$ & $62.31_{\pm 1.73}$ & $0.339_{\pm 0.069}$ \\
            COLA & $92.26_{\pm 0.08}$ & $93.12_{\pm 0.11}$ & $0.010_{\pm 0.006}$
                 & $61.08_{\pm 0.23}$ & $65.24_{\pm 0.43}$ & $-0.037_{\pm 0.006}$\\
            \bottomrule
        \end{tabular}}

    \caption{
   Performance of MU methods on white-box MIAs (Activation, Gradient) and IDI for single-class forgetting on ResNet-18. MIA values represent the attack success rate (\%) for distinguishing forgetting samples. ``Random'' refers to a model randomly initialized without prior training.
    }
    \label{tab:main_white}
\end{table}

While black-box MIA, adapted from privacy studies, is widely used as an evaluation tool in unlearning literature, we explore the potential of white-box MIA, which has not traditionally been employed for this purpose, and compare it with IDI.
Specifically, we evaluate two white-box MIA methods: one leveraging model activations and another utilizing gradients~\citep{nasr2019comprehensive}.
\Cref{tab:main_white} presents the results of white-box MIA and IDI in single-class forgetting scenarios. 
White-box MIA delivers consistent results on CIFAR-10 but becomes unstable as the dataset scales to CIFAR-100, with significant variability in MIA values across algorithms. 
This instability is further highlighted with a randomly initialized model, which produces MIA values comparable to Retrain despite no actual training.
In contrast, IDI provides stable and interpretable results, yielding strongly negative values for randomly initialized models, accurately reflecting their lack of residual information. 
This underscores IDI's reliability as a robust and interpretable metric for unlearning evaluation.

\section{Conclusion}
Black-box metrics often overlook critical intermediate features, limiting their effectiveness in assessing unlearning. 
To address this, we propose the Information Difference Index (IDI), a practical white-box metric that quantifies residual information in model features.
Our experiments confirm IDI's effectiveness for robust unlearning assessment across diverse datasets and architectures.
Additionally, we introduce COLA, a novel unlearning algorithm that collapses and realigns feature representations to eliminate residual information. 
By examining deep representational traces rather than just superficial outputs, our work provides a more comprehensive framework for evaluating unlearning techniques.

\bibliography{main}

%%%%%%%%%%%%%%%%%%%%%%%%%%%%%%%%%%%%%%%%%%%%%%%%%%%%%%%%%%%%%%%%%%%%%%%%%%%%%%%
%%%%%%%%%%%%%%%%%%%%%%%%%%%%%%%%%%%%%%%%%%%%%%%%%%%%%%%%%%%%%%%%%%%%%%%%%%%%%%%
% APPENDIX
%%%%%%%%%%%%%%%%%%%%%%%%%%%%%%%%%%%%%%%%%%%%%%%%%%%%%%%%%%%%%%%%%%%%%%%%%%%%%%%
%%%%%%%%%%%%%%%%%%%%%%%%%%%%%%%%%%%%%%%%%%%%%%%%%%%%%%%%%%%%%%%%%%%%%%%%%%%%%%%
\newpage
\appendix

\section{Random Data Forgetting} \label{app:random_data_forgetting}
Another scenario in machine unlearning (MU) is \textit{random data forgetting}, which involves forgetting a randomly selected subset of data across multiple classes. This differs from the \textit{class-wise forgetting} task, which aims to forget entire data from single or multiple classes.

\subsection{{Information Difference Index} for 
Random Data Forgetting}\label{app:idi_random}

To calculate the information difference index (IDI) for class-wise forgetting, we employ a binary label \(Y\) to determine whether a sample belongs to the retain or forget set. 
However, this approach is inadequate for random data forgetting, where samples span multiple classes and a minor fraction of each class is targeted for forgetting. 
As a result, no single class is completely removed.
To address this, we transform the binary label \(Y\) into a multiclass label \(Y_C\), which reflects the ground-truth class label of each sample. 
Consequently, we define the IDI for random data forgetting as follows:
\begin{equation}
    \mathbf{IDI}_{random}(\mathbf{\theta_u}) 
    = \frac{\mathbf{ID}_{random}(\mathbf{\theta_u})}{\mathbf{ID}_{random}(\mathbf{\theta_0})},
\end{equation}
where $\mathbf{ID}_{random}(\mathbf{\theta_u}) = \sum_{\ell=1}^{L} \bigl( I(\mathbf{Z}^{(\mathbf{u})}_\ell; Y_C) - I(\mathbf{Z}^{(\mathbf{r})}_\ell; Y_C) \bigr)$. 
Unlike the ID computed in a class-wise forgetting, \(\mathbf{ID}_{random}(\cdot)\) utilizes only the forget set \(\mathcal{D}_f\). 
Intuitively, we expect the mutual information \(I(\mathbf{Z}^{(\mathbf{o})}; Y_C)\) to be higher than \(I(\mathbf{Z}^{(\mathbf{r})}; Y_C)\) because Original explicitly learned the relationship between the forget samples and their ground truth labels, while Retrain did not. 
Although the labels have transitioned from binary to multiple classes, the function \(f_{\nu_\ell}\) remains unchanged. 
For \(g_{\eta_\ell}\), it now employs the \(C\) dimension of vectors, where \(C\) represents the total number of data classes.\\

\subsection{COLA+}
The core idea behind COLA is to induce catastrophic forgetting within the model's encoder in the collapse phase, making the influence of the forget set vanish implicitly. 
This approach is effective for class-wise forgetting tasks, where the forget set includes distinct classes. 
However, it may be less effective for random data forgetting, where the forget set and retain set samples generally share the same classes and are not easily distinguishable.
To address this, we aim to explicitly remove the information of the forget set through pseudo-labeling. 
This variant, called COLA+, assigns the second-highest predicted label to the forget set samples before unlearning with supervised contrastive loss~\citep{khosla2020supervised}. 
This pseudo-labeling effectively collapsing the forget set features into the retain set clusters of other classes, while reducing the confusion of the knowledge of the retain set.
The results of the COLA+ experiment on the random data forgetting task are presented in Appendix~\ref{app:random_data_forgetting_results}.

\section{Network Parametrizations for InfoNCE}\label{app:parameterization} 
This section provides a detailed explanation of the parameterization of the neural network critic functions used in the InfoNCE loss, including layer-specific adaptations.
\subsection{Critic Functions for InfoNCE Loss}
To compute the InfoNCE loss, we parameterize two critic functions: \( f_{\nu} \) and \( g_{\eta} \), where \( \nu \) and \( \eta \) represent the learnable parameters of their respective neural networks. For each layer \( \ell \) in the network, these functions are defined as follows:
\begin{enumerate}
    \item \textbf{Critic \( f_{\nu_\ell} \):}
    \begin{itemize}
        \item \( f_{\nu_\ell}: \mathcal{Z}_\ell \to \mathbb{R}^d \), where \( \mathcal{Z}_\ell \) represents the feature space at layer \( \ell \).
        \item This function maps raw or intermediate features \( \mathbf{Z}_\ell \) to a \( d \)-dimensional embedding space. In earlier layers, \( f_{\nu_\ell} \) must process raw, less interpretable features, making it more complex. For later layers, where features are more structured, \( f_{\nu_\ell} \) can leverage the refined representations for better alignment with \( Y \).
    \end{itemize}
\item \textbf{Critic \( g_{\eta_\ell} \):}
    \begin{itemize}
   \item \( g_{\eta_\ell}: \{0, 1\} \to \mathbb{R}^d \).
   \item For the binary variable \( Y \), \( g_{\eta_\ell} \) is parameterized as a pair of trainable \( d \)-dimensional vectors: \( g_{\eta_\ell}(0) \) and \( g_{\eta_\ell}(1) \). Depending on the label \( Y \), the corresponding vector is selected to represent the target embedding for contrastive learning.
    \end{itemize}
\end{enumerate}
\subsection{Layer-Specific Parameterization}
Each layer \( \ell \) has independent sets of parameters \( \nu_\ell \) and \( \eta_\ell \). This design allows the model to adapt to the varying complexity of feature representations across the network. Specifically:
\begin{itemize}
\item  In earlier layers, \( f_{\nu_\ell} \) focuses on extracting information from raw features \( \mathbf{Z}_\ell \), which are less structured and more challenging to interpret.
\item  In later layers, \( f_{\nu_\ell} \) benefits from more refined features, enabling a more direct alignment with \( Y \).
\end{itemize}

\section{Experiment Details}
\label{app:experiment_detail}

\subsection{Detailed explanation for Head Distillation} \label{app:detailed_head_distillation}
\begin{figure}[htbp] % [h]here, [t]top, [b]bottom, [p]page
    \centering
    \includegraphics[width=0.7\linewidth]{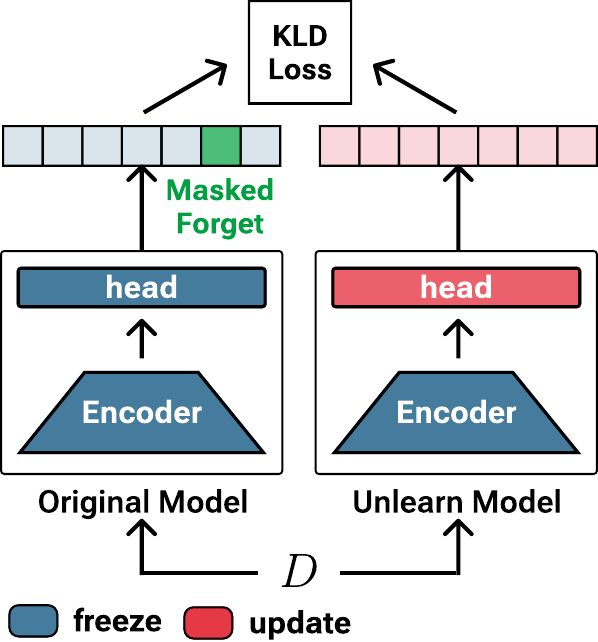}
    \caption{Overview of head distillation (HD): Original distills knowledge into the unlearn model's head by masking the forgetting class logit, with the encoder kept frozen.}
    \label{fig:illust__hd}
\end{figure}
Head Distillation (HD) employs logit distillation from the original model \(\mathbf{\theta_o}\) for efficient machine unlearning. The unlearned model \(\mathbf{\theta_u}\) is initialized from \(\mathbf{\theta_o}\) with the encoder frozen and only the head remaining trainable. During the unlearning process, the head is finetuned on training dataset \(\mathcal{D}\) using KL-divergence loss to follow the masked output from \(\mathbf{\theta_o}\), where the logit for the forgetting class is set to negative infinity while preserving the logits for the remaining classes. The illustration of HD is shown in~\Cref{fig:illust__hd}. This enables \(\mathbf{\theta_u}\) to mimic a pseudo-retrained model, as the masked logits closely resemble those of Retrain.

In random data forgetting scenarios where forget samples can come from any class, we modify the HD approach while maintaining its core principle of modifying only the last layer. Instead of using logit masking, we apply gradient descent on the retain set and gradient ascent on the forget set while training only the model's head. Our experimental results demonstrate that HD achieves strong performance on black-box metrics (accuracy, MIA, and JSD) within seconds, successfully deceiving these metrics while maintaining computational efficiency (\Cref{fig:exp__hd_rte_ta_larger,fig:exp__hd_rte_taM5_larger,fig:exp__hd_rte_taM20_larger,fig:exp__hd_rte_taS_larger}).

\subsection{Evaluation Metrics Detail} \label{app:evaluation_metrics_detail}
\paragraph{UA, RA, TA}
We compute accuracy as follows:
\begin{equation}
\mathbf{\text{Acc}_\mathcal{D}}(\mathbf{\theta}) 
= \frac{1}{|\mathcal{D}|} \sum_{(x, y) \in \mathcal{D}} \mathbf{1}\left[\arg\max\left(f(x;\mathbf{\theta})\right) = y\right],
\end{equation}
where \(f(x; \mathbf{\theta})\) represents the model's output logits for input \(x\) with parameters \(\mathbf{\theta}\), and \(y\) is the ground truth label.
Unlearning accuracy (UA), which quantifies the model's task performance on forgetting data, is defined as \(UA(\mathbf{\theta_u}) = 1 - \text{Acc}_{\mathcal{D}_f}(\mathbf{\theta_u})\). 
Remaining accuracy (RA) measures the model’s performance on the retain set \( \mathcal{D}_r \), which should be preserved after unlearning, and is defined as \(RA(\mathbf{\theta_u}) = \text{Acc}_{\mathcal{D}_r}(\mathbf{\theta_u})\).
Finally, testing accuracy (TA) evaluates generalization to unseen data, and is defined as \(TA(\mathbf{\theta_u}) = \text{Acc}_{\mathcal{D}_{test}}(\mathbf{\theta_u})\).
It is important to note that better unlearning in terms of accuracy reflects a smaller performance gap between the unlearned model and Retrain, meaning that higher accuracy levels are not necessarily better.
Refer to~\citep{liu2024l1sparse} for detailed explanation.

\paragraph{MIA.}
Membership Inference Attack (MIA)~\citep{shokri2017mia1, carlini2022mia2} determines whether a specific data record was part of a model's training set by leveraging auxiliary classifiers to distinguish between training and non-training data based on the model's output.

In the context of unlearning, membership inference attack (MIA) is primarily used as an evaluation metric, rather than representing an adversarial scenario where an attacker attempts to extract membership information from the unlearned model.
Consequently, a comparable MIA success rate on the forgetting data relative to Retrain signifies a more effective unlearning algorithm.
Unlike the original MIA implementation~\citep{shokri2017mia1}, which utilizes multiple shadow models, MIA variants in the unlearning often employ a single auxiliary classifier for each unlearning method~\citep{liu2024l1sparse}. 
A detailed comparison of these approaches can be found in~\citep{hayes2024inexact}.

The MIA implementation in our study has two phases: the \textit{training phase} and the \textit{testing phase}.

During the \textit{training phase}, we create a balanced dataset by equally sampling from the retain set ($\mathcal{D}_r$) and the test set, explicitly excluding the forget set ($\mathcal{D}_f$). 
We then use this balanced dataset to train the MIA predictor with two output categories (train, non-train), 
allowing it to differentiate between training and non-training samples.

In the \textit{testing phase}, the trained MIA predictor is used to evaluate the efficacy of the unlearning methods. 
Specifically, the \textbf{MIA} metric is calculated by applying the MIA predictor to the unlearned model ($\theta_u$) using the forget set ($\mathcal{D}_f$). 
The objective is to determine how many samples within $\mathcal{D}_f$ are identified as training samples by the MIA predictor.

Formally, MIA is defined as:

\begin{equation}
    \text{MIA} = 1 - \frac{\text{TN}}{|\mathcal{D}_f|}
\end{equation}

where \textit{TN} represents the number of true negatives (\ie, the number of forget samples correctly predicted as non-training examples by the MIA predictor), 
and $|\mathcal{D}_f|$ denotes the total number of samples in the forget set. 
Overall, MIA leverages privacy attack mechanisms to validate the effectiveness of the unlearning process,
providing a quantitative measure of how successfully the model has `forgotten' specific data resembling Retrain.

We consider two widely adopted variants of MIA.
The first variant, \textbf{C-MIA (Confidence-based MIA)}, assesses membership based on the confidence score, which is the predicted probability of the true class~\citep{fan2023salun, liu2024l1sparse}.
The second variant, \textbf{E-MIA (Entropy-based MIA)}, infers membership by examining the entropy of the model's outputs, 
calculated as $H(x) = -\sum_i \mathbf{p}_i(x) \cdot \log {\mathbf{p}_i(x)}$ \blue{~\citep{chundawat2023badt, foster2024ssd, kurmanji2024scrub}.}
Higher entropy indicates greater uncertainty in the model's predictions, 
often signaling non-training samples. 
We primarily report results using E-MIA due to its more pronounced differences across various baselines compared to C-MIA.
It is noteworthy that our head distillation (HD) method achieves similar performance outcomes with both E-MIA and C-MIA.

\paragraph{U-LiRA}
U-LiRA is a variant of black-box membership inference attack (MIA) designed to evaluate the privacy protection of unlearning algorithms~\citep{hayes2024inexact}. 
For our LiRA MIA experiments in~\cref{app:generalization_hd}, we followed the U-LiRA methodology from~\citet{hayes2024inexact}, training 128 ResNet-18 models on random splits of half the CIFAR-10 training set, ensuring that each sample is included in 64 and excluded from 64 models on average. 
We applied the unlearning algorithm to 40 random forget sets (200 samples each) per model, resulting in 5,120 unlearned models. 
For evaluation, we used 2,560 shadow and 2,560 target models, focusing on class 4 samples. 
Testing each method required 300–500 GPU hours, highlighting the cost-intensive nature of LiRA when adopting to unlearning; additional details can be found in~\citep{hayes2024inexact}.

\paragraph{JSD}
Jensen-Shannon divergence (JSD) is presented in Bad-T~\citep{chundawat2023badt}. It measures the distance between the output distributions of the unlearned model and Retrain. 
JSD is measured as follows:
\begin{equation}
\begin{split}
\mathbf{JSD}_\mathcal{D}(\mathbf{\theta_u}, \mathbf{\theta_r}) &= 0.5 \cdot KL(f(x;\theta_u)\mid\mid m) \\
&\quad + 0.5 \cdot KL(f(x;\theta_r)\mid\mid m),
\end{split}
\end{equation}
where $KL(\cdot)$ is Kullback-Leibler divergence, $x$ is data from \(\mathcal{D}\), and m = $\frac{f(x;\mathbf{\theta_u}) + f(x;\mathbf{\theta_r})}{2}$.
Here, $f(x; \mathbf{\theta})$ represents the model's output probability distribution for input $x$ with parameters \(\mathbf{\theta}\).
A smaller distance means better unlearning as the unlearned model better mimics Retrain.

\paragraph{RTE}
Runtime efficiency (RTE) measures the time that an algorithm spends to complete the unlearning, where smaller RTE indicates more efficient unlearning~\citep{fan2023salun, liu2024l1sparse, foster2024ssd}. Since it measures the experiment wall-clock time, it has high variance depending on the experiment environment. 

% \paragraph{AIN, RT}
% Time-based metrics, such as the Anamnesis Index (AIN)~\cite{chundawat2023ain} and Relearn Time (RT)~\cite{tarun2023unsir, golatkar2020eternal, golatkar2020forgetting},
% measure the time (epochs) required for a model to regain a specified performance level on forgotten data. The original paper~\cite{chundawat2023ain} defined AIN as follows:
% \[
% \mathbf{\text{AIN}} = \frac{r_t(\theta_o, \theta_u, \alpha)}{r_t(\theta_r, \theta_u, \alpha)},
% \]
% where $r_t(\cdot)$ is the number of iterations count required to reach the accuracy threshold $\alpha$.

\subsection{Approximate MU Baselines.}
\label{app:mu_baselines}
We conduct our experiments on several widely used or recent approximate MU baselines: Finetuning (\textbf{FT})~\citep{golatkar2020eternal} finetunes Original \( \mathbf{\theta_o}\) with retain set \( \mathcal{D}_r \), inducing catastrophic forgetting~\citep{french1999catastrophic_1,kirkpatrick2017catastrophic_2} of \( \mathcal{D}_f \). 
Random labeling (\textbf{RL})~\citep{golatkar2020eternal} involves finetuning \(\mathbf{\theta_o}\) with randomly labeled forget set \( \mathcal{D}_f \).  
Gradient ascent (\textbf{GA})~\citep{thudi2022unrolling} trains \( \mathbf{\theta_o}\) with reverse gradient steps using \( \mathcal{D}_f \).
\textbf{Bad-T}~\citep{chundawat2023badt} uses a teacher-student framework that utilizes distillation techniques,
distinguishing between beneficial and detrimental influences through good and bad teachers to refine the learning process.
Catastrophic forgetting-k (\textbf{CF-k}) and exact unlearning-k (\textbf{EU-k})~\citep{goel2022evaluating} 
involve either finetuning (CF-k) or retraining (EU-k) the last \( k \) layers of the model using \( \mathcal{D}_r \) while freezing the prior layers.
\textbf{SCRUB}~\citep{kurmanji2024scrub} employs a technique of positive distillation from \(\mathbf{\theta_o}\) using the \( \mathcal{D}_r \),
and negative distillation on the \( \mathcal{D}_f \), which helps in selectively retaining beneficial knowledge while discarding the unwanted influences.
\textbf{\(\mathbf{\ell_1}\)-sparse}~\citep{liu2024l1sparse} enhances the model's ability to forget by strategically inducing weight sparsity in \(\mathbf{\theta_o}\).
\textbf{SALUN}~\citep{fan2023salun} finetunes the salient weights of \( \mathbf{\theta_o}\) using a method that incorporates random labeling.
For all experiments, we conduct five independent runs.
% \blue{
% \textbf{BoundaryShrink}~\citep{chen2023boundary} reassigns the \( \mathcal{D}_f \) to their nearest but incorrect labels, splitting the decision space of the forgetting class.  
% \textbf{BoundaryExpand}~\citep{chen2023boundary} maps \( \mathcal{D}_f \) to an extra shadow class, bypassing the need to find nearest labels.}
% \textbf{BoundrayShi}

\subsection{Datasets and Models}
% We conduct experiments using widely adopted datasets -- CIFAR-10, CIFAR-100~\cite{krizhevsky2009cifar}, and ImageNet-1K~\cite{DenDon09Imagenet} -- and models, ResNet-18, ResNet-50~\cite{he2016resnet}, and ViT~\cite{dosovitskiy2020vit}, in the image classification task. 
% CIFAR-10 and CIFAR-100 consist of 10 and 100 classes, respectively, with 60,000 images of size 32 x 32. 
% ImageNet-1K contains 1,000 classes with a total of 1,281,167 training images. 
% We resize ImageNet images to 224 x 224 for all experiments.
% Similarly, we resize CIFAR images to 224 x 224 for ViT experiments. 
% We apply simple data augmentation techniques, including random cropping and random horizontal flipping, in all training phases, including the pretraining and the unlearning.

We conduct image classification experiments utilizing well-established datasets and models. 
The datasets include CIFAR-10, CIFAR-100 \citep{krizhevsky2009cifar}, and ImageNet-1K \citep{DenDon09Imagenet}; and the models are ResNet-18, ResNet-50 \citep{he2016resnet}, and Vision Transformer (ViT) \citep{dosovitskiy2020vit}. 
CIFAR-10 and CIFAR-100 each comprise 50,000 training images distributed across 10 and 100 classes, respectively, each with an original resolution of 32 x 32 pixels. 
In our experiments, we resize the images in ImageNet-1K, which consists of 1,281,167 training images across 1,000 classes, to 224 x 224 pixels. 
Similarly, for the ViT experiments, we resize CIFAR images to 224 x 224 pixels to accommodate the architecture's requirements. 
Throughout the training process, including pretraining and unlearning phases, we employ basic data augmentation techniques such as random cropping and random horizontal flipping.

\subsection{Pretraining Settings}
To perform unlearning, we require two models: \textbf{Original}, trained on the entire dataset \(\mathcal{D}\), and \textbf{Retrain}, trained on the retain set \(\mathcal{D}_r\). 
Original initializes the unlearning model. 
After unlearning, Retrain evaluates them.
Table~\ref{tab:pretraining_detail} summarizes the training configurations for each dataset and model combination. We train ResNet models from scratch and initialize ViT models with ImageNet-21K pretrained weights. For training on ImageNet-1K, we follow the configurations provided by Pytorch\footnote{https://github.com/pytorch/examples/tree/main/imagenet}.

\begin{table*}[t!]
    \centering
    \begin{tabular}{ccccc}
        \toprule[1pt]
        \midrule
        \multirow{2}{*}{Settings} & \multicolumn{2}{c}{CIFAR-10 / CIFAR-100}&  \multicolumn{2}{c}{ImageNet-1K} \\
        \cmidrule(lr){2-5}
         & Resnet-18 / Resnet-50 & ViT & ResNet-50 & ViT \\
        \midrule
        Epochs & 300 & 3 & 90 & 30 \\
        \midrule
        Batch Size & \multicolumn{2}{c}{128} & 256 & 512 \\
        \midrule
        LR & 0.1 & 0.00002 & 0.1 & 0.02 \\
        \midrule
        Optimizer & \multicolumn{4}{c}{SGD} \\
        \midrule
        Momentum & \multicolumn{4}{c}{0.9} \\
        \midrule
        L2 regularization & \multicolumn{2}{c}{0.0005} & \multicolumn{2}{c}{0} \\
        \midrule
        Scheduler & \multicolumn{4}{c}{CosineAnnealing} \\
        \midrule
        \bottomrule[1pt]
    \end{tabular}
\caption{Training configuration for Original and Retrain.}
\label{tab:pretraining_detail}
\end{table*}

\subsection{Unlearning Settings}
We aim to follow the hyperparameters provided by the original papers. 
However, many hyperparameters are missing since most existing works do not experiment with large-scale datasets and models. 
Additionally, some values from the original papers result in poor performance, likely due to different experiment settings, as most previous work performed unlearning without any data augmentation, unlike our experiments. 
Therefore, we conduct thorough hyperparameter searches for each baseline.
Optimal hyperparameters were identified through a fine-grained search across the range $[10^{-5}, 10^{-2}]$, evaluating values at linear steps within each decade.
The detailed hyperparameters of each baseline, including our method COLA and COLA+, are shown in Table~\ref{tab:hyperparam_class} and Table~\ref{tab:hyperparam_sample}.
We use the same optimizer and batch size from the original papers and focus on finding the best epoch number and learning rate in terms of unlearning accuracy (UA) and testing accuracy (TA). 
Note that we implement gradient ascent (GA) from SCRUB~\citep{kurmanji2024scrub} (referred to as 'NegGrad+') due to its strong performance.

\subsection{COLA and COLA+ Pseudo Code}\label{app:cola_pseudocode}
\begin{figure}[htbp]
    \centering
    \includegraphics[width=\linewidth]{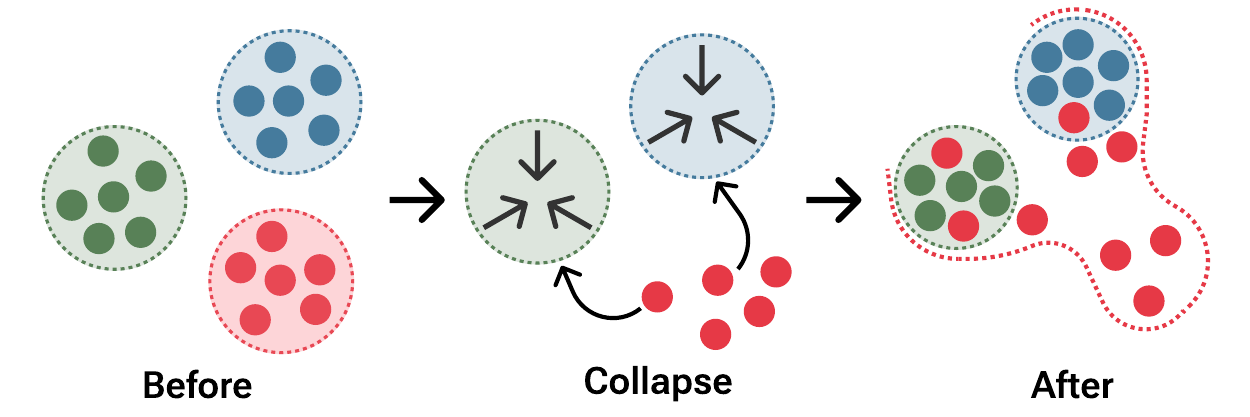}
    \caption{Illustration of the collapse phase of COLA. Features (post-encoder, pre-head) from forget set \(\mathcal{D}_f\) are represented in red, while features from retain set \(\mathcal{D}_r\) are represented in green and blue. 
    The figure shows a class-wise forgetting task. Best viewed in color.}
    \label{fig:COLA_illust_figure}
\end{figure}
Algorithm~\ref{algo:cola} shows the pseudo code of our two-step framework COLA. Only using the retain set $\mathcal{D}_r$, in the \textit{collapse phase} (see Figure~\ref{fig:COLA_illust_figure}), We first train the encoder of the model using supervised contrastive loss~\citep{khosla2020supervised} as follows:
\begin{equation}
\begin{split}
\text{SupConLoss}(b, \mathbf{\theta_{enc}}) &= \frac{1}{|b|}\sum_{i} \frac{1}{|P(i)|} \sum_{p \in P(i)} \\
&\quad -\log \frac{\exp(\mathbf{z}_i \cdot \mathbf{z}_p / \tau)}{\sum_{a \in A(i)} \exp(\mathbf{z}_i \cdot \mathbf{z}_a / \tau)},
\end{split}
\label{eq:supconloss}
\end{equation}

where $P(i)$ is the set of indices of positive samples sharing the same label as sample \(i\), $A(i)$ is the set of all indices excluding sample \(i\), $\tau$ is a temperature, and ${z}_i = F(x_i;\mathbf{\theta_{enc}})$, the output feature of the model encoder. 
Then we train the whole network using cross-entropy loss in the \textit{align phase}.
COLA+ additionally utilizes forget set $\mathcal{D}_f$ in the collapse phase, where the label of forget samples is changed to the class label closest to the original label, determined by the logit output of the head of Original. 
Its pseudo code is presented in Algorithm~\ref{algo:cola+}.

% \input{figures/cola_algorithm}
% \input{figures/cola_plus_algorithm}
% Preamble
% \usepackage{algorithm}
% \usepackage{algorithmic}
% \usepackage{float} % for [H]

% In the document
\begin{figure*}[t]
\centering

% --- LEFT: COLA ---
\begin{minipage}[t]{0.49\textwidth}
\begin{algorithm}[H]
\caption{Pseudo Code of \textit{COLA}}
\label{algo:cola}
\begin{algorithmic}
    \REQUIRE learning rate $\eta$, number of epochs $E_1, E_2$, retain set
    $\mathcal{D}_r = \{(x_i, y_i) \mid (x_i, y_i) \in \mathcal{D}_r\}$, encoder
    $F(\cdot;\mathbf{\theta})$, and model weight
    $\mathbf{\theta} = \{\mathbf{\theta_{enc}}, \mathbf{\theta_{head}}\}$
    \vspace{1em}
    \STATE $\mathbf{\theta_{u,enc}} \gets \mathbf{\theta_{o,enc}}$ \hfill $\triangleright$ Collapse phase
    \FOR{$e \gets 0$ \textbf{to} $E_1 - 1$}
        \FOR{all batches $b$ of $\mathcal{D}_r$}
            \STATE $L = \text{SupConLoss}(b, \mathbf{\theta_{u,enc}})$ \hfill $\triangleright$ Equation~\ref{eq:supconloss}
            \STATE $\mathbf{\theta_{u,enc}} \gets \mathbf{\theta_{u,enc}} - \eta \nabla_{\mathbf{\theta_{u,enc}}} L$
        \ENDFOR
    \ENDFOR
    \vspace{1em}
    \STATE $\mathbf{\theta_{u,head}} \gets \text{random initialization}$ \hfill $\triangleright$ Align phase
    \FOR{$e \gets 0$ \textbf{to} $E_2 - 1$}
        \FOR{all batches $b$ of $\mathcal{D}_r$}
            \STATE $\mathbf{\theta_{u}} \gets \mathbf{\theta_{u}} - \eta \nabla_{\mathbf{\theta_{u}}} L_{CE}$
        \ENDFOR
    \ENDFOR \\
    \textbf{return} $\mathbf{\theta_u} = \{\mathbf{\theta_{u,enc}}, \mathbf{\theta_{u,head}}\}$
\end{algorithmic}
\end{algorithm}
\end{minipage}
\hfill
% --- RIGHT: COLA+ ---
\begin{minipage}[t]{0.49\textwidth}
\begin{algorithm}[H]
\caption{Pseudo Code of \textit{COLA+}}
\label{algo:cola+} % kept exactly as in your original file
\begin{algorithmic}
    \REQUIRE learning rate $\eta$, number of epochs $E_1, E_2$, retain set
    $\mathcal{D}_r = \{(x_i, y_i) \mid (x_i, y_i) \in \mathcal{D}_r\}$, forget set
    $\mathcal{D}_f = \{(x'_i, y'_i) \mid (x'_i, y'_i) \in \mathcal{D}_f\}$,
    encoder $F(\cdot;\mathbf{\theta})$, head $G(\cdot;\mathbf{\theta})$, and model weight
    $\mathbf{\theta} = \{\mathbf{\theta_{enc}}, \mathbf{\theta_{head}}\}$
    \vspace{1em}
    \STATE $\mathbf{\theta_{u,enc}} \gets \mathbf{\theta_{o,enc}}$ \hfill $\triangleright$ Collapse phase
    \STATE $\mathbf{\theta_{u,head}} \gets \mathbf{\theta_{o,head}}$
    \FOR{$e \gets 0$ \textbf{to} $E_1 - 1$}
        \FOR{$\{b_r, b_f\}$ in all batches of $\{\mathcal{D}_r, \mathcal{D}_f\}$}
            \FOR{$x'_i \in b_f$}
                \STATE $y'_i \gets \argmax_y \text{softmax}(G(F(x'_i;\mathbf{\theta_{u,enc}});\mathbf{\theta_{u,head}})) \cdot \mathbb{I}[y \ne y'_i]$ \hfill $\triangleright$ Pseudo-labeling
            \ENDFOR
            \STATE $b \gets b_r + b_f$
            \STATE $L = \text{SupConLoss}(b, \mathbf{\theta_{u,enc}})$ \hfill $\triangleright$ Equation~\ref{eq:supconloss}
            \STATE $\mathbf{\theta_{u,enc}} \gets \mathbf{\theta_{u,enc}} - \eta \nabla_{\mathbf{\theta_{u,enc}}} L$
        \ENDFOR
    \ENDFOR
    \STATE \hfill $\triangleright$ Align phase
    \FOR{$e \gets 0$ \textbf{to} $E_2 - 1$}
        \FOR{all batches $b$ of $\mathcal{D}_r$}
            \STATE $\mathbf{\theta_{u}} \gets \mathbf{\theta_{u}} - \eta \nabla_{\mathbf{\theta_{u}}} L_{CE}$
        \ENDFOR
    \ENDFOR\\
    \textbf{return} $\mathbf{\theta_u} = \{\mathbf{\theta_{u,enc}}, \mathbf{\theta_{u,head}}\}$
\end{algorithmic}
\end{algorithm}
\end{minipage}

\end{figure*}

\subsection{IDI Details}
\label{app: IDI implementation}
To derive IDI from features, it is necessary to train the critic functions \(f_{\nu_\ell}\) and \(g_{\eta_\ell}\), as referenced in Section~\ref{subsection_MI_estimation}.
For the training of \(g_{\eta_\ell}\), a learning rate of $5\cdot10^{-4}$ is applied in all architectures and datasets. 
Meanwhile, for \(f_{\nu_\ell}\), the learning rates are set at $2\cdot10^{-5}$ for CIFAR10 ResNet-18, $2\cdot10^{-6}$ for ViT ImageNet-1K, and $1\cdot10^{-5}$ for the remaining architectures of the data set.

To get IDI, we analyzed the outputs from the layers of different models. 
Specifically, we evaluated the last two block outputs for ResNet18 and the final three for ResNet50. 
For Vision Transformer (ViT), we examined the outputs of the final three transformer encoder blocks.
Note that these selections of layers is based on the observation that the information differences of outputs from the initial layers of both original and retrained models are similar. 
For empirical justifications of these selections, please refer to~\cref{Effect_of_number_of_Layers_for_IDI}.

\subsection{System specification}
For fair comparison, all experiments are executed in Python 3.10, on an Ubuntu 18.04 machine with 72 CPU cores, 4 Nvidia RTX A6000 GPUs and 512GB memory.

\section{Additional Unlearning Results}
\label{app:additional_results}
In this section, we provide the full experiment results on various machine unlearning settings, extending the results in~\cref{section_residualinformation,} and~\cref{section_evaluation}.

\subsection{
Head Distillation (HD) Results}

\label{app:generalization_hd}

By achieving strong black-box performance while modifying only the model head and retaining intermediate layer information (\ie, preserving the information of the forget samples), we highlighted the limitations of black-box metrics in~\cref{section_residualinformation}. 
To examine whether HD consistently exposes these limitations across diverse metrics (e.g., Accuracy, MIA, JSD, ZRF~\citep{chundawat2023badt}, AUS~\citep{cotogni2023duck}, LiRA MIA~\citep{carlini2022mia2}) and scenarios (single-class, multi-class, and random data forgetting), we extend our analysis to include a broader range of evaluations. 
We compare HD with five other methods: FT, RL, GA, $\ell_1$-sparse, and SALUN.

\begin{figure*}[t!]
    \vskip 2em
    \centering
    \includegraphics[width=\linewidth]{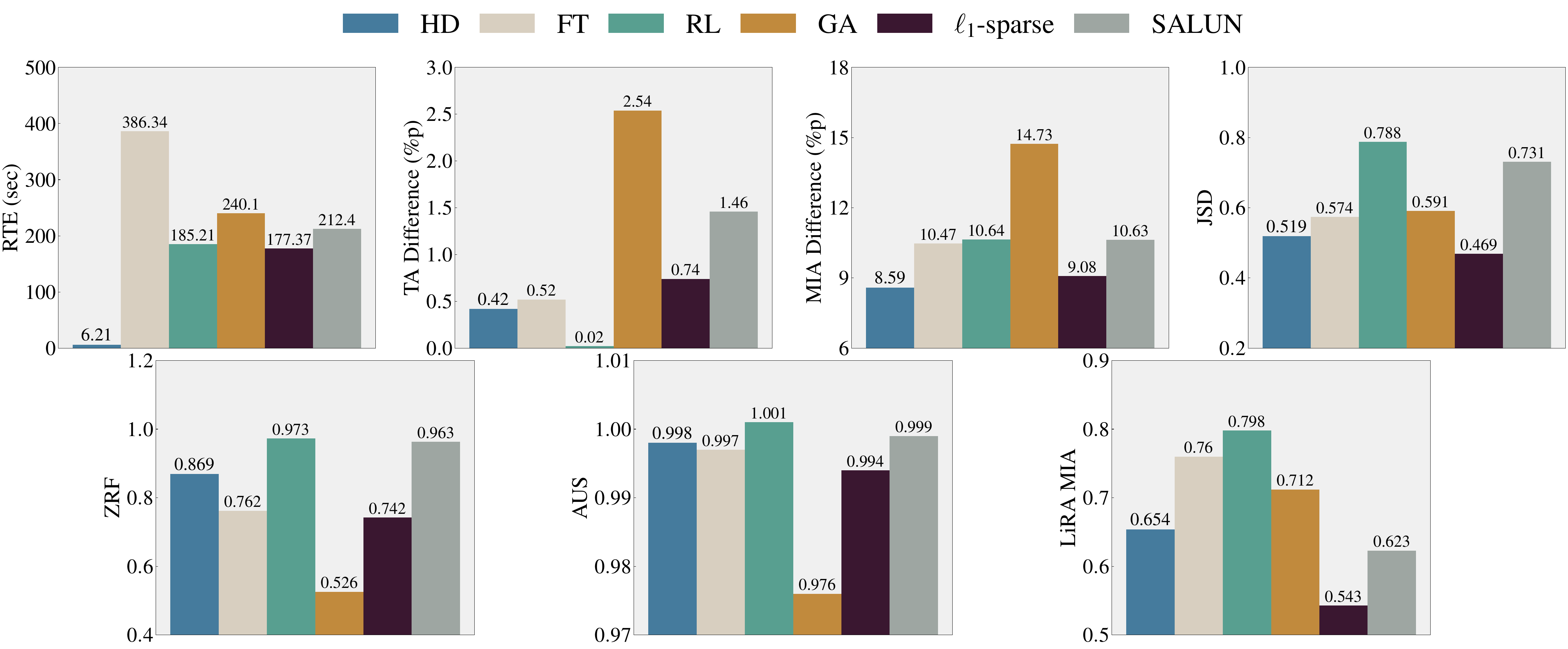}
    \centering
    \caption{
   Performance of six methods (HD, FT, RL, GA, $\ell_1$-sparse, SALUN) on (CIFAR-10, ResNet-18), evaluated in efficiency (RTE), accuracy (TA), and efficacy (MIA, JSD, ZRF, AUS, LiRA MIA) in single-class forgetting scenarios. Lower differences from Retrain in TA, MIA, and JSD indicate closer similarity to Retrain, while higher values for ZRF and AUS represent better efficacy. Additionally, LiRA MIA values closer to 0.5 reflect higher efficacy.
   }
    \label{fig:exp__hd_rte_ta_larger}
    \vskip 2em
\end{figure*}

\paragraph{Single-Class Forgetting.}
\Cref{fig:exp__hd_rte_ta_larger} presents the results for single-class forgetting on CIFAR-10.
Despite modifying only the last layer, HD achieves the best MIA performance among the five methods. 
Additionally, it delivers competitive results across accuracy, JSD, ZRF, and AUS metrics, completing the task in under ten seconds (RTE). 
Notably, HD also performs comparably on LiRA MIA~\citep{hayes2024inexact}, one of a recently proposed black-box MIA metrics. 
The strong performance of HD across various black-box metrics underscores the need for robust white-box metrics to more effectively assess unlearning quality.

\begin{figure*}[t!]
    \centering
    \includegraphics[width=0.92\linewidth]{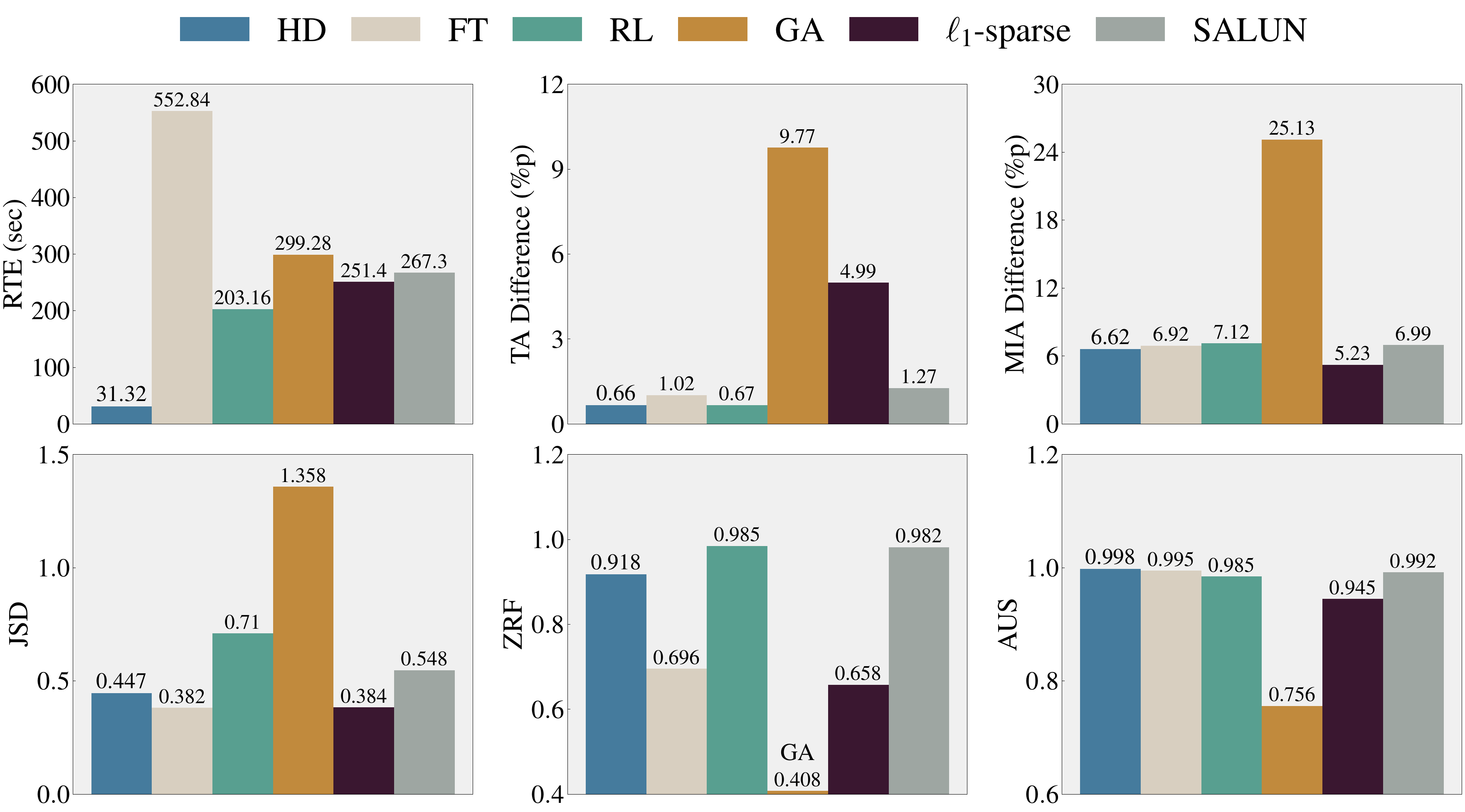}
    \centering

    \caption{
   Performance of six methods (HD, FT, RL, GA, $\ell_1$-sparse, SALUN) on (CIFAR-100, ResNet-18), evaluated in efficiency (RTE), accuracy (TA), and efficacy (MIA, JSD, ZRF, AUS) in multi-class forgetting scenarios (5 classes). 
   For TA, MIA, and JSD, lower differences from Retrain are preferred, indicating similarity to Retrain. For ZRF and AUS, higher values reflect better efficacy.
   }
    \label{fig:exp__hd_rte_taM5_larger}
\end{figure*}

\begin{figure*}[t!]
    \centering
    \includegraphics[width=0.92\linewidth]{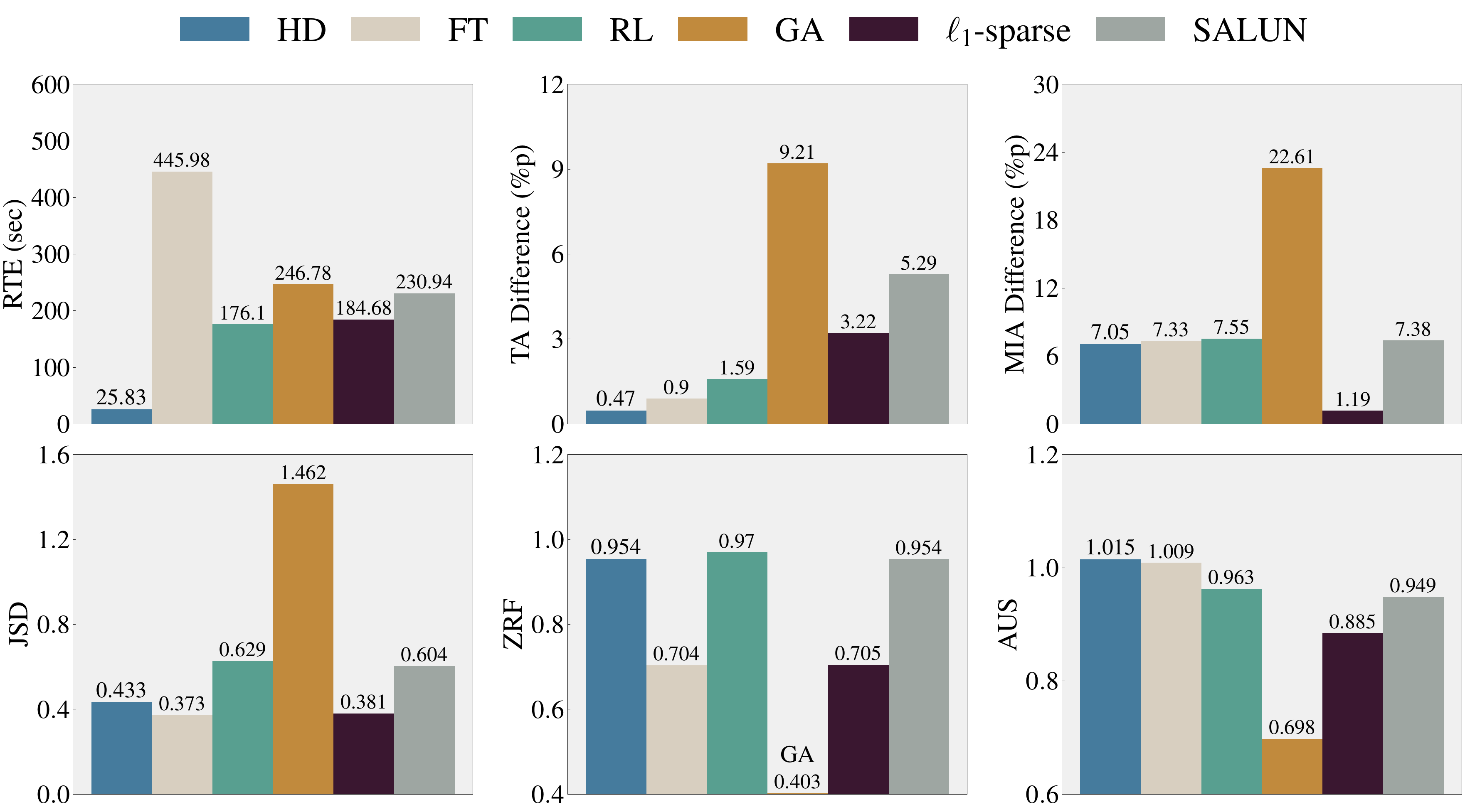}
    \centering
    \caption{
   Performance of six methods (HD, FT, RL, GA, $\ell_1$-sparse, SALUN) on (CIFAR-100, ResNet-18), evaluated in efficiency (RTE), accuracy (TA), and efficacy (MIA, JSD, ZRF, AUS) in multi-class forgetting scenarios (20 classes). 
   For TA, MIA, and JSD, lower differences from Retrain are preferred, indicating similarity to Retrain. For ZRF and AUS, higher values reflect better efficacy.
   }
    \label{fig:exp__hd_rte_taM20_larger}
\end{figure*}

\paragraph{Multi-Class Forgetting.}
For the multi-class forgetting scenario, we extend the logit masking technique of HD, as described in~\cref{section_residualinformation}, by incorporating additional masking for multiple classes. 
As shown in~\cref{fig:exp__hd_rte_taM5_larger,fig:exp__hd_rte_taM20_larger}, HD achieves comparable performance across TA, MIA, JSD, ZRF, and AUS, completing the tasks within the shortest time frame (31.32 seconds for forgetting 5 classes and 25.83 seconds for forgetting 20 classes). 
Similar to the single-class forgetting scenario, HD's comparable performance in this extended setup further highlights the limitations of black-box metrics.

\begin{figure*}[t!]
    \centering
    \includegraphics[width=0.88\linewidth]{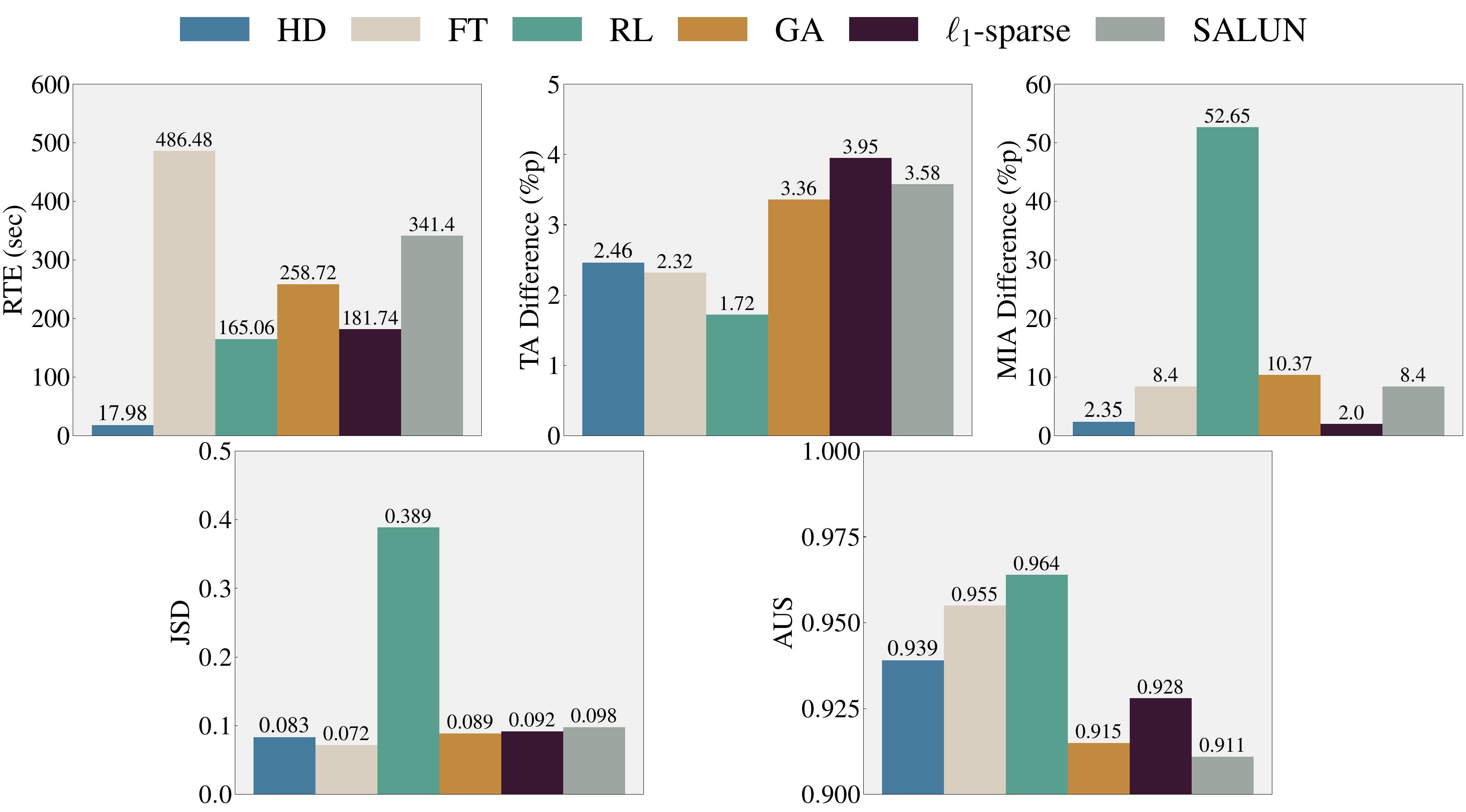}
    \centering
    \caption{
   Performance of six methods (HD, FT, RL, GA, $\ell_1$-sparse, SALUN) on (CIFAR-10, ResNet-18), evaluated in efficiency (RTE), accuracy (TA), and efficacy (MIA, JSD, AUS) in random data forgetting scenarios (500 samples per class). 
   For TA, MIA, and JSD, lower differences from Retrain are preferred, indicating similarity to Retrain. For AUS, higher values reflect better efficacy.
   }
    \label{fig:exp__hd_rte_taS_larger}
\end{figure*}

\paragraph{Random Data Forgetting.}
In random data forgetting scenarios, HD cannot be directly applied, as all classes are included in the retain set. 
To address this, we use gradient descent on the retain set and gradient ascent on the forget set while training only the model's head. 
While this approach differs from HD, which uses logit masking, we retain the name to emphasize its defining characteristic of modifying only the last layer.
As shown in~\cref{fig:exp__hd_rte_taS_larger}, HD achieves strong performance on black-box metrics (accuracy, MIA, and JSD) within less than 20 seconds (RTE).
This result demonstrates that HD can still deceive black-box metrics with comparable performance.
Note that ZRF is omitted in this scenario, as it is not ideally suited for random data forgetting; for more details, refer to~\citep{chundawat2023badt}.

\subsection{Standard Deviation} \label{app: C.2}
Due to the visual complexity of \cref{fig:mi__resnet18_resnet50_cifar10_class,fig:exp__ratio}, representing the standard deviation directly in these figures is challenging. 
Therefore, we include the standard deviation values separately in \cref{tab:resnet18_cifar10_std,tab:resnet50_cifar10_std} for \cref{fig:mi__resnet18_resnet50_cifar10_class}, and in \cref{tab:ratio_std} for \cref{fig:exp__ratio}.

\subsection{Single-Class Forgetting Results}

\paragraph{CIFAR-10 with Various Architectures}
Table~\ref{tab:single_class_cifar10} shows the full experiment results of the single-class forgetting experiments from Table~\ref{tab:main_baseline} on CIFAR-10 using different models. 
This extended table includes Jensen-Shannon divergence (JSD) and the unlearning results with ResNet-50 and ViT. 
Although many baselines show promising results on output based evaluation metrics, they generally exhibit poor feature-level unlearning.
In contrast, COLA not only outperforms existing baselines in IDI but also shows decent results in other metrics, demonstrating its effectiveness in removing the influence of the forget set within the encoder of the model.

\paragraph{CIFAR-100 with Various Architectures}
As demonstrated in Table~\ref{tab:single_class_cifar100}, we also compare COLA with other baselines on CIFAR-100. 
The results consistently highlight the difficulty of comparing and validating the efficacy of each unlearning method using existing output-based metrics. 
With the help of IDI, it is clear that COLA shows robustness in model unlearning on datasets with a large number of classes across various model architectures. 
Although SCRUB has achieved IDI near 0 for the CIFAR-10 ResNet-18 experiment, it shows significant variations in feature-level unlearning across different datasets and architectures. 

\subsection{Multi-Class Forgetting Results}\label{app:multi_class_forget}

\paragraph{Multi-Class Forgetting on CIFAR-10 and CIFAR-100}
Table~\ref{tab:multi_class_unlearning} presents the results of multi-class forgetting experiments on CIFAR-10 and CIFAR-100 using ResNet-18, which involves erasing the information of more than one class in the training set. 
We remove two classes from CIFAR-10 and five and twenty classes from CIFAR-100. 
Notably, many baselines exhibit higher IDI values as the number of forgetting class increases, demonstrating that the tendency to modify the head of the model strengthens with the difficulty of the unlearning tasks.
In contrast, COLA shows remarkable effectiveness, achieving metric values closely aligned with Retrain.
Specifically, COLA consistently achieves the lowest IDI values among the evaluated methods, indicating the necessity of the collapse phase for effective feature-level unlearning no matter the number of class to forget.

\paragraph{Multi-Class Forgetting on ImageNet-1K}
We conduct 5-class unlearning on ImageNet-1K using ResNet-50 and ViT.  
Table~\ref{tab:5_class_unlearning_imagenet} provides the complete results of~\cref{tab:main_baseline} for ImageNet-1K, including all evaluation metrics and outcomes on the ViT architecture.
However, it is important to note that IDI alone should not be used to assess unlearned models, as a low IDI might indicate a loss of overall information, including that from the retain set, which should be maintained at the same level as Original.
This issue is evident in the RA, TA, and IDI of EU-10 and CF-10 in Table~\ref{tab:5_class_unlearning_imagenet}. 
In contrast, COLA consistently achieves IDI near 0 while maintaining accuracy measurements comparable to Retrain, demonstrating the scalability of our framework to the large-scale datasets.

\subsection{Random Data Forgetting Results}
\label{app:random_data_forgetting_results}
Table~\ref{tab:random_sample_unlearning} presents the results of the random data forgetting task conducted on ResNet-18. 
For CIFAR-10 and CIFAR-100 datasets, we randomly selected 500 and 50 forget samples per class, respectively.
COLA+, which incorporates pseudo-labeling, successfully eliminates the influence of the forgetting data while maintaining competitive performance.
\section{Additional Discussions}\label{app:IDI and t-SNE}

\subsection{Mutual Information Curves} \label{app: D.1}
Figure~\ref{fig:mutual_information_curves} illustrates the estimated mutual information $I(\mathbf{Z}_{\ell};Y)$ of the features from the $\ell$-th layer $\mathbf{Z}_{\ell}$ and the binary label $Y$, computed by the InfoNCE loss across various architectures and datasets. 
We compute mutual information (MI) for all layers from the ResNet encoder and last five layers from the ViT encoder based single-class forgetting retain and forget sets. The upper bound of MI is given by the entropy $ H(Y) \geq I(\mathbf{Z}_{\ell};Y) = H(Y) - H(Y \mid \mathbf{Z}_{\ell})$. 
The estimated MI values fall within the range of the upper and lower bounds (0), validating the use of InfoNCE for MI estimation.
Notably, all MI curves consistently show a larger difference between Original and Retrain in the later layers of the encoder across various datasets and architectures, while differences are minimal in the earlier layers.
These observations underscore the validity of computing the information difference (ID) for the last few layers to quantify unlearning.
Furthermore, the difference between Original and Retrain becomes more significant with increasing numbers of forget classes, as shown in~\cref{fig:representation_mi_for_mutliclasses}.

\begin{table*}[t!]
% \captionsetup{
%   labelfont={color=blue}, % "표 1" 등의 레이블을 파란색으로
%   textfont={color=blue}   % 캡션 텍스트를 파란색으로
% }
    \centering

\caption{IDI values of methods on ResNet-18 with CIFAR-10 singleclass forgetting, computed using the last \(n\) selected layers, where \(n=1\) considers only the final representation, and larger \(n\) incrementally include earlier layers. $\star$ marks the IDI values reported in our work.
}
    \resizebox{0.75\textwidth}{!}{%
    \begin{tabular}{l|ccccc}
        \toprule
        Methods & Full Layers & $n=4$ & $n=3$ & $n=2^\star$ & $n=1$ \\ \midrule
        FT  & $0.670_{\pm 0.011}$  & $0.670_{\pm 0.011}$  & $0.670_{\pm 0.013}$  & $0.671_{\pm 0.008}$  & $0.673_{\pm 0.012}$ \\
        RL  & $0.833_{\pm 0.005}$  & $0.833_{\pm 0.004}$  & $0.830_{\pm 0.004}$  & $0.830_{\pm 0.005}$  & $0.837_{\pm 0.002}$ \\
        GA  & $0.338_{\pm 0.006}$  & $0.336_{\pm 0.007}$  & $0.334_{\pm 0.007}$  & $0.334_{\pm 0.014}$  & $0.333_{\pm 0.008}$ \\
        Bad-T  & $1.020_{\pm 0.023}$  & $1.016_{\pm 0.023}$  & $1.012_{\pm 0.023}$  & $1.014_{\pm 0.004}$  & $1.016_{\pm 0.022}$ \\
        EU-5  & $0.531_{\pm 0.004}$  & $0.531_{\pm 0.005}$  & $0.530_{\pm 0.006}$  & $0.528_{\pm 0.005}$  & $0.524_{\pm 0.007}$ \\
        CF-5  & $0.674_{\pm 0.023}$  & $0.675_{\pm 0.023}$  & $0.673_{\pm 0.024}$  & $0.675_{\pm 0.027}$  & $0.679_{\pm 0.026}$ \\
        EU-10  & $-0.352_{\pm 0.007}$  & $-0.347_{\pm 0.007}$  & $-0.344_{\pm 0.006}$  & $-0.349_{\pm 0.019}$  & $-0.310_{\pm 0.018}$ \\
        CF-10  & $-0.058_{\pm 0.010}$  & $-0.060_{\pm 0.010}$  & $-0.061_{\pm 0.010}$  & $-0.060_{\pm 0.017}$  & $-0.056_{\pm 0.008}$ \\
        SCRUB  & $-0.055_{\pm 0.028}$  & $-0.053_{\pm 0.029}$  & $-0.051_{\pm 0.027}$  & $-0.056_{\pm 0.008}$  & $-0.048_{\pm 0.026}$ \\
        SALUN  & $0.941_{\pm 0.029}$  & $0.937_{\pm 0.029}$  & $0.935_{\pm 0.029}$  & $0.936_{\pm 0.012}$  & $0.935_{\pm 0.027}$ \\
        l1-sparse  & $0.292_{\pm 0.011}$  & $0.293_{\pm 0.012}$  & $0.294_{\pm 0.011}$  & $0.293_{\pm 0.012}$  & $0.297_{\pm 0.013}$ \\
        COLA  & $0.010_{\pm 0.009}$  & $0.012_{\pm 0.009}$  & $0.012_{\pm 0.008}$  & $0.010_{\pm 0.006}$  & $0.015_{\pm 0.013}$ \\
        \bottomrule

\end{tabular}
}
\label{tab:diverse_idi_model}
\end{table*}
\begin{table*}[t!]
% \captionsetup{
  % labelfont={color=blue}, % "표 1" 등의 레이블을 파란색으로
  % textfont={color=blue}   % 캡션 텍스트를 파란색으로
% }
    \centering

\caption{IDI values of methods on ResNet-50 with CIFAR-10 single class forgetting, computed using the last \(n\) selected layers, where \(n=1\) considers only the final representation, and larger \(n\) incrementally include earlier layers. $\star$ marks the IDI values reported in our work.
}
    \resizebox{0.85\textwidth}{!}{%
    \begin{tabular}{l|cccccc}
        \toprule
        Methods & Full Layers & $n=5$ & $n=4$ & $n=3^\star$& $n=2$ & $n=1$ \\ \midrule
        FT  & $0.617_{\pm 0.006}$  & $0.618_{\pm 0.009}$  & $0.618_{\pm 0.013}$  & $0.607_{\pm 0.009}$  & $0.610_{\pm 0.013}$  & $0.563_{\pm 0.014}$ \\
        RL  & $0.808_{\pm 0.012}$  & $0.808_{\pm 0.012}$  & $0.811_{\pm 0.006}$  & $0.804_{\pm 0.006}$  & $0.814_{\pm 0.003}$  & $0.797_{\pm 0.000}$ \\
        GA  & $0.334_{\pm 0.018}$  & $0.338_{\pm 0.018}$  & $0.337_{\pm 0.018}$  & $0.334_{\pm 0.023}$  & $0.339_{\pm 0.017}$  & $0.269_{\pm 0.015}$ \\
        Bad-T  & $1.156_{\pm 0.016}$  & $1.151_{\pm 0.020}$  & $1.152_{\pm 0.021}$  & $1.153_{\pm 0.026}$  & $1.157_{\pm 0.018}$  & $1.163_{\pm 0.024}$ \\
        EU-5  & $1.044_{\pm 0.009}$  & $1.043_{\pm 0.008}$  & $1.050_{\pm 0.005}$  & $1.047_{\pm 0.005}$  & $1.061_{\pm 0.002}$  & $1.080_{\pm 0.002}$ \\
        CF-5  & $0.904_{\pm 0.005}$  & $0.906_{\pm 0.005}$  & $0.910_{\pm 0.006}$  & $0.906_{\pm 0.002}$  & $0.916_{\pm 0.001}$  & $0.914_{\pm 0.002}$ \\
        EU-10  & $0.760_{\pm 0.014}$  & $0.766_{\pm 0.011}$  & $0.766_{\pm 0.011}$  & $0.757_{\pm 0.011}$  & $0.756_{\pm 0.010}$  & $0.715_{\pm 0.010}$ \\
        CF-10  & $0.592_{\pm 0.015}$  & $0.594_{\pm 0.017}$  & $0.590_{\pm 0.018}$  & $0.579_{\pm 0.009}$  & $0.582_{\pm 0.018}$  & $0.516_{\pm 0.024}$ \\
        SCRUB  & $0.067_{\pm 0.005}$  & $0.073_{\pm 0.007}$  & $0.071_{\pm 0.007}$  & $0.067_{\pm 0.020}$  & $0.076_{\pm 0.008}$  & $0.011_{\pm 0.005}$ \\
        SALUN  & $0.831_{\pm 0.014}$  & $0.833_{\pm 0.011}$  & $0.832_{\pm 0.019}$  & $0.832_{\pm 0.027}$  & $0.842_{\pm 0.009}$  & $0.771_{\pm 0.005}$ \\
        l1-sparse  & $0.185_{\pm 0.005}$  & $0.183_{\pm 0.007}$  & $0.181_{\pm 0.007}$  & $0.184_{\pm 0.023}$  & $0.185_{\pm 0.016}$  & $0.191_{\pm 0.007}$ \\
        COLA  & $0.019_{\pm 0.006}$  & $0.023_{\pm 0.009}$  & $0.021_{\pm 0.009}$  & $0.019_{\pm 0.025}$  & $0.022_{\pm 0.009}$  & $0.007_{\pm 0.011}$ \\
        \bottomrule
\end{tabular}
}
\label{tab:diverse_idi_model2}
\end{table*}
\subsection{Effect of number of Layers for IDI}
\label{Effect_of_number_of_Layers_for_IDI}
\begin{figure*}[!t]
    \begin{subfigure}[b]{0.32\linewidth}
        \centering
        \resizebox{\linewidth}{!}{
            \includegraphics{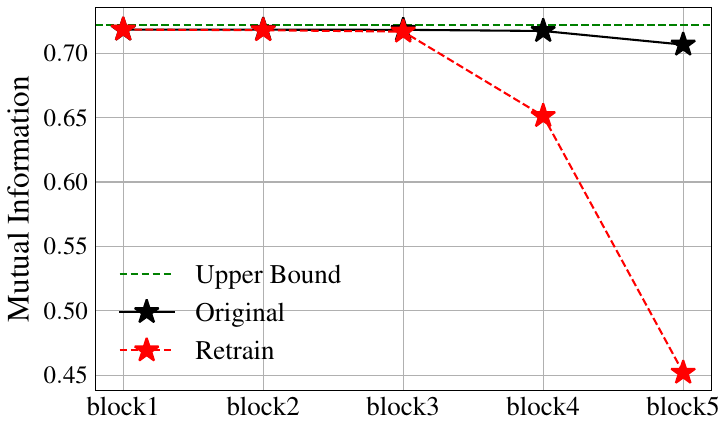}
        }
        \vspace{-1.5em}
        \caption{{CIFAR10 (2 classes)}}
    \end{subfigure}
    \begin{subfigure}[b]{0.32\linewidth}
        \centering
        \resizebox{\linewidth}{!}{
            \includegraphics{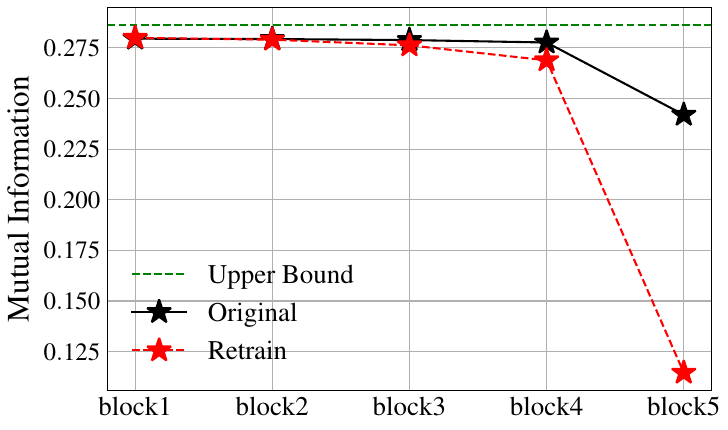}
        }
        \vspace{-1.5em}
        \caption{CIFAR-100 (5 classes)}
    \end{subfigure} 
    \begin{subfigure}[b]{0.32\linewidth}
        \centering
        \resizebox{\linewidth}{!}{
            \includegraphics{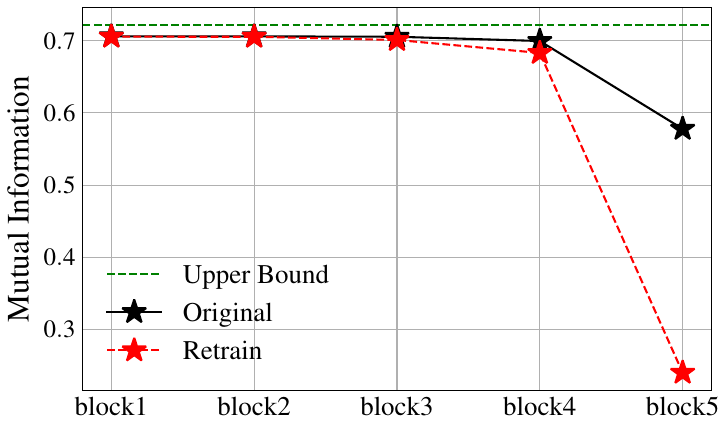}
        }
        \vspace{-1.5em}
        \caption{CIFAR-100 (20 classes)}
    \end{subfigure}
    \caption{
    Mutual information curves for multiple class unlearning in ResNet-18 architecture.  It illustrates the estimated mutual information $I(\mathbf{Z}_{\ell};Y)$ of the features from the $\ell$-th layer $\mathbf{Z}_{\ell}$ and the binary label $Y$, computed by the InfoNCE loss. 
   }
    \label{fig:representation_mi_for_mutliclasses} 
    \vskip -.25in
\end{figure*}

Conceptually, estimating mutual information for IDI involves all intermediate layers, as introduced in~\cref{section_idi}. 
However, in practice, earlier layers exhibit similar mutual information levels across models, as shown in~\cref{fig:mi__resnet18_resnet50_cifar10_class},~\cref{fig:mutual_information_curves}, and~\cref{fig:representation_mi_for_mutliclasses}.
Consequently, estimating mutual information from only a few later layers is sufficient for evaluation. 
This observation aligns with findings in~\citet{yosinski2014transferable, zeiler2014visualizing}, which indicate that earlier layers primarily capture general features, while later layers focus on distinctive features, resulting in greater variability in mutual information.
To validate this approach, we measure IDI using different numbers of accumulated layers from the back, as presented in~\cref{tab:diverse_idi_model} and~\cref{tab:diverse_idi_model2}. 
These experiments use the same settings discussed in~\cref{fig:mi__resnet18_resnet50_cifar10_class}. 
\begin{table*}[t!]
    \centering

    \resizebox{\textwidth}{!}{%
    \begin{tabular}{lcccccccccc}
        \toprule
        &\multicolumn{5}{c}{\textbf{CIFAR-10}} &\multicolumn{5}{c}{\textbf{CIFAR-100}} \\
        \cmidrule(lr){1-1} \cmidrule(lr){2-6} \cmidrule(lr){7-11}
        Methods & \textbf{Order} & $\mathbf{\theta_s} = \text{Retrain}$ & $\mathbf{\theta_s} = \text{COLA}$ & $\mathbf{\theta_s} = \text{EU-10}$ & $\mathbf{\theta_s} = \text{FT}^\star$ & \textbf{Order}
        & $\mathbf{\theta_s} = \text{Retrain}$ & $\mathbf{\theta_s} = \text{COLA}$ & $\mathbf{\theta_s} = \text{EU-10}$ & $\mathbf{\theta_s} = \text{FT}^\star$\\
        \cmidrule(lr){1-1} \cmidrule(lr){2-6} \cmidrule(lr){7-11}
        FT & 8 & $0.671_{\pm 0.008}$ & $0.668_{\pm 0.008}$ & $0.756_{\pm 0.006}$ & $0.662_{\pm 0.008}$ & 11 & $0.610_{\pm 0.022}$ & $0.624_{\pm 0.021}$ & $0.680_{\pm 0.018}$ & $0.481_{\pm 0.029}$ \\
        RL & 10 & $0.830_{\pm 0.005}$ & $0.828_{\pm 0.005}$ & $0.874_{\pm 0.004}$ & $0.825_{\pm 0.005}$ & 9 & $0.467_{\pm 0.010}$ & $0.486_{\pm 0.010}$ & $0.563_{\pm 0.008}$ & $0.291_{\pm 0.013}$ \\
        GA & 6 & $0.334_{\pm 0.014}$ & $0.328_{\pm 0.014}$ & $0.506_{\pm 0.010}$ & $0.315_{\pm 0.014}$ & 8 & $0.392_{\pm 0.021}$ & $0.414_{\pm 0.020}$ & $0.502_{\pm 0.017}$ & $0.191_{\pm 0.028}$\\
        Bad-T & 12 & $1.014_{\pm 0.004}$ & $1.014_{\pm 0.004}$ & $1.010_{\pm 0.003}$ & $1.014_{\pm 0.004}$ & 12 & $1.079_{\pm 0.024}$ & $1.076_{\pm 0.023}$ & $1.065_{\pm 0.020}$ & $1.105_{\pm 0.032}$\\
        EU-5 & 7 & $0.528_{\pm 0.005}$ & $0.523_{\pm 0.005}$ & $0.650_{\pm 0.004}$ & $0.515_{\pm 0.005}$ & 3 & $0.064_{\pm 0.037}$ & $0.098_{\pm 0.036}$ & $0.233_{\pm 0.030}$ & $-0.245_{\pm 0.049}$\\
        CF-5 & 9 & $0.675_{\pm 0.027}$ & $0.672_{\pm 0.027}$ & $0.759_{\pm 0.020}$ & $0.666_{\pm 0.028}$ & 7 & $0.388_{\pm 0.010}$ & $0.410_{\pm 0.010}$ & $0.499_{\pm 0.008}$ & $0.186_{\pm 0.013}$\\
        EU-10 & 1 & $-0.349_{\pm 0.019}$ & $-0.362_{\pm 0.019}$ & $0.0_{\pm 0.014}$ & $-0.387_{\pm 0.020}$ & 1 & $-0.221_{\pm 0.009}$ & $-0.177_{\pm 0.009}$ & $0.0_{\pm 0.007}$ & $-0.624_{\pm 0.012}$\\
        CF-10 & 2 & $-0.060_{\pm 0.017}$ & $-0.070_{\pm 0.017}$ & $0.214_{\pm 0.013}$ & $-0.090_{\pm 0.017}$ & 4 & $0.175_{\pm 0.040}$ & $0.205_{\pm 0.039}$ & $0.324_{\pm 0.033}$ & $-0.097_{\pm 0.053}$\\
        SCRUB & 3 & $-0.056_{\pm 0.008}$ & $-0.066_{\pm 0.008}$ & $0.217_{\pm 0.006}$ & $-0.086_{\pm 0.008}$ & 6 & $0.339_{\pm 0.069}$ & $0.363_{\pm 0.067}$ & $0.458_{\pm 0.057}$ & $0.121_{\pm 0.092}$\\
        SALUN & 11 & $0.936_{\pm 0.012}$ & $0.935_{\pm 0.012}$ & $0.953_{\pm 0.009}$ & $0.934_{\pm 0.012}$ & 10 & $0.529_{\pm 0.022}$ & $0.546_{\pm 0.021}$ & $0.614_{\pm 0.018}$ & $0.373_{\pm 0.029}$\\
        \(\mathbf{\ell_1}\)-sparse & 5 & $0.293_{\pm 0.012}$ & $0.286_{\pm 0.012}$ & $0.476_{\pm 0.009}$ & $0.273_{\pm 0.012}$ & 5 & $0.334_{\pm 0.026}$ & $0.358_{\pm 0.025}$ & $0.454_{\pm 0.021}$ & $0.114_{\pm 0.035}$\\
        COLA & 4 & $0.010_{\pm 0.006}$ & $0.0_{\pm 0.006}$ & $0.266_{\pm 0.004}$ & $-0.018_{\pm 0.006}$ & 2 & $-0.038_{\pm 0.006}$ & $0.0_{\pm 0.006}$ & $0.150_{\pm 0.005}$ & $-0.381_{\pm 0.008}$\\
        \bottomrule
\end{tabular}
    }
\caption{IDI for four different reference models (Retrain, COLA, EU-10, and $\text{FT}^\star$). $\text{FT}^\star$ is finetuned with a learning rate of 5e-5, while FT is finetuned with a learning rate of 1e-5. Since FT typically does not remove all residual information while maintaining test accuracy, using a higher learning rate for $\text{FT}^\star$ can be justified if you want to use it as the reference model. The `Order' has been arranged in ascending sequence according to the IDI values.
}
\label{tab:diverse_base_model}
\end{table*}

Our results demonstrate minimal differences in IDI as $n$ increases, indicating a negligible contribution of earlier layers to IDI. Specifically, when comparing the two columns (``Full layers'' and ``$n$ with $*$''), the discrepancy between the ideal IDI and our practical approach is minimal, empirically supporting the validity of focusing on the last selected layers.
This property is particularly beneficial for reducing computational costs, as mutual information computations for later layers require less time due to the shallower $g$ networks involved (refer to~\cref{subsection_MI_estimation}).
Practitioners can determine the appropriate $n$ by observing the information gap per layer between Original and Retrain for a given unlearning setup.

\begin{figure*}[!t]
    \begin{subfigure}[b]{0.155\linewidth}
        \centering
        \resizebox{\linewidth}{!}{
            \includegraphics{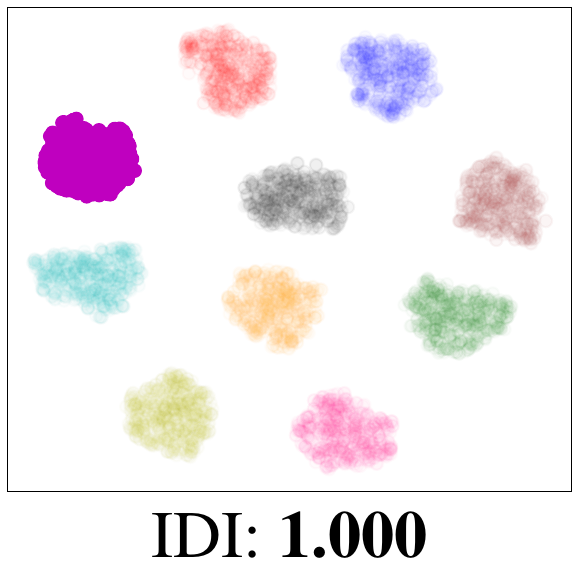}
        }
        \vspace{-1.5em}
        \caption{{Original}}
    \end{subfigure}
    \hfill
    \begin{subfigure}[b]{0.155\linewidth}
        \centering
        \resizebox{\linewidth}{!}{
            \includegraphics{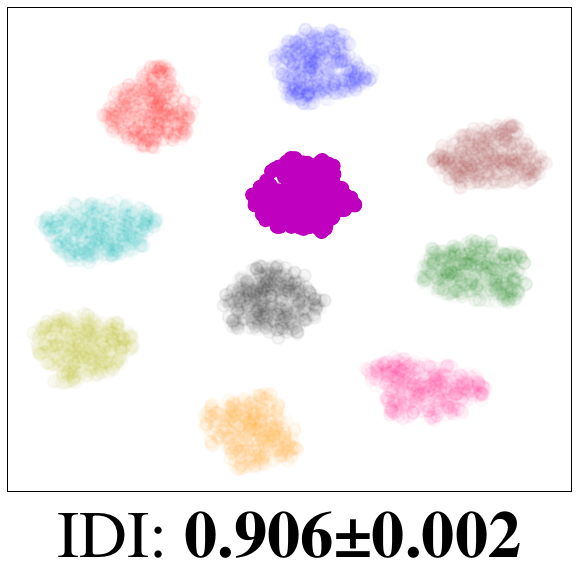}
        }
        \vspace{-1.5em}
        \caption{CF-5}
    \end{subfigure}
    \begin{subfigure}[b]{0.155\linewidth}
        \centering
        \resizebox{\linewidth}{!}{
            \includegraphics{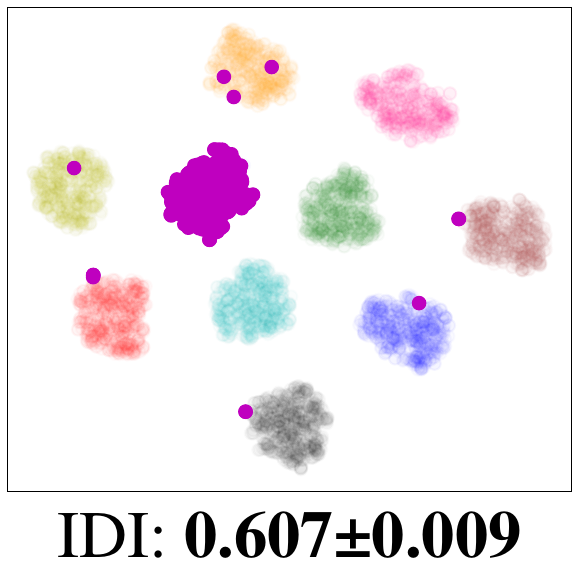}
        }
        \vspace{-1.5em}
        \caption{FT}
    \end{subfigure}
    \begin{subfigure}[b]{0.155\linewidth}
        \centering
        \resizebox{\linewidth}{!}{
            \includegraphics{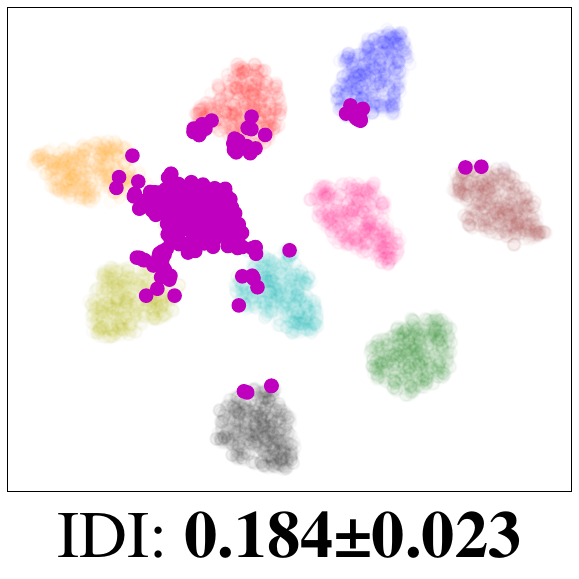}
        }
        \vspace{-1.5em}
        \caption{\(\mathbf{\ell_1}\)-sparse}
    \end{subfigure} 
    \begin{subfigure}[b]{0.155\linewidth}
        \centering
        \resizebox{\linewidth}{!}{
            \includegraphics{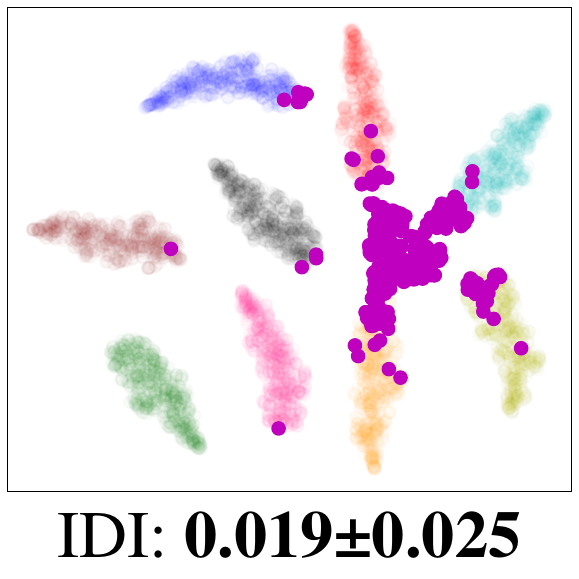}
        }
        \vspace{-1.5em}
        \caption{COLA}
    \end{subfigure}
    \hfill
    \begin{subfigure}[b]{0.155\linewidth}
        \centering
        \resizebox{\linewidth}{!}{
            \includegraphics{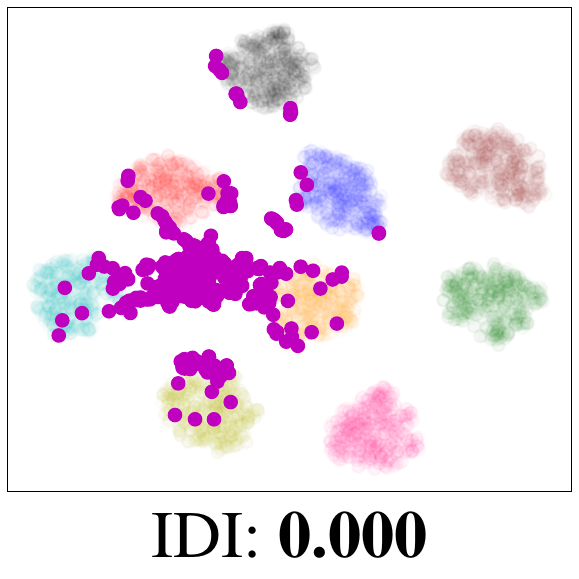}
        }
        \vspace{-1.5em}
        \caption{{Retrain}}
    \end{subfigure}
    \caption{
    t-SNE visualizations of encoder outputs for Original, Retrain, and unlearned models from four MU methods (SALUN, \(\mathbf{\ell_1}\)-sparse, SCRUB, EU-10) on single-class forgetting with (CIFAR-10, ResNet-50). In each t-SNE plot, features of the forgetting class are represented in purple. 
   }
    \label{fig:analysis_t-sne} 
\end{figure*}

\subsection{IDI without Retrain Model}  \label{app: D.2}
In real-world applications, using the Retrain is often infeasible. 
In such cases, any reasonable model can be used as the \textbf{standard model}, 
denoted as $\mathbf{\theta_s}$. 
Although the absence of a Retrain inevitably affects how the IDI value is interpreted
(\ie, an IDI value of zero indicates that unlearning has been properly achieved, equivalent to the Retrain), 
it still provides useful insights into the degree of unlearning achieved relative to the chosen reference. 
To accommodate this, we introduce an extended version of the ID metric. Differences are highlighted with $\textcolor{teal}
{(\cdot)}$:
\begin{equation}
\mathbf{ID}(\mathbf{\theta_u}, \mathbf{\textcolor{teal}{\theta_s}}) = \sum_{\ell=1}^{L} \left( I(\mathbf{Z}^{(\mathbf{u})}_\ell; Y_C) - I(\mathbf{Z}^{(\mathbf{\textcolor{teal}{s}})}_\ell; Y_C) \right).
\label{eq:ID_extension}
\end{equation}
The main difference from the original ID is that $\mathbf{\theta_s}$ can be set as any model including $\mathbf{\theta_r}$ (Retrain), while the previous version fixed $\mathbf{\theta_s} = \mathbf{\theta_r}$.  This extension also leads to the modified IDI metric:
\begin{equation}
\begin{split}
\mathbf{IDI}(\mathbf{\theta_u}, \textcolor{teal}{\mathbf{\theta_s}}) &= \frac{\mathbf{ID}(\mathbf{\theta_u}, \textcolor{teal}{\mathbf{\theta_s}})}{\mathbf{ID}(\mathbf{\theta_0}, \textcolor{teal}{\mathbf{\theta_s}})} \\&= \frac{\sum_{\ell=1}^{L} \left(I(\mathbf{Z}^{(\mathbf{u})}_\ell; Y) - I(\mathbf{Z}^{(\mathbf{\textcolor{teal}{s}})}_\ell; Y)\right)}{\sum_{\ell=1}^{L} \left(I(\mathbf{Z}^{(\mathbf{o})}_\ell; Y) - I(\mathbf{Z}^{(\mathbf{\textcolor{teal}{s}})}_\ell; Y)\right)}
.
\end{split}
\label{eq:IDI_extension}
\end{equation}
We test IDI using different reference models, as demonstrated in \cref{tab:diverse_base_model}. 
Intuitively, since only the standard model changes in \cref{eq:IDI_extension}, the order of the IDI values remains consistent.
\begin{table*}[t!]
% \captionsetup{
%   labelfont={color=blue}, % "표 1" 등의 레이블을 파란색으로
%   textfont={color=blue}   % 캡션 텍스트를 파란색으로
% }
    
    \centering

    \resizebox{\textwidth}{!}{
    \begin{tabular}{lccccccccccccc}
        \toprule[1pt]
        \midrule
        \multicolumn{6}{c}{ResNet-18} & \multicolumn{7}{c}{ResNet-50} \\
        \cmidrule(lr){1-6} \cmidrule(lr){7-13}
        Ratios & Block1 & Block2 & Block3 & Block4 & Block5 & Ratios & Block1 & Block2 & Block3 & Block4 & Block5  & Block6 \\
        \cmidrule(lr){1-6} \cmidrule(lr){7-13}
        10\%
        & $1.61_{\pm 0.05}$
        & $1.55_{\pm 0.21}$
        & $1.63_{\pm 0.09}$
        & $1.49_{\pm 0.11}$
        & $1.42_{\pm 0.13}$
        & 10\%
        & $4.17_{\pm 0.13}$
        & $3.85_{\pm 0.04}$
        & $3.43_{\pm 0.01}$
        & $3.42_{\pm 0.15}$
        & $3.35_{\pm 0.02}$
        & $3.24_{\pm 0.15}$ \\
        Full
        & $6.64_{\pm 0.08}$
        & $6.51_{\pm 0.17}$
        & $6.32_{\pm 0.12}$
        & $6.04_{\pm 0.10}$
        & $5.90_{\pm 0.15}$
        & Full
        & $19.91_{\pm 0.18}$
        & $17.70_{\pm 0.01}$
        & $15.75_{\pm 0.02}$
        & $15.32_{\pm 0.11}$
        & $14.98_{\pm 0.07}$
        & $14.83_{\pm 0.04}$ \\
        \cmidrule(lr){1-6} \cmidrule(lr){7-13}
        \toprule[0.5pt]
        \multicolumn{13}{c}{ViT} \\
        \midrule
        Ratios & Block1 & Block2 & Block3 & Block4 & Block5 & Block6 & Block7 & Block8 & Block9 & Block10 & Block11  & Block12 \\
        \midrule
        10\%
        & $32.46$
        & $32.12$
        & $31.43$
        & $30.99$
        & $30.64$
        & $30.21$
        & $29.50$
        & $29.01$
        & $28.75$
        & $28.43$
        & $27.95$
        & $27.14$
        \\ 
        Full
        & $167.00$
        & $162.81$
        & $160.64$
        & $159.56$
        & $157.65$
        & $155.35$
        & $151.82$
        & $147.78$
        & $146.05$
        & $144.25$
        & $142.81$
        & $139.61$
        \\ \midrule
        \bottomrule
    \end{tabular}
    }
\caption{Runtime (in minutes) for mutual information estimation at the final layer of each block. A `block' refers to a group of residual layers in ResNet (commonly referred to as stages, with four blocks in ResNet-18) or a transformer block in ViT. Results are presented for evaluations conducted using 10\% of the retain set and the full dataset in the CIFAR-10 single-class forgetting scenario.}
\label{tab:mi_runtime_cifar100}
\end{table*}

\subsection{IDI and t-SNE Relationship} \label{app: D.3}

In~\cref{subsection_residual_tsne_recovery}, we identified significant residual information in unlearned models through their tightly clustered t-SNE plots (see~\cref{fig:representation_tsne}) and their ability to easily recover forgotten information (see~\cref{fig:exp__recovery}). 
Black-box assessments failed to detect these residuals, as shown by the success of HD (see~\cref{fig:exp__hd_rte_ta}), which only altered the last layer.
In contrast, IDI effectively captures these hidden residuals, showing a strong correlation with t-SNE plots (see~\cref{fig:analysis_t-sne}), and aligning with accuracy recovered across unlearned models (see~\cref{fig:exp__recovery}).
By complementing existing metrics, IDI offers a comprehensive evaluation of approximate MU methods, addressing crucial aspects to ensure strong unlearning beyond superficial modifications.

\begin{figure*}[t!]
    \hfill
    \begin{subfigure}[b]{0.3\linewidth}
        \centering
        \includegraphics[width=\linewidth]{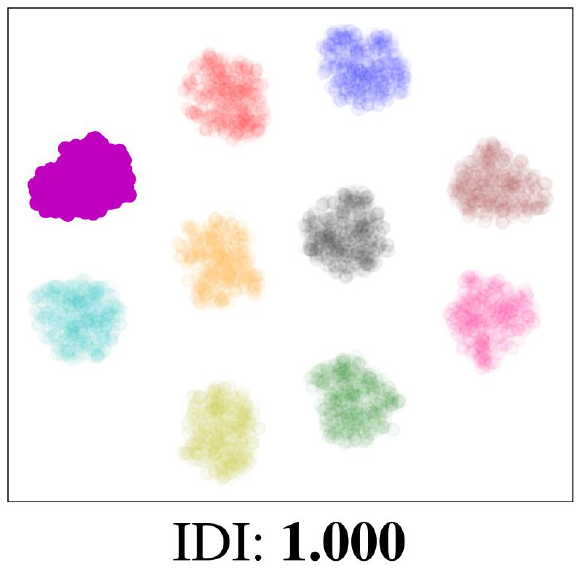}
        \caption{Original}
    \end{subfigure}
    \hspace{4em}
    \begin{subfigure}[b]{0.3\linewidth}
        \centering
        \includegraphics[width=\linewidth]{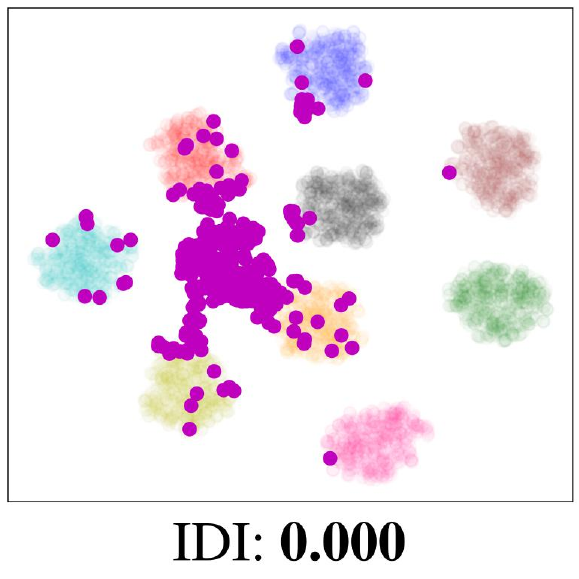}
        \caption{Retrain}
    \end{subfigure}
    \hfill
    \vspace{1em}

    \begin{tabular}{cccc}
        \begin{subfigure}[b]{0.23\linewidth}
            \centering
            \includegraphics[width=\linewidth]{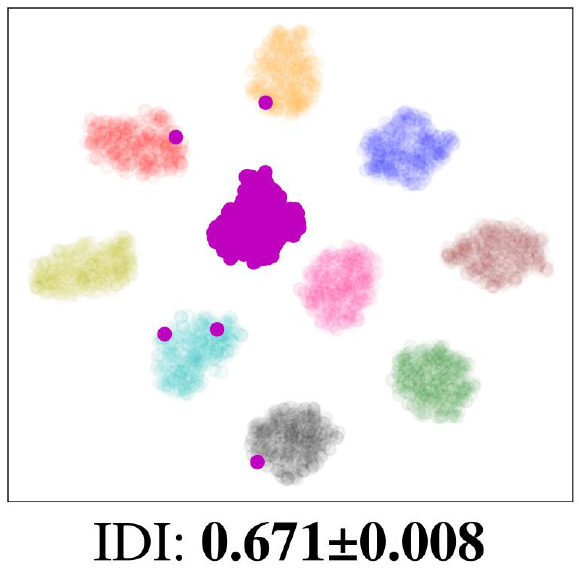}
            \caption{FT}
        \end{subfigure} &
        \begin{subfigure}[b]{0.23\linewidth}
            \centering
            \includegraphics[width=\linewidth]{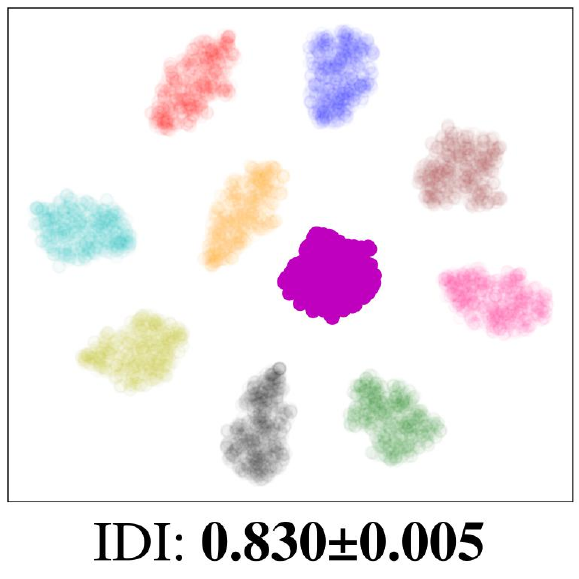}
            \caption{RL}
        \end{subfigure} &
        \begin{subfigure}[b]{0.23\linewidth}
            \centering
            \includegraphics[width=\linewidth]{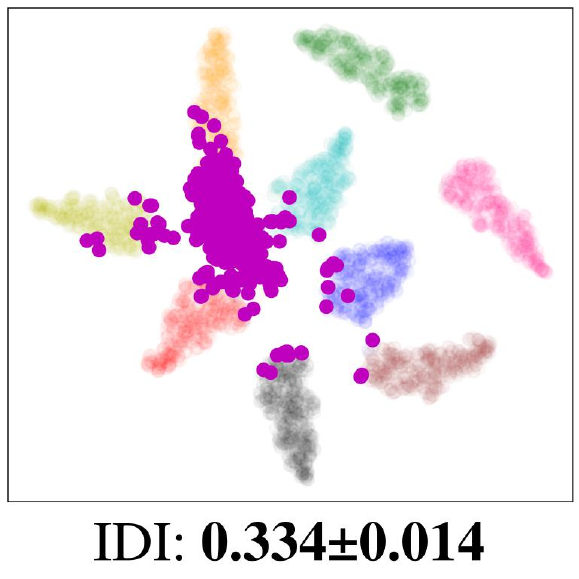}
            \caption{GA}
        \end{subfigure} &
        \begin{subfigure}[b]{0.23\linewidth}
            \centering
            \includegraphics[width=\linewidth]{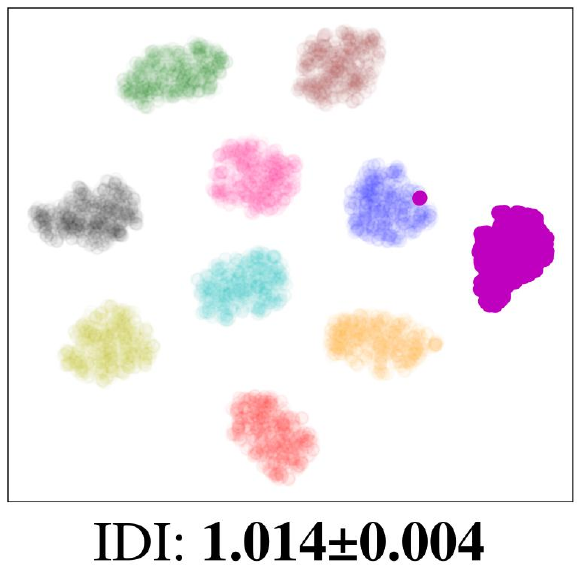}
            \caption{Bad-T}
        \end{subfigure} \\

        \begin{subfigure}[b]{0.23\linewidth}
            \centering
            \includegraphics[width=\linewidth]{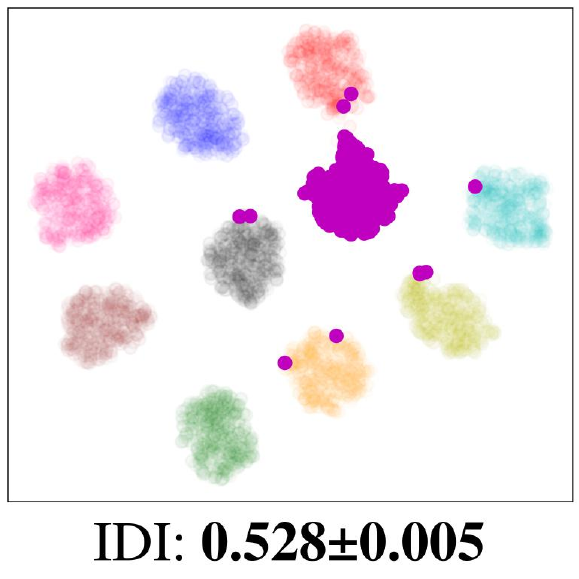}
            \caption{EU-5}
        \end{subfigure} &
        \begin{subfigure}[b]{0.23\linewidth}
            \centering
            \includegraphics[width=\linewidth]{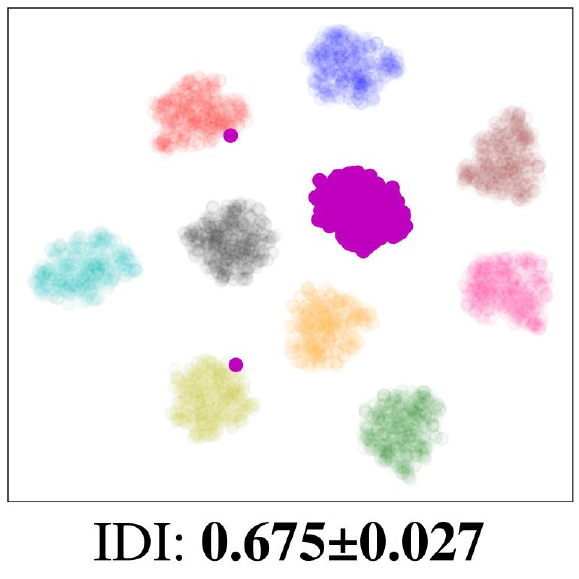}
            \caption{CF-5}
        \end{subfigure} &
        \begin{subfigure}[b]{0.23\linewidth}
            \centering
            \includegraphics[width=\linewidth]{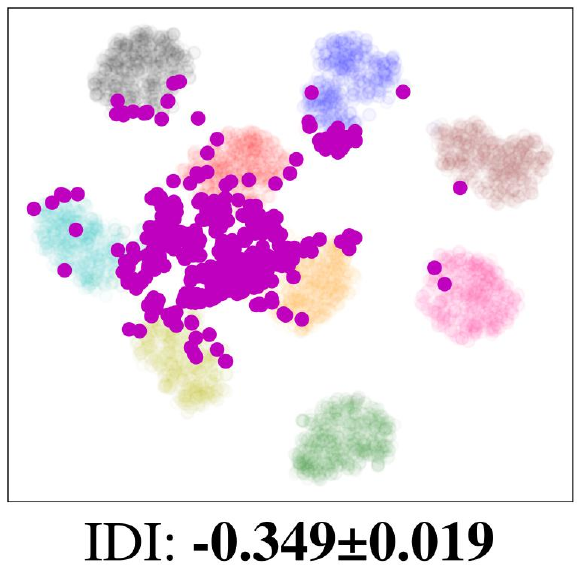}
            \caption{EU-10}
        \end{subfigure} &
        \begin{subfigure}[b]{0.23\linewidth}
            \centering
            \includegraphics[width=\linewidth]{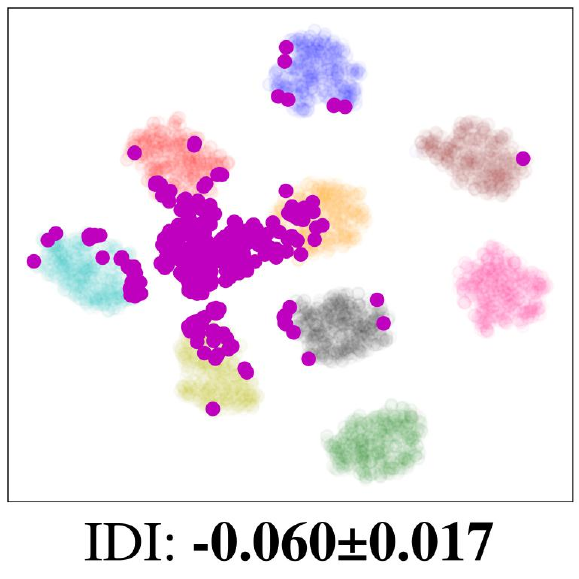}
            \caption{CF-10}
        \end{subfigure} \\

        \begin{subfigure}[b]{0.23\linewidth}
            \centering
            \includegraphics[width=\linewidth]{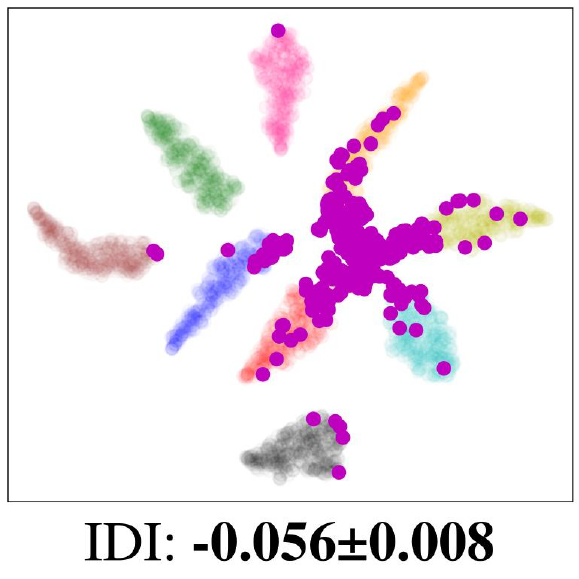}
            \caption{SCRUB}    
        \end{subfigure} &
        \begin{subfigure}[b]{0.23\linewidth}
            \centering
            \includegraphics[width=\linewidth]{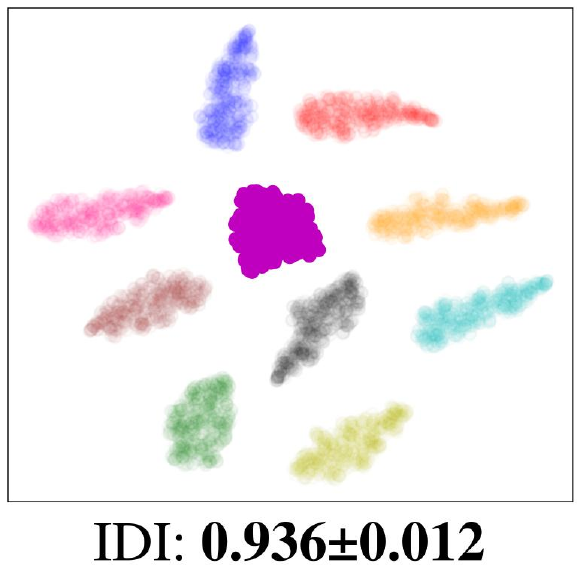}
            \caption{SALUN}
        \end{subfigure} &
        \begin{subfigure}[b]{0.23\linewidth}
            \centering
            \includegraphics[width=\linewidth]{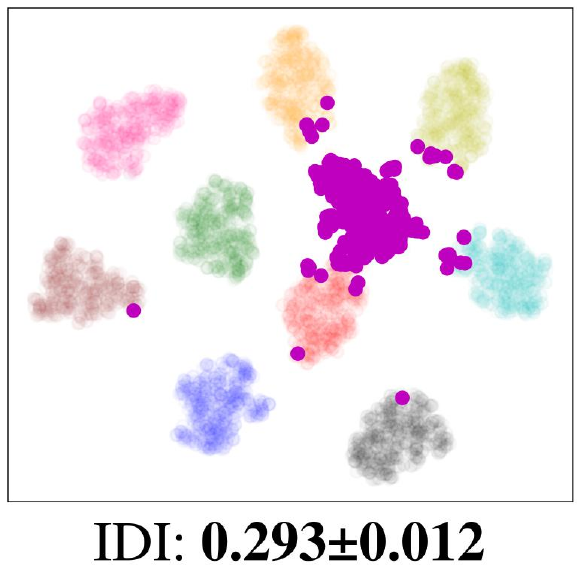}
            \caption{\(\mathbf{\ell_1}\)-sparse}
        \end{subfigure} &
        \begin{subfigure}[b]{0.23\linewidth}
            \centering
            \includegraphics[width=\linewidth]{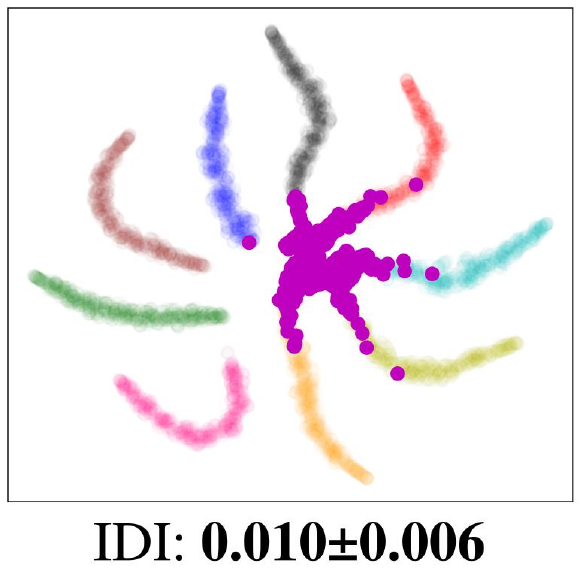}
            \caption{COLA}
        \end{subfigure} \\
    \end{tabular}
    \caption{t-SNE visualizations of features of Original, Retrain, and unlearned models (FT, RL, GA, Bad-T, EU-5, CF-5, EU-10, CF-10, SCRUB, SALUN, \(\mathbf{\ell_1}\)-sparse, and COLA) on CIFAR-10 with ResNet-18. The forgetting class is represented in purple, while rest of the points represents the remaining class.}
    \label{fig:resnet18_cifar10_alltsne}
\end{figure*}

\begin{figure*}[t!]
    \hfill
    \begin{subfigure}[b]{0.3\linewidth}
        \centering
        \includegraphics[width=\linewidth]{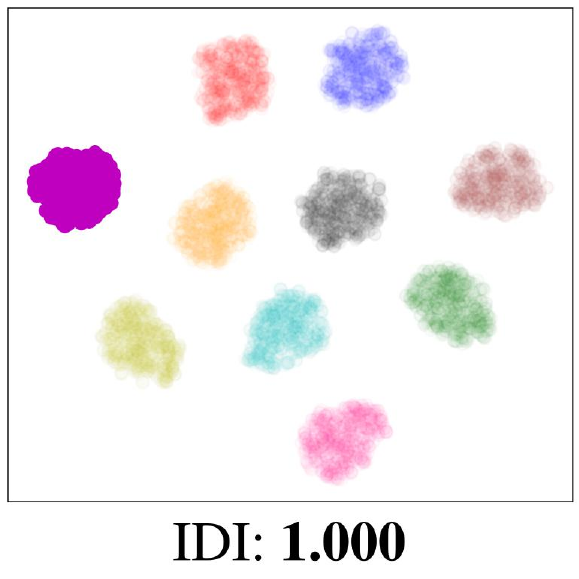}
        \caption{Original}
    \end{subfigure}
    \hspace{4em}
    \begin{subfigure}[b]{0.3\linewidth}
        \centering
        \includegraphics[width=\linewidth]{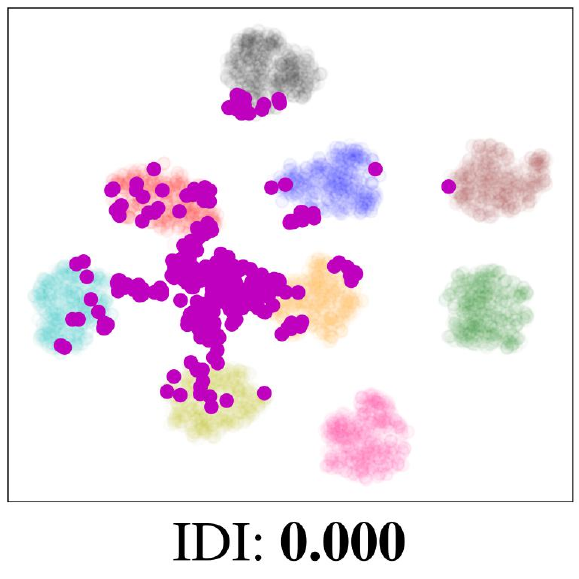}
        \caption{Retrain}
    \end{subfigure}
    \hfill
    \vspace{1em}

    \begin{tabular}{cccc}
        \begin{subfigure}[b]{0.23\linewidth}
            \centering
            \includegraphics[width=\linewidth]{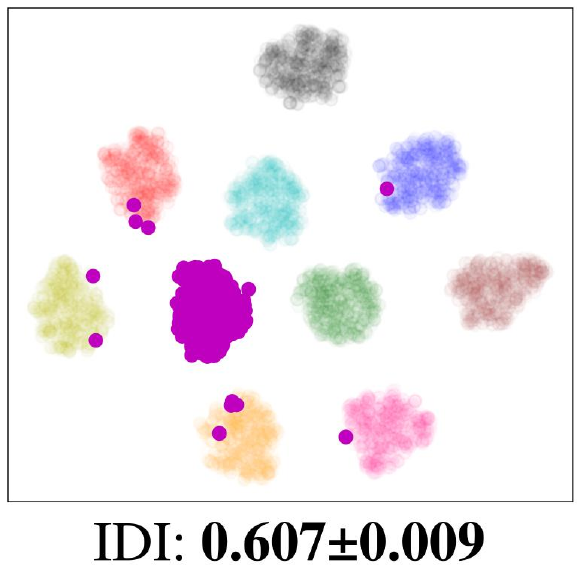}
            \caption{FT}
        \end{subfigure} &
        \begin{subfigure}[b]{0.23\linewidth}
            \centering
            \includegraphics[width=\linewidth]{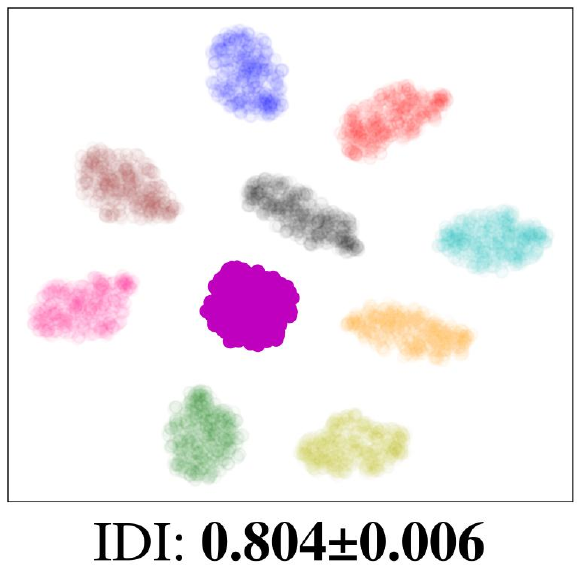}
            \caption{RL}
        \end{subfigure} &
        \begin{subfigure}[b]{0.23\linewidth}
            \centering
            \includegraphics[width=\linewidth]{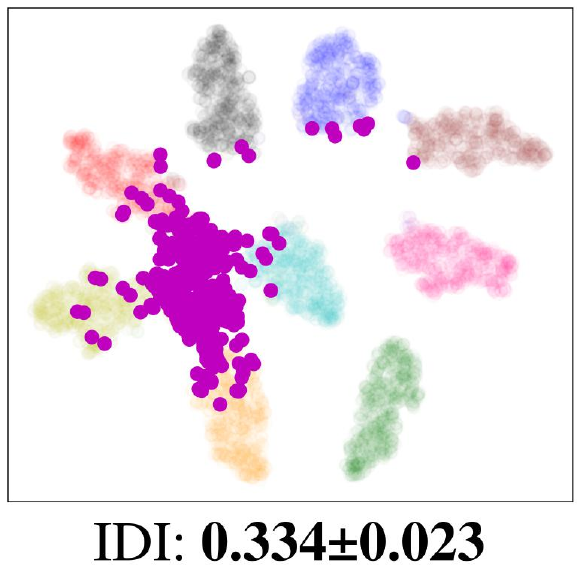}
            \caption{GA}
        \end{subfigure} &
        \begin{subfigure}[b]{0.23\linewidth}
            \centering
            \includegraphics[width=\linewidth]{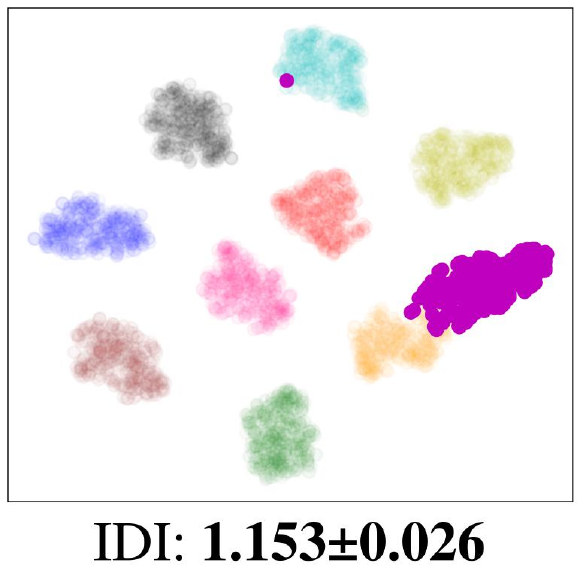}
            \caption{Bad-T}
        \end{subfigure} \\ \\

        \begin{subfigure}[b]{0.23\linewidth}
            \centering
            \includegraphics[width=\linewidth]{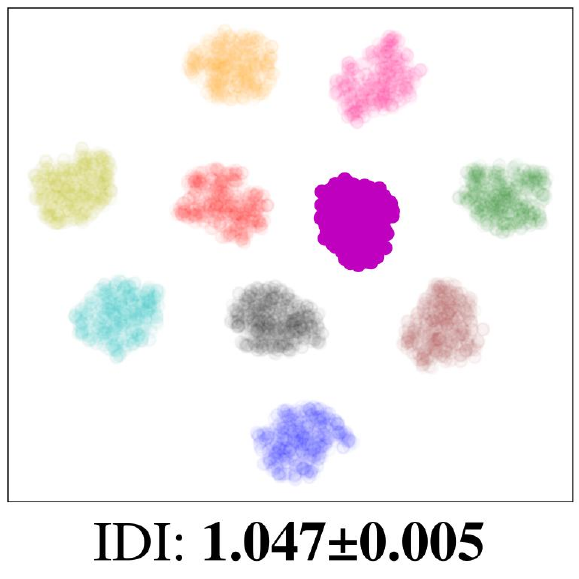}
            \caption{EU-5}
        \end{subfigure} &
        \begin{subfigure}[b]{0.23\linewidth}
            \centering
            \includegraphics[width=\linewidth]{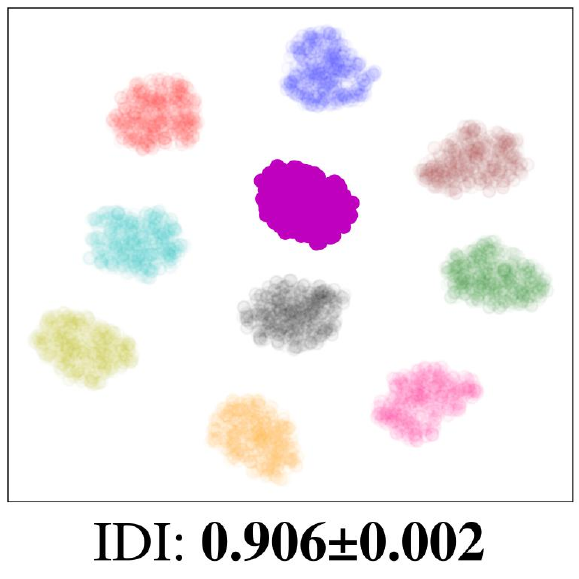}
            \caption{CF-5}
        \end{subfigure} &
        \begin{subfigure}[b]{0.23\linewidth}
            \centering
            \includegraphics[width=\linewidth]{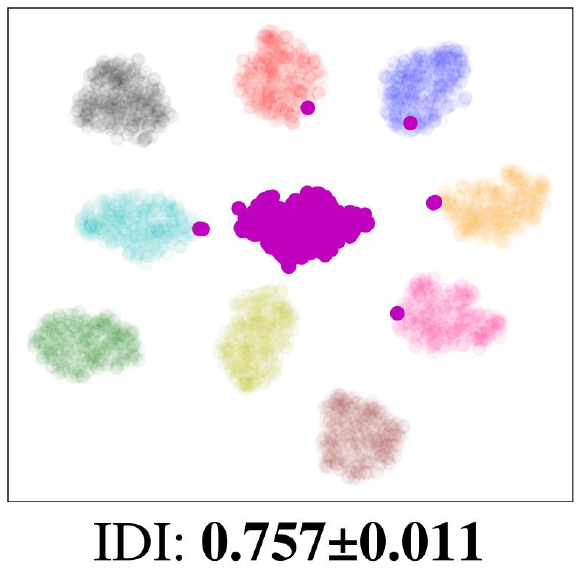}
            \caption{EU-10}
        \end{subfigure} &
        \begin{subfigure}[b]{0.23\linewidth}
            \centering
            \includegraphics[width=\linewidth]{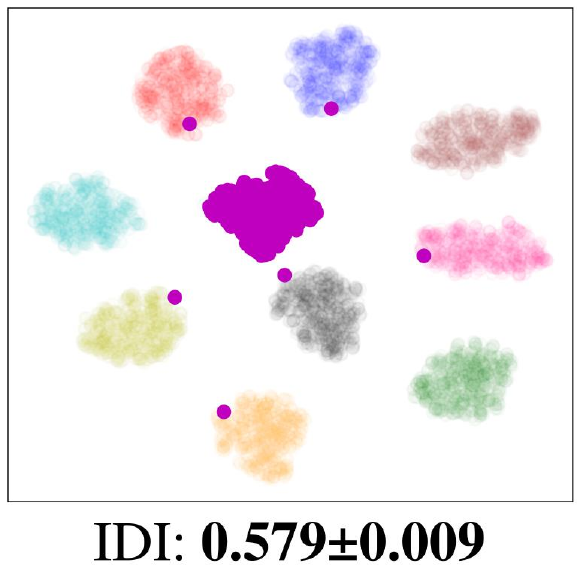}
            \caption{CF-10}
        \end{subfigure} \\ \\

        \begin{subfigure}[b]{0.23\linewidth}
            \centering
            \includegraphics[width=\linewidth]{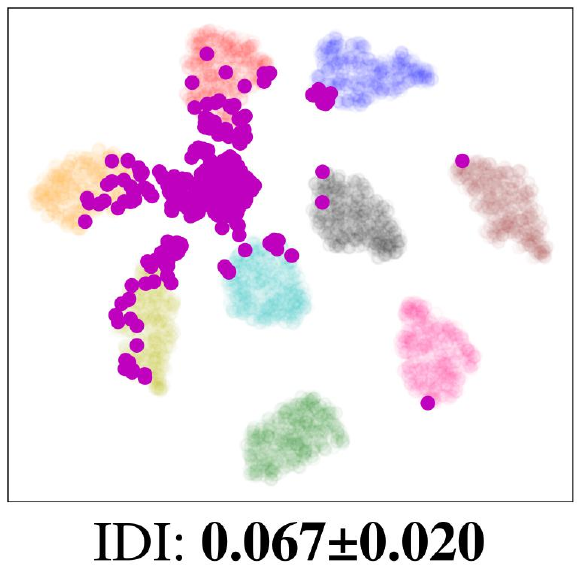}
            \caption{SCRUB}
        \end{subfigure} &
        \begin{subfigure}[b]{0.23\linewidth}
            \centering
            \includegraphics[width=\linewidth]{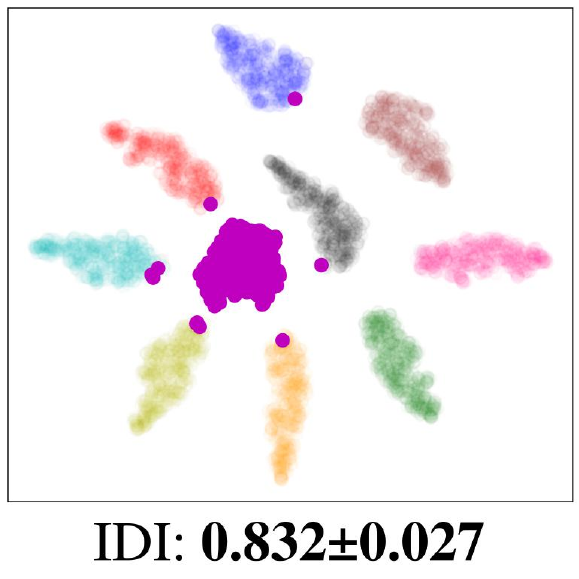}
            \caption{SALUN}
        \end{subfigure} &
        \begin{subfigure}[b]{0.23\linewidth}
            \centering
            \includegraphics[width=\linewidth]{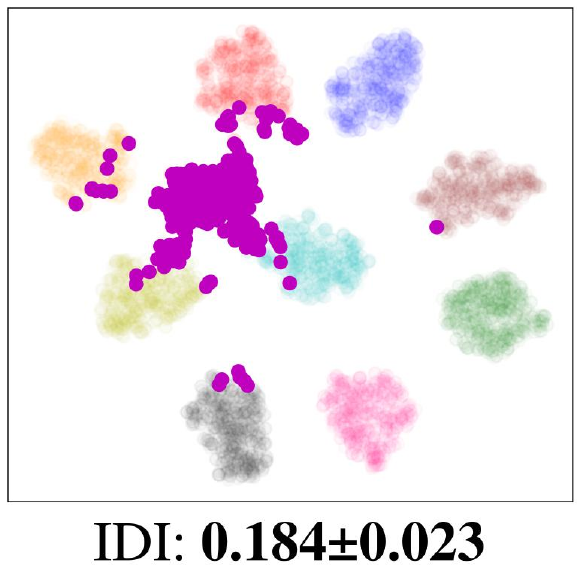}
            \caption{\(\mathbf{\ell_1}\)-sparse}
        \end{subfigure} &
        \begin{subfigure}[b]{0.23\linewidth}
            \centering
            \includegraphics[width=\linewidth]{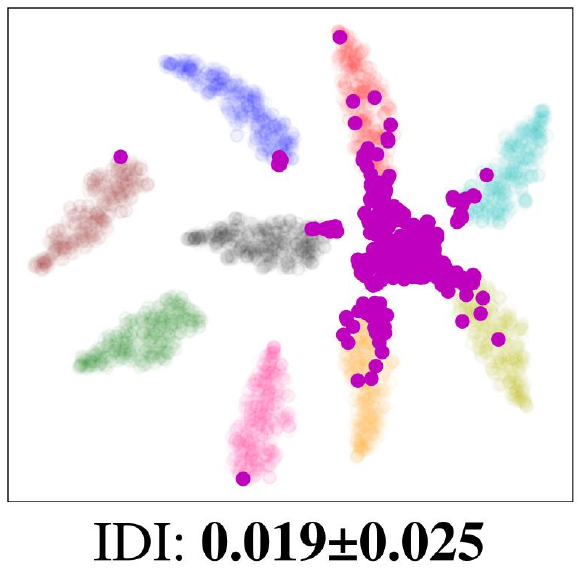}
            \caption{COLA}
        \end{subfigure} \\
    \end{tabular}
    \caption{t-SNE visualizations of feature of Original, Retrain, and unlearned models (FT, RL, GA, Bad-T, EU-5, CF-5, EU-10, CF-10, SCRUB, SALUN, \(\mathbf{\ell_1}\)-sparse, and COLA) on CIFAR-10 with ResNet-50. The forgetting class is represented in purple, while rest of the points represents the remaining class.}
    \label{fig:resnet50_cifar10_alltsne}
\end{figure*}

\begin{figure*}[t!]
    \hfill
    \begin{subfigure}[b]{0.3\linewidth}
        \centering
        \includegraphics[width=\linewidth]{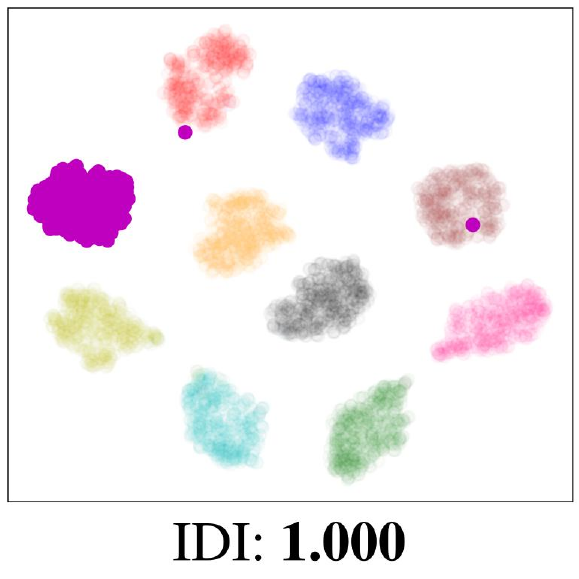}
        \caption{Original}
    \end{subfigure}
    \hspace{4em}
    \begin{subfigure}[b]{0.3\linewidth}
        \centering
        \includegraphics[width=\linewidth]{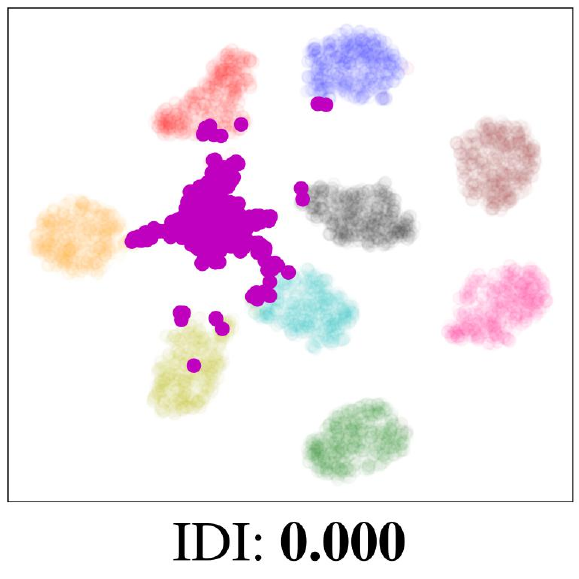}
        \caption{Retrain}
    \end{subfigure}
    \hfill
    \vspace{1em}

    \begin{tabular}{cccc}
        \begin{subfigure}[b]{0.23\linewidth}
            \centering
            \includegraphics[width=\linewidth]{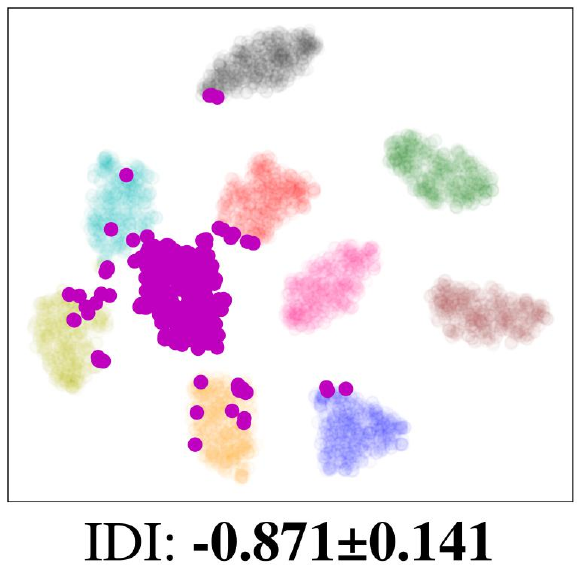}
            \caption{FT}
        \end{subfigure} &
        \begin{subfigure}[b]{0.23\linewidth}
            \centering
            \includegraphics[width=\linewidth]{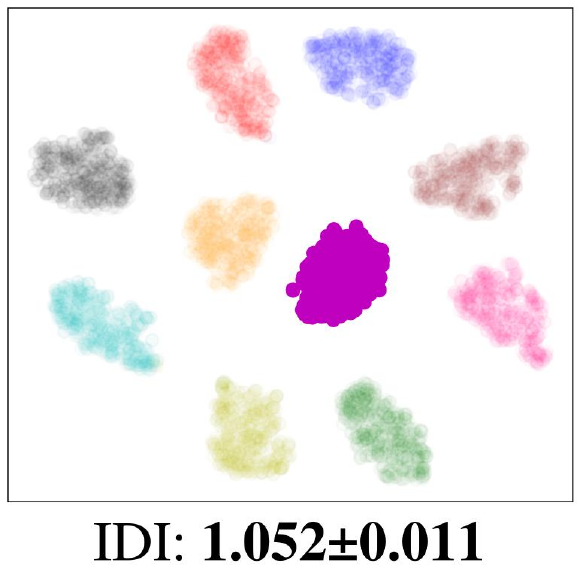}
            \caption{RL}
        \end{subfigure} &
        \begin{subfigure}[b]{0.23\linewidth}
            \centering
            \includegraphics[width=\linewidth]{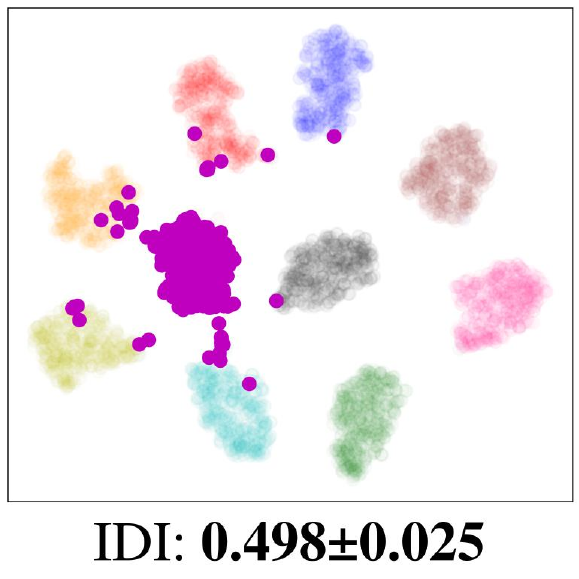}
            \caption{GA}
        \end{subfigure} &
        \begin{subfigure}[b]{0.23\linewidth}
            \centering
            \includegraphics[width=\linewidth]{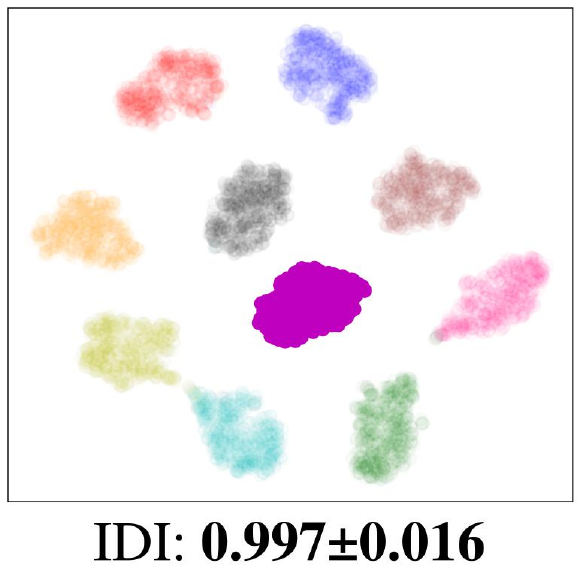}
            \caption{Bad-T}
        \end{subfigure} \\

        \begin{subfigure}[b]{0.23\linewidth}
            \centering
            \includegraphics[width=\linewidth]{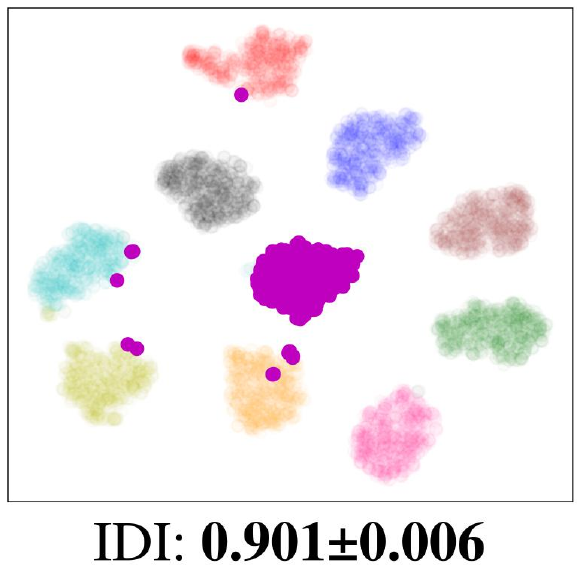}
            \caption{EU-5}
        \end{subfigure} &
        \begin{subfigure}[b]{0.23\linewidth}
            \centering
            \includegraphics[width=\linewidth]{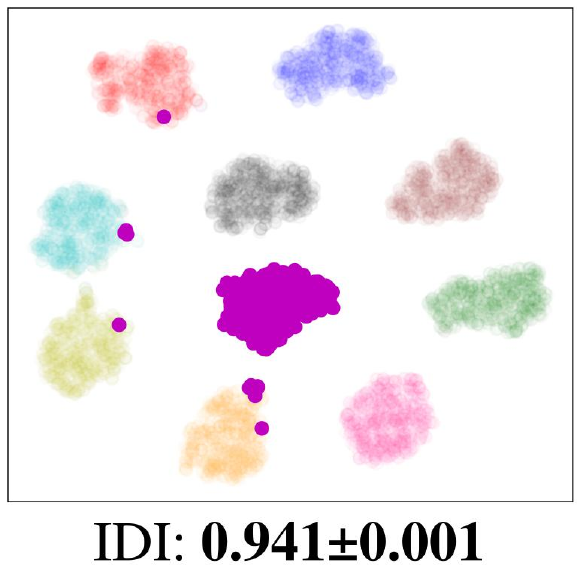}
            \caption{CF-5}
        \end{subfigure} &
        \begin{subfigure}[b]{0.23\linewidth}
            \centering
            \includegraphics[width=\linewidth]{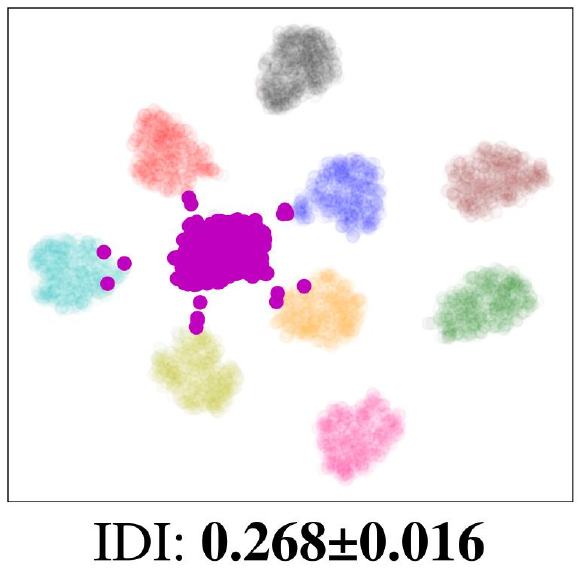}
            \caption{EU-10}
        \end{subfigure} &
        \begin{subfigure}[b]{0.23\linewidth}
            \centering
            \includegraphics[width=\linewidth]{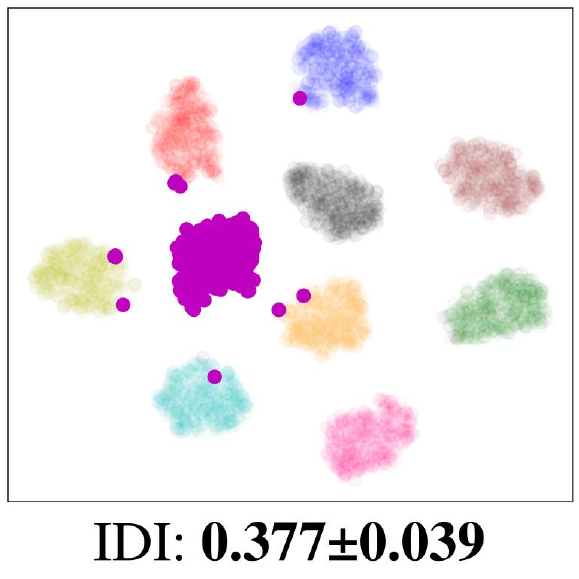}
            \caption{CF-10}
        \end{subfigure} \\

        \begin{subfigure}[b]{0.23\linewidth}
            \centering
            \includegraphics[width=\linewidth]{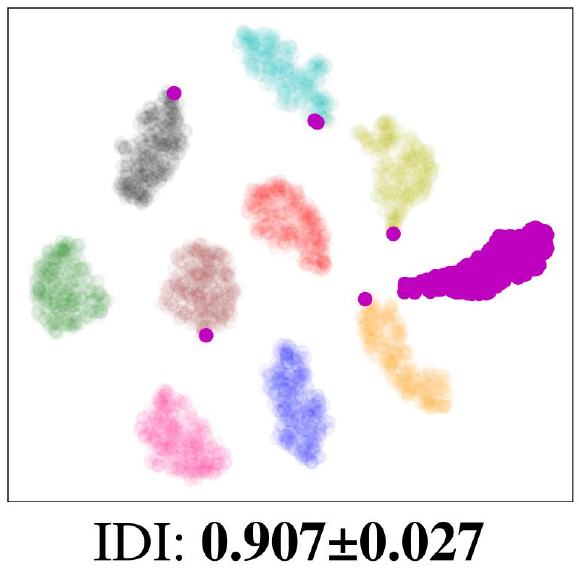}
            \caption{SCRUB}
        \end{subfigure} &
        \begin{subfigure}[b]{0.23\linewidth}
            \centering
            \includegraphics[width=\linewidth]{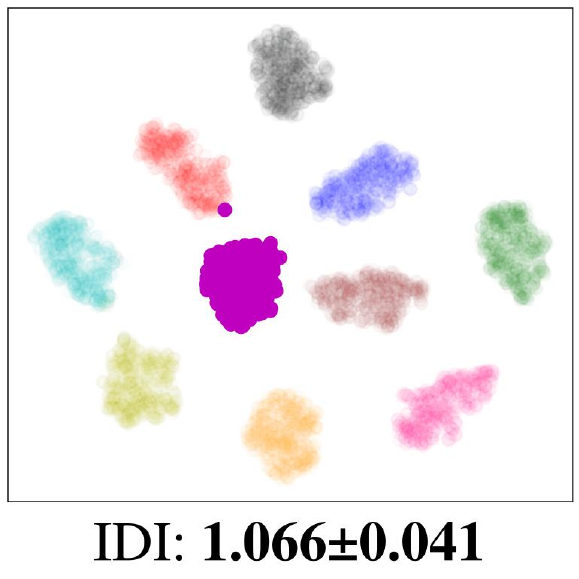}
            \caption{SALUN}
        \end{subfigure} &
        \begin{subfigure}[b]{0.23\linewidth}
            \centering
            \includegraphics[width=\linewidth]{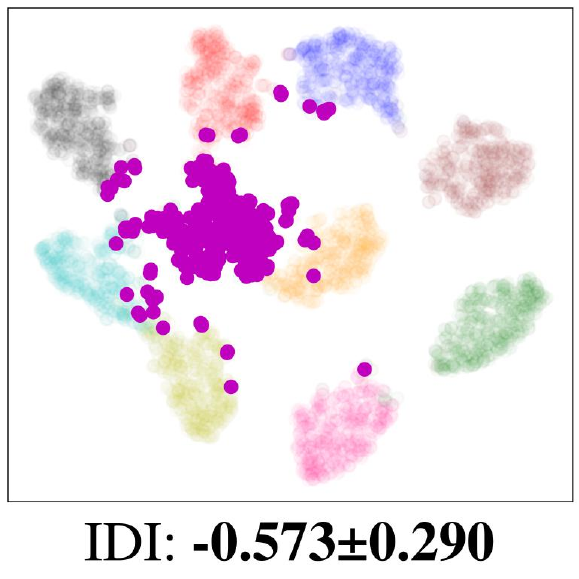}
            \caption{\(\mathbf{\ell_1}\)-sparse}
        \end{subfigure} &
        \begin{subfigure}[b]{0.23\linewidth}
            \centering
            \includegraphics[width=\linewidth]{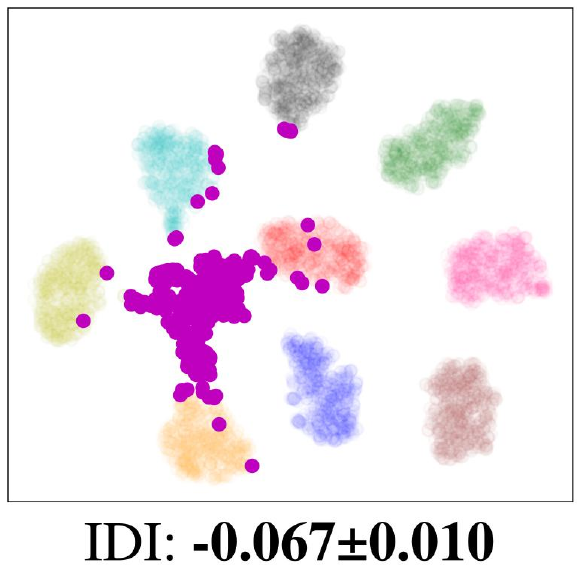}
            \caption{COLA}
        \end{subfigure} \\
    \end{tabular}
    \caption{t-SNE visualizations of features of Original, Retrain, and unlearned models (FT, RL, GA, Bad-T, EU-5, CF-5, EU-10, CF-10, SCRUB, SALUN, \(\mathbf{\ell_1}\)-sparse, and COLA) on CIFAR-10 with ViT. The forgetting class is represented in purple, while rest of 
the points represents the remaining class.}
    \label{fig:vit_cifar10_alltsne}
\end{figure*}

Figure~\ref{fig:resnet18_cifar10_alltsne} presents the full t-SNE plots illustrating the intermediate features and corresponding IDI measurements of MU baselines on the single-class unlearning on CIFAR-10 with ResNet-18. 
A high IDI corresponds to better clustering and similarity among features of the forgetting class, as seen in (l) SALUN and (f) Bad-T, which show inadequate unlearning performance. 
These examinations show the high relationship between IDI and the residual information of forget set.
Additionally, the IDI metric reveals instances of over-unlearning, where the forgetting class becomes excessively dispersed, as demonstrated in (i) EU-10. 
Among the evaluated methods, (n) COLA has the closest IDI to Retrain, suggesting its high efficacy in achieving the desired removal of the forget set influence in the intermediate layers of the model. 
This trend is also visible in ResNet-50 (see Figure~\ref{fig:resnet50_cifar10_alltsne}) and ViT (see Figure~\ref{fig:vit_cifar10_alltsne}).

\begin{figure*}[t!]
    \centering
    \begin{subfigure}[b]{0.4\linewidth}
        \centering
        \includegraphics[width=\linewidth]{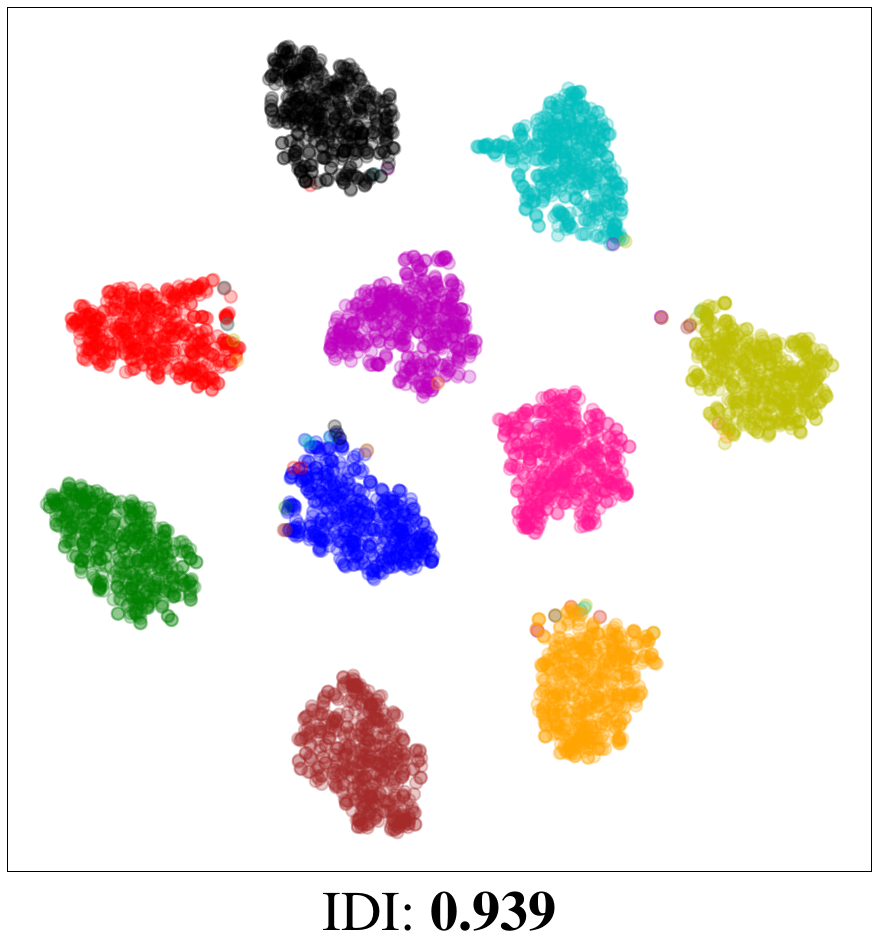}
        \caption{Bad-T}
    \end{subfigure}
    \hspace{4em}
    \begin{subfigure}[b]{0.4\linewidth}
        \centering
        \includegraphics[width=\linewidth]{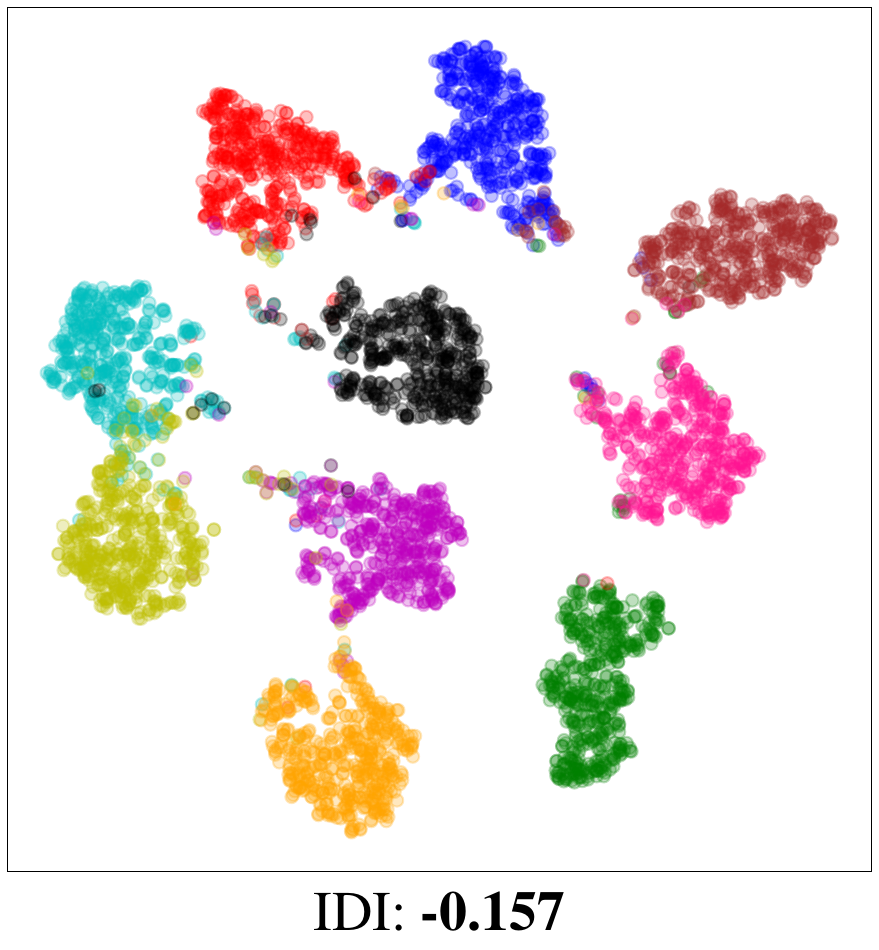}
        \caption{\(\mathbf{\ell_1}\)-sparse}
    \end{subfigure}
    \caption{t-SNE visualizations of features of forget samples of Bad-T and \(\mathbf{\ell_1}\)-sparse in a random data forgetting task on (CIFAR-10, ResNet-18). The clusters of \(\mathbf{\ell_1}\)-sparse are more disperse than those of Bad-T.}

    \label{fig:sample_unlearn_tsne}
\end{figure*}

Furthermore, we confirm that IDI for the random data forgetting correctly captures the encoder`s information, similar to IDI for the class-wise forgetting.
In Figure~\ref{fig:sample_unlearn_tsne}, the t-SNE plots of forget sample features for two baselines with the same unlearning accuracy (UA) -- Bad-T and \(\mathbf{\ell_1}\)-sparse -- and their IDI values in the random data forgetting task is visualized. 
Comparing them, IDI successfully reflects the residual information in the features, as the features of Bad-T form more compact clusters than those of \(\mathbf{\ell_1}\)-sparse, indicating more influence of the forget set remains in Bad-T.
IDI the for random data forgetting captures the hidden information that cannot be noticed from existing metrics, which may suggest that both methods unlearn similarly due to their same forget accuracy.

\subsection{Mutual Information and Accuracy}

We extend the experiment to measure the accuracy of the intermediate features of the model's encoder. 
Similar to measuring MI using the InfoNCE loss, we freeze the layers up to the $\ell$-th layer of the encoder and train the remaining encoder layers and an additional head using cross-entropy loss.
The additional head perform binary classification to determine whether the input belongs to the retain or forget set. 

\begin{figure*}[t!]
    \centering
    \begin{subfigure}[b]{0.48\textwidth}
        \centering
        \includegraphics[width=\textwidth]{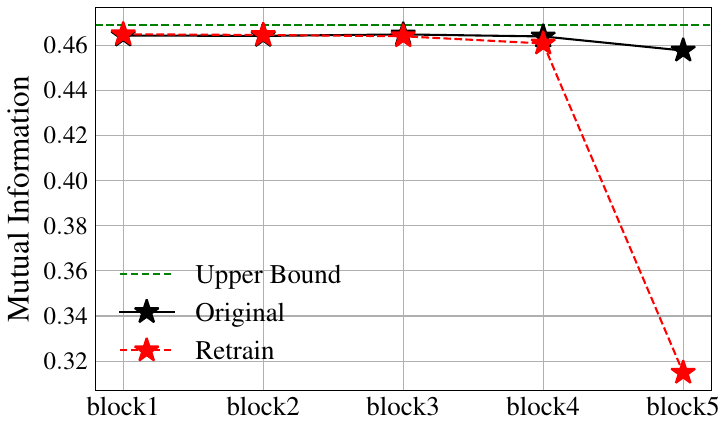}
        \caption{CIFAR-10 - ResNet-18}
        \label{fig:figure1}
    \end{subfigure}
    \hfill
    \begin{subfigure}[b]{0.48\textwidth}
        \centering
        \includegraphics[width=\textwidth]{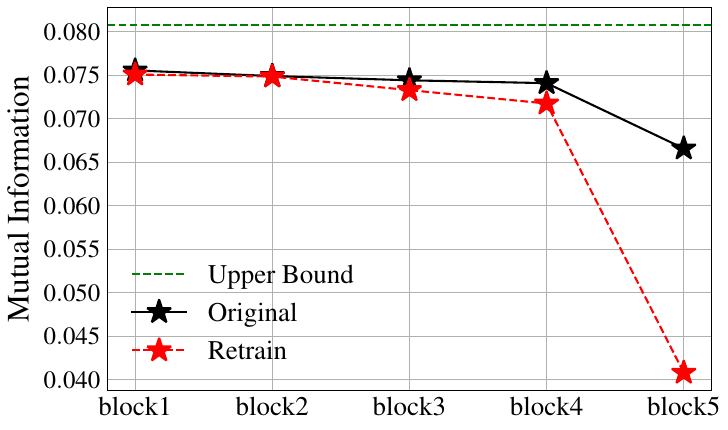}
        \caption{CIFAR-100 - ResNet-18}
        \label{fig:figure2}
    \end{subfigure}
    
    \vspace{1cm} % Adjust the vertical space between the rows
    
    \begin{subfigure}[b]{0.48\textwidth}
        \centering
        \includegraphics[width=\textwidth]{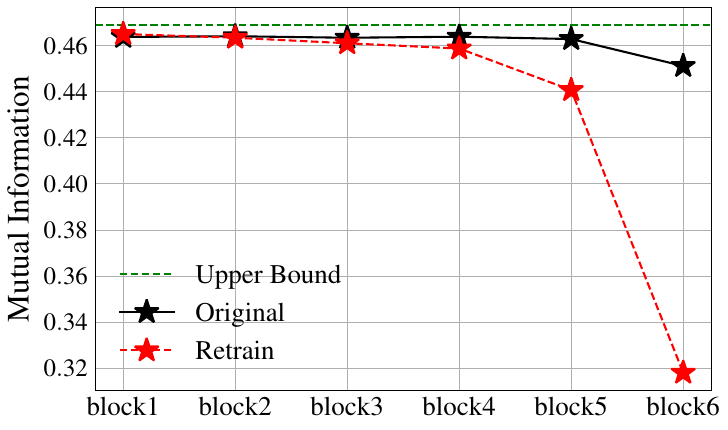}
        \caption{CIFAR-10 - ResNet-50}
        \label{fig:figure3}
    \end{subfigure}
    \hfill
    \begin{subfigure}[b]{0.48\textwidth}
        \centering
        \includegraphics[width=\textwidth]{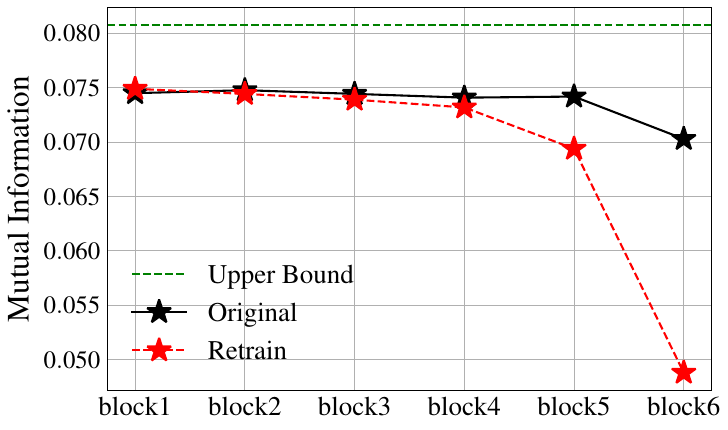}
        \caption{CIFAR-100 - ResNet-50}
        \label{fig:figure4}
    \end{subfigure}
    
    \vspace{1cm} % Adjust the vertical space between the rows
    
    \begin{subfigure}[b]{0.48\textwidth}
        \centering
        \includegraphics[width=\textwidth]{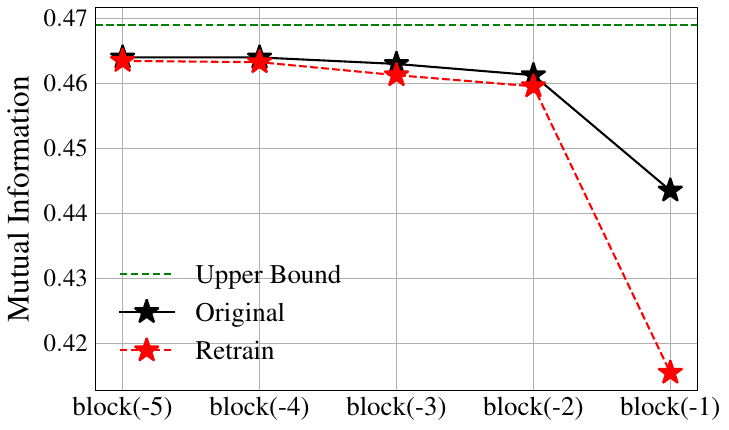}
        \caption{CIFAR-10 - ViT}
        \label{fig:figure5}
    \end{subfigure}
    \hfill
    \begin{subfigure}[b]{0.48\textwidth}
        \centering
        \includegraphics[width=\textwidth]{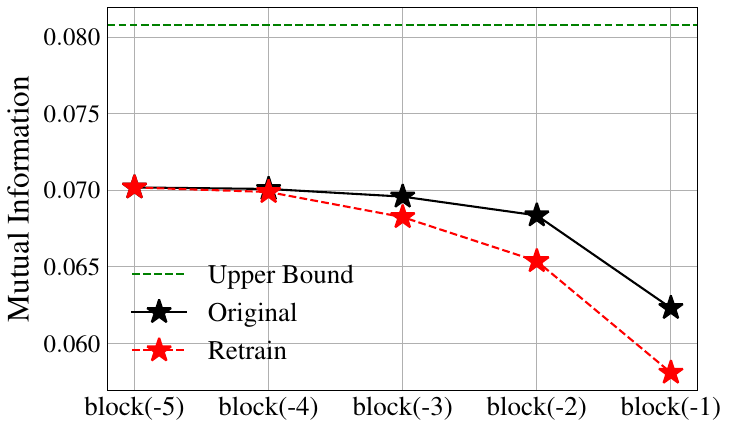}
        \caption{CIFAR-100 - ViT}
        \label{fig:figure6}
    \end{subfigure}
    
    \caption{Mutual information curves across various datasets and model architectures. It illustrates the estimated mutual information $I(\mathbf{Z}_{\ell};Y)$ of the features from the $\ell$-th layer $\mathbf{Z}_{\ell}$ and the binary label $Y$, computed by the InfoNCE loss. `block(-k)' means the k block front from the last layer. }
    \label{fig:mutual_information_curves}
\end{figure*}

\begin{figure*}[t!]
    \centering
    \includegraphics[width=0.4\textwidth]{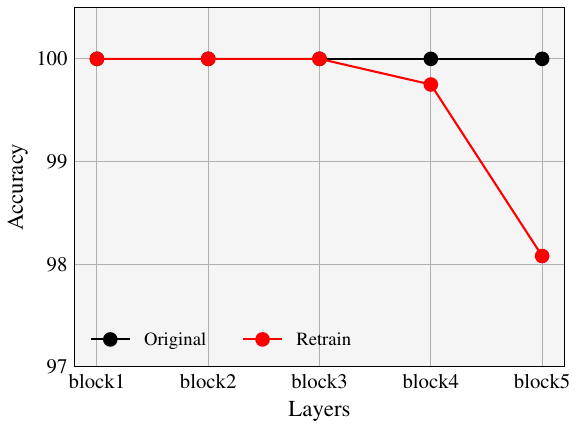}
    \caption{Binary train accuracy on CIFAR-10 in single-class forgetting with retain and forget sets. Interestingly, it shows similar results with mutual information plots shown~\cref{fig:mutual_information_curves}.}
    \label{fig:intermediate_accuracy}
\end{figure*}

Figure~\ref{fig:intermediate_accuracy} shows the train accuracy curves on the CIFAR-10 single-class forgetting dataset with ResNet-18. 
For Original encoder, the trained model readily classifies the retain and forget sets. 
However, for Retrain encoder, the model fails to classifies all samples at the last two layers, with the accuracy dropping more in the later layer. 
These curves correspond to the those from Figure~\ref{fig:mutual_information_curves}, indicating that the estimated MI accurately reflects the model's knowledge of the retain and forget sets.
In addition, the small accuracy gap between Original and Retrain provides the necessity of MI for accurate residual information quantification.

\subsection{Computational Complexity of IDI}
\begin{figure}[htbp]
\vskip -.17in
    \centering
    \includegraphics[width=0.82\linewidth]{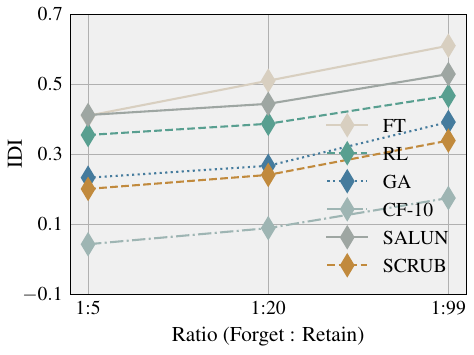}
    \centering
    \vskip -.8em
    \caption{
    IDI of six methods with varying binary label ratios, in the single-class forgetting on (CIFAR-100, ResNet-18), where \(a:b\) denotes the ratio of forget to retain samples.
    }
    \label{fig:exp__ratio}
\vskip -.1in
\end{figure}
\label{computational_complexity_of_idi}
\Cref{tab:mi_runtime_cifar100} presents the runtime of mutual information (MI) computation for intermediate features from each block, 
using the MI estimation method proposed in~\cref{subsection_MI_estimation}, in the CIFAR-100 single-class forgetting setup with ResNet-18, ResNet-50, and ViT. 
Although MI estimation across all layers can be time-consuming, our selected layers for IDI computation (\ie, features from the last two blocks for ResNet-18 and the last three blocks for ResNet-50 and ViT, as detailed in~\cref{app: IDI implementation}, and empirically justified in~\cref{Effect_of_number_of_Layers_for_IDI}) significantly reduce runtime without harming metric performance.
Specifically, the runtime decreases by factors of 2.63, 2.18, and 4.30 for ResNet-18, ResNet-50, and ViT, respectively, when the estimation is sequentially processed for each block.

Furthermore, using only 10\% of the retain set improves runtime by an additional 4 to 5 times without affecting the general trend, as shown in~\cref{fig:exp__ratio}. 
Since training the latter layers requires fewer FLOPs compared to earlier layers, the computational complexity of IDI is further reduced. 
These techniques can effectively alleviate potential computational challenges when applying our metric in practice.

\section{Broader Impact}
Our work on improving machine unlearning focuses on foundational research aimed at enhancing privacy and data removal. However, there is a potential risk that our methodology could be misused to evade data retention policies or obscure accountability. Despite this possibility, it is unlikely that our work will introduce new harmful practices beyond what existing unlearning methods already permit, as we are not introducing new capabilities. Therefore, while there might be concerns related to privacy, security, and fairness, our work does not pose a greater risk compared to other foundational research in machine unlearning.
\section{Limitations}
Our methodology accomplishes its main objective, but there are a few limitations we point out.
Although our IDI successfully investigates hidden information in intermediate features,
its computation requires multiple training runs, which can be computationally intensive.
For instance, The computation of IDI for ResNet-50 on the CIFAR-100 dataset takes approximately 40-50 minutes.
However, one can mitigate this by computing mutual information for only the last few layers,
as the early stages of the encoder are largely similar for both the Retrain and Original models.
Thus, this approach requires fine-tuning only the later layers, reducing the overall computational burden.
Additionally, by adjusting the forget-to-retain ratio, it is possible to improve efficiency and possibly decrease the processing time to merely 3-4 minutes.
\newpage
\begin{table*}[t!]
    
    \centering
    \resizebox{\textwidth}{!}{%
    \begin{tabular}{c|c|c|c}
        \toprule[1pt]
        \midrule
        \multicolumn{4}{c}{\textbf{Class-wise Forgetting}} \\
        \midrule
        \multirow{2}{*}{Settings} & CIFAR-10 & CIFAR-100 & ImageNet-1K \\
        & \multicolumn{2}{c|}{ResNet-18 / ResNet-50 / ViT} & ResNet-50 / ViT \\
        \midrule
        \multirow{3}{*}{FT}
        & \multicolumn{2}{c|}{25 Epochs, Adam} & 3/4 Epochs, Adam \\
        & \multicolumn{2}{c|}{LR $10^{-5}$/$10^{-5}$/$10^{-4}$} & LR $10^{-5}$ \\
        & \multicolumn{2}{c|}{Retain Batch Size 64} & Retain Batch Size 128
        \\
        \midrule
        \multirow{4}{*}{RL}
        & 7 / 7 / 10 Epochs, SGD & 7 / 7 / 10 Epochs, SGD & 3 Epochs, SGD \\
        & LR $10^{-5}$ / $2\cdot10^{-5}$ / $10^{-3}$ & LR $2\cdot10^{-5}$ / $10^{-4}$ / $10^{-4}$ & LR $10^{-3}$ / $10^{-4}$ \\
        & Retain Batch Size 64 & Retain Batch Size 64 & Retain Batch Size 128 \\
        & Forget Batch Size 16 & Forget Batch Size 16 & Forget Batch Size 16 \\
        \midrule
        \multirow{4}{*}{GA}
        & 10 Epochs, SGD & 10 Epochs, SGD & 3 Epochs, SGD \\
        & LR $2\cdot10^{-3}$ / $2\cdot10^{-3}$ / $5\cdot10^{-3}$ & LR $9\cdot10^{-4}$ / $9\cdot10^{-4}$ / $5\cdot10^{-3}$ & LR $2\cdot10^{-3}$ / $10^{-3}$ \\
        & Retain Batch Size 64 & Retain Batch Size 64 & Retain Batch Size 128 \\
        & Forget Batch Size 16 & Forget Batch Size 16 & Forget Batch Size 16 \\
        \midrule
        \multirow{3}{*}{Bad-T}
        & \multicolumn{2}{c|}{10 Epochs, Adam} & 3 Epochs, Adam \\
        & \multicolumn{2}{c|}{LR $10^{-5}$} & LR $10^{-5}$ \\
        & \multicolumn{2}{c|}{Batch Size 256} & Batch Size 256 \\
        \midrule
        \multirow{3}{*}{\begin{tabular}{@{}c@{}}EU-5 / \\EU-10\end{tabular}}
        & \multicolumn{2}{c|}{14 Epochs, SGD} & 2 Epochs, SGD \\
        & \multicolumn{2}{c|}{LR $10^{-2}$} & LR $5\cdot10^{-3}$ \\
        & \multicolumn{2}{c|}{Retain Batch Size 64} & Retain Batch Size 128 \\
        \midrule
        \multirow{3}{*}{\begin{tabular}{@{}c@{}}CF-5 / \\CF-10\end{tabular}}
        & \multicolumn{2}{c|}{14 / 14 / 18 Epochs, SGD} & 5 Epochs, SGD \\
        & \multicolumn{2}{c|}{LR $10^{-2}$ / $10^{-2}$ / $3\cdot10^{-2}$} & LR $5\cdot10^{-3}$ \\
        & \multicolumn{2}{c|}{Retain Batch Size 64} & Retain Batch Size 128 \\
        \midrule
        \multirow{4}{*}{SCRUB}
        & 3(2) Epochs, SGD & 3(2) Epochs, SGD & 2(2) Epochs, SGD \\
        & LR $5\cdot10^{-4}$ / $5\cdot10^{-4}$ / $10^{-4}$ & LR $5\cdot10^{-4}$ & LR $5\cdot10^{-4}$ / $10^{-4}$ \\
        & Retain Batch Size 64 & Retain Batch Size 128 & Retain Batch Size 128 \\
        & Forget Batch Size 256 / 256 / 64 & Forget Batch Size 8 & Forget Batch Size 256 \\
        \midrule
        \multirow{4}{*}{SALUN}
        & 10 Epochs, SGD & 15 Epochs, SGD & 5/2 Epochs, SGD \\
        & LR $5\cdot10^{-4}$ / $10^{-3}$ / $10^{-3}$ & LR $10^{-3}$ & LR $10^{-3}$ \\
        & Retain Batch Size 64 & Retain Batch Size 64 & Retain Batch Size 128 \\
        & Forget Batch Size 16 & Forget Batch Size 16 & Forget Batch Size 16 \\
        \midrule
        \multirow{3}{*}{\(\mathbf{\ell_1}\)-sparse}
        & 10 Epochs, SGD & 10 Epochs, SGD & 5 Epochs, SGD \\
        & LR $2\cdot10^{-4}$ / $2\cdot10^{-4}$ / $9\cdot10^{-4}$ & LR $2\cdot10^{-4}$ / $2\cdot10^{-4}$ / $5\cdot10^{-4}$ & LR $9\cdot10^{-4}$ \\
        & Retain Batch Size 64 & Retain Batch Size 64 & Retain Batch Size 128 \\
        \midrule
        \multirow{4}{*}{COLA}
        & 10+10 Epochs, Adam & 10+10 Epochs, Adam & 1+2 Epochs, Adam \\
        & {\tiny Contrast} LR $2\cdot10^{-4}$ / $2\cdot10^{-4}$ / $1.5\cdot10^{-4}$ & {\tiny Contrast} LR $5\cdot10^{-4}$ / $5\cdot10^{-4}$ / $5\cdot10^{-4}$ & {\tiny Contrast} LR $2\cdot10^{-5}$ / $5\cdot10^{-5}$ \\
        & {\tiny Finetune} LR $5\cdot10^{-6}$ / $10^{-5}$ / $5\cdot10^{-5}$ & {\tiny Finetune} LR $5\cdot10^{-6}$ / $10^{-5}$ / $5\cdot10^{-5}$ & {\tiny Finetune} LR $1\cdot10^{-5}$ / $5\cdot10^{-5}$ \\
        & Retain Batch Size 64 & Retain Batch Size 256 & Retain Batch Size 256 \\
        \midrule
        \bottomrule[1pt]

    \end{tabular}
}
\caption{Hyperparameters of baselines for \textit{class-wise forgetting}. Retain Batch Size is the batch size of retain set \(\mathcal{D}_r\) and Forget Batch Size is the batch size of forget set \(\mathcal{D}_f\). Baselines without Forget Batch Size imply that they do not use forget set \(\mathcal{D}_f\). Bad-T uses the entire dataset \(\mathcal{D}\), so there is no separation of retain and forget of Batch Size. SCRUB has separate epochs for retain set and forget set, which is visualized as Retain Epochs (Forget Epochs). For COLA, A + B Epochs indicates collapse epochs A and align epochs B.}
\label{tab:hyperparam_class}
\end{table*}
\newpage
\begin{table*}[t!]
    
    \centering
    \begin{tabular}{c|c|c}
        \toprule[1pt]
        \midrule
        \multicolumn{3}{c}{\textbf{Random Data Forgetting}} \\
        \midrule
        \multirow{2}{*}{Settings} & CIFAR-10 & CIFAR-100  \\
        & \multicolumn{2}{c}{ResNet-18}  \\
        \midrule
        \multirow{3}{*}{FT}
        & \multicolumn{2}{c}{25 Epochs, Adam} \\
        & LR $10^{-4}$ & LR $2\cdot10^{-4}$ \\
        & \multicolumn{2}{c}{Retain Batch Size 64}
        \\
        \midrule
        \multirow{4}{*}{RL}
        & \multicolumn{2}{c}{7 Epochs, SGD} \\
        & LR $10^{-3}$ & LR $5\cdot10^{-4}$\\
        & \multicolumn{2}{c}{Retain Batch Size 64} \\
        & \multicolumn{2}{c}{Forget Batch Size 16} \\
        \midrule
        \multirow{4}{*}{GA}
        & 10 Epochs, SGD & 10 Epochs, SGD \\
        & LR $2.5\cdot10^{-3}$ & LR $1\cdot10^{-3}$ \\
        & \multicolumn{2}{c}{Retain Batch Size 64} \\
        & \multicolumn{2}{c}{Forget Batch Size 16} \\
        \midrule
        \multirow{3}{*}{Bad-T}
        & \multicolumn{2}{c}{10 Epochs, Adam} \\
        & \multicolumn{2}{c}{LR $1\cdot10^{-5}$}  \\
        & \multicolumn{2}{c}{Batch Size 256} \\
        \midrule
        \multirow{3}{*}{\begin{tabular}{@{}c@{}}EU-5 / \\EU-10\end{tabular}}
        & \multicolumn{2}{c}{14 Epochs, SGD} \\
        & LR $10^{-1}$ & LR $5\cdot10^{-2}$ \\
        & \multicolumn{2}{c}{Retain Batch Size 64} \\
        \midrule
        \multirow{3}{*}{\begin{tabular}{@{}c@{}}CF-5 / \\CF-10\end{tabular}}
        & \multicolumn{2}{c}{14 Epochs, SGD} \\
        & LR $10^{-1}$ & LR $5\cdot10^{-2}$  \\
        & \multicolumn{2}{c}{Retain Batch Size 64} \\
        \midrule
        \multirow{4}{*}{SCRUB}
        & \multicolumn{2}{c}{5(5) Epochs, SGD} \\
        & LR $2.5\cdot10^{-5}$ & LR $5.4\cdot10^{-4}$ \\
        & \multicolumn{2}{c}{Retain Batch Size 16} \\
        & \multicolumn{2}{c}{Forget Batch Size 64} \\
        \midrule
        \multirow{4}{*}{SALUN}
        & 10 Epochs, SGD & 15 Epochs, SGD \\
        & LR $8.3\cdot10^{-4}$ & LR $5\cdot10^{-4}$ \\
        & \multicolumn{2}{c}{Retain Batch Size 64} \\
        & \multicolumn{2}{c}{Forget Batch Size 16} \\
        \midrule
        \multirow{3}{*}{\(\mathbf{\ell_1}\)-sparse}
        & \multicolumn{2}{c}{10 Epochs, SGD} \\
        & LR $4\cdot10^{-4}$ & LR $3\cdot10^{-4}$ \\
        & Retain Batch Size 64 & Retain Batch Size 64 \\
        \midrule
        \multirow{5}{*}{\begin{tabular}{@{}c@{}} COLA+\end{tabular}}
        & 10+10 Epochs, Adam & 10+10 Epochs, Adam \\
        & {\tiny Contrast} LR $2\cdot10^{-4}$ & {\tiny Contrast} LR $2.5\cdot10^{-4}$ \\
        & {\tiny Finetune} LR $1\cdot10^{-4}$ & {\tiny Finetune} LR $2\cdot10^{-5}$ \\
        & Retain Batch Size 32 & Retain Batch Size 64 \\
        & Forget Batch Size 64 & Forget Batch Size 192\\
        \midrule
        \bottomrule[1pt]

    \end{tabular}
\caption{Hyperparameters of baselines for \textit{random data forgetting}. Retain Batch Size is the batch size of retain set \(\mathcal{D}_r\) and Forget Batch Size is the batch size of forget set \(\mathcal{D}_f\). Baselines without Forget Batch Size imply that they do not use forget set \(\mathcal{D}_f\). Bad-T uses the entire dataset \(\mathcal{D}\), so there is no separation of retain and forget of Batch Size. SCRUB uses separate epochs for retain set and forget set, which is visualized as Retain Epochs (Forget Epochs). For COLA+, A + B Epochs indicates collapse epochs A and align epochs B.}
\label{tab:hyperparam_sample}
\end{table*}
\newpage
\begin{table*}[t!]
    
    \centering
    \resizebox{0.82\textwidth}{!}{%
    \begin{tabular}{lcccccccccccc}
        \toprule[1pt]
        \midrule
        \multicolumn{8}{c}{\textbf{CIFAR-10 - ResNet-18}} \\
        \midrule
        Methods & UA & RA & TA & MIA & JSD & \textbf{IDI} & RTE (min)\\
        \midrule
        Original
        & $0.0$
        & $100.0$
        & $95.46$
        & $91.50$
        & $3.21$
        & $1.000$
        & $170.32$
        \\ 
        Retrain
        & $100.0$
        & $100.0$
        & $95.64$
        & $10.64$
        & $0.0$
        & $0.0$
        & $154.56$
        \\ \midrule
        FT
        & $\textbf{100.0}_{\pm 0.0}$
        & $\textbf{100.0}_{\pm 0.0}$
        & $95.12_{\pm 0.09}$
        & $0.17_{\pm 0.05}$
        & $0.57_{\pm 0.03}$ 
        & $0.671_{\pm 0.008}$ 
        & $6.44_{\pm 0.07}$\\
        RL
        & $99.93_{\pm 0.01}$ 
        & $\textbf{100.0}_{\pm 0.0}$
        & $\textbf{95.66}_{\pm 0.09}$ 
        & $0.0_{\pm 0.0}$
        & $0.79_{\pm 0.01}$ 
        & $0.830_{\pm 0.005}$
        & $3.09_{\pm 0.03}$\\
        GA
        & $\textbf{100.0}_{\pm 0.0}$
        & $99.06_{\pm 0.25}$
        & $93.10_{\pm 0.50}$ 
        & $25.37_{\pm 3.24}$
        & $0.59_{\pm 0.05}$
        & $0.334_{\pm 0.014}$
        & $4.00_{\pm 0.08}$\\ 
        Bad-T
        & $99.90_{\pm 0.14}$ 
        & $\underline{99.99}_{\pm 0.0}$
        & $94.99_{\pm 0.12}$ 
        & $68.17_{\pm 42.80}$
        & $3.69_{\pm 0.85}$
        & $1.014_{\pm 0.004}$ 
        & $4.64_{\pm 0.05}$\\ 
        EU-5
        & $\textbf{100.0}_{\pm 0.0}$ 
        & $\textbf{100.0}_{\pm 0.0}$
        & $95.25_{\pm 0.02}$ 
        & $0.06_{\pm 0.03}$
        & $0.53_{\pm 0.02}$ 
        & $0.528_{\pm 0.005}$
        & ${1.54}_{\pm 0.00}$\\ 
        CF-5
        & $98.13_{\pm 1.39}$ 
        & $\textbf{100.0}_{\pm 0.0}$
        & $\underline{95.54}_{\pm 0.09}$ 
        & $0.0_{\pm 0.0}$
        & $0.56_{\pm 0.04}$
        & $0.675_{\pm 0.027}$
        & ${1.57}_{\pm 0.03}$\\
        EU-10
        & $\textbf{100.0}_{\pm 0.0}$
        & $99.50_{\pm 0.02}$
        & $93.61_{\pm 0.08}$
        & $15.24_{\pm 1.08}$
        & $\textbf{0.40}_{\pm 0.01}$
        & $-0.349_{\pm 0.019}$
        & $2.42_{\pm 0.11}$\\ 
        CF-10
        & $\textbf{100.0}_{\pm 0.0}$
        & $99.98_{\pm 0.0}$
        & $94.95_{\pm 0.05}$
        & $\textbf{11.61}_{\pm 0.91}$
        & $\underline{0.41}_{\pm 0.01}$
        & $-0.060_{\pm 0.017}$ 
        & $2.31_{\pm 0.03}$\\ 
        SCRUB
        & $\textbf{100.0}_{\pm 0.0}$
        & $\textbf{100.0}_{\pm 0.0}$
        & $95.37_{\pm 0.04}$
        & $19.73_{\pm 1.92}$
        & $0.47_{\pm 0.01}$
        & $\underline{-0.056}_{\pm 0.008}$
        & $3.49_{\pm 0.02}$\\ 
        SALUN
        & $\underline{99.99}_{\pm 0.01}$ 
        & $\textbf{100.0}_{\pm 0.0}$
        & $95.42_{\pm 0.12}$
        & $0.01_{\pm 0.01}$
        & $0.73_{\pm 0.04}$
        & $0.936_{\pm 0.012}$
        & $3.54_{\pm 0.11}$\\ 
        \(\mathbf{\ell_1}\)-sparse
        & $\textbf{100.0}_{\pm 0.0}$
        & $99.93_{\pm 0.02}$
        & $94.90_{\pm 0.10}$
        & $1.56_{\pm 0.09}$
        & $0.47_{\pm 0.03}$
        & $0.293_{\pm 0.012}$ 
        & $2.96_{\pm 0.03}$\\ 
        \midrule
        \textbf{COLA}
        & $\textbf{100.0}_{\pm 0.0}$
        & $\textbf{100.0}_{\pm 0.00}$
        & $95.36_{\pm 0.06}$
        & $\underline{12.64}_{\pm 0.92}$
        & $0.44_{\pm 0.04}$
        & $\textbf{0.010}_{\pm 0.006}$
        & $4.91_{\pm 0.04}$\\
%%%%%%%%%%%%%%%%%%%%%%%%%%%%%%%%%%%%%%%%%%%
        \toprule[1pt]
        \midrule
        \multicolumn{8}{c}{\textbf{CIFAR-10 - ResNet-50}} \\
        \midrule
        Methods & UA & RA & TA & MIA & JSD & \textbf{IDI} & RTE (min)\\
        \midrule
        Original
        & $0.0$
        & $100.0$
        & $95.42$
        & $95.58$
        & $4.11$
        & $1.000$
        & $341.86$
        \\ 
        Retrain
        & $100.0$
        & $100.0$
        & $95.49$
        & $14.92$
        & $0.0$
        & $0.0$
        & $312.24$
        \\ \midrule
        FT
        & $\textbf{100.0}_{\pm 0.0}$
        & $\underline{99.99}_{\pm 0.0}$
        & $95.28_{\pm 0.11}$
        & $2.17_{\pm 1.28}$
        & $0.73_{\pm 0.02}$
        & $0.607_{\pm 0.009}$
        & $14.50_{\pm 0.34}$\\
        RL
        & $\textbf{100.0}_{\pm 0.0}$
        & $\textbf{100.0}_{\pm 0.0}$
        & $95.56_{\pm 0.03}$
        & \textbf{$0.0_{\pm 0.0}$}
        & $0.99_{\pm 0.02}$
        & $0.804_{\pm 0.006}$
        & $6.26_{\pm 0.04}$\\
        GA
        & $\textbf{100.0}_{\pm 0.0}$
        & $98.06_{\pm 0.34}$
        & $92.07_{\pm 0.63}$
        & $20.56_{\pm 3.87}$
        & $0.66_{\pm 0.06}$
        & $0.334_{\pm 0.023}$
        & $8.69_{\pm 0.03}$\\ 
        Bad-T
        & $\textbf{100.0}_{\pm 0.0}$
        & $99.94_{\pm 0.04}$
        & $94.74_{\pm 0.24}$
        & $49.95_{\pm 40.74}$
        & $3.02_{\pm 0.64}$
        & $1.153_{\pm 0.026}$
        & $10.19_{\pm 0.32}$\\ 
        EU-5
        & $\textbf{100.0}_{\pm 0.0}$
        & $\textbf{100.0}_{\pm 0.0}$
        & $95.59_{\pm 0.08}$
        & $0.0_{\pm 0.0}$
        & $0.78_{\pm 0.08}$
        & $1.047_{\pm 0.005}$
        & $\underline{4.86}_{\pm 0.43}$\\ 
        CF-5
        & $\underline{17.84}_{\pm 0.93}$
        & $\textbf{100.0}_{\pm 0.0}$
        & $95.64_{\pm 0.11}$
        & $0.0_{\pm 0.0}$
        & $1.43_{\pm 0.04}$
        & $0.906_{\pm 0.002}$
        & $\textbf{4.84}_{\pm 0.10}$\\ 
        EU-10
        & $\textbf{100.0}_{\pm 0.0}$
        & $\textbf{100.0}_{\pm 0.0}$
        & $\underline{95.51}_{\pm 0.12}$
        & $0.17_{\pm 0.05}$
        & $0.65_{\pm 0.02}$
        & $0.757_{\pm 0.011}$
        & $6.92_{\pm 0.02}$\\ 
        CF-10
        & $\textbf{100.0}_{\pm 0.0}$
        & $\textbf{100.0}_{\pm 0.0}$
        & $\textbf{95.49}_{\pm 0.13}$
        & $0.07_{\pm 0.03}$
        & $0.67_{\pm 0.08}$
        & $0.579_{\pm 0.009}$
        & $7.09_{\pm 0.02}$\\ 
        SCRUB
        & $\textbf{100.0}_{\pm 0.0}$
        & $\textbf{100.0}_{\pm 0.0}$
        & $95.23_{\pm 0.20}$
        & $\underline{18.19}_{\pm 0.10}$
        & $0.59_{\pm 0.01}$
        & $\underline{0.067}_{\pm 0.020}$
        & $8.69_{\pm 0.03}$\\ 
        SALUN
        & $\textbf{100.0}_{\pm 0.0}$
        & $99.67_{\pm 0.17}$
        & $93.90_{\pm 0.48}$
        & $1.58_{\pm 0.98}$
        & $0.67_{\pm 0.03}$
        & $0.832_{\pm 0.027}$
        & $11.00_{\pm 0.06}$\\ 
        \(\mathbf{\ell_1}\)-sparse
        & $\textbf{100.0}_{\pm 0.0}$
        & $99.88_{\pm 0.06}$
        & $94.49_{\pm 0.29}$
        & $4.06_{\pm 0.91}$
        & $\textbf{0.47}_{\pm 0.01}$
        & $0.184_{\pm 0.023}$
        & $12.33_{\pm 0.04}$\\ 
        \midrule
        \textbf{COLA}
        & $\textbf{100.0}_{\pm 0.0}$
        & $\underline{99.99}_{\pm 0.0}$
        & $95.45_{\pm 0.05}$
        & $\textbf{13.69}_{\pm 0.84}$
        & $\underline{0.52}_{\pm 0.02}$
        & $\textbf{0.019}_{\pm 0.025}$
        & $11.98_{\pm 0.03}$\\
%%%%%%%%%%%%%%%%%%%%%%%%%%%%%%%%%%%%%%%%%%%%%%%%%%%%%%%%%%
        \toprule[1pt]
        \midrule
        \multicolumn{8}{c}{\textbf{CIFAR-10 - ViT}} \\
        \midrule
        Methods & UA & RA & TA & MIA & JSD & \textbf{IDI} & RTE (min)\\
        \midrule
        Original
        & $0.36$
        & $99.55$
        & $98.40$
        & $89.12$
        & $3.96$
        & $1.000$
        & $100.68^\star$
        \\ 
        Retrain
        & $100.0$
        & $99.40$
        & $97.96$
        & $4.96$
        & $0.0$
        & $0.0$
        & $90.96^\star$
        \\ \midrule
        FT
        & $98.10_{\pm 0.24}$
        & $99.85_{\pm 0.06}$
        & $\textbf{97.58}_{\pm 0.36}$
        & $21.14_{\pm 0.92}$
        & $0.71_{\pm 0.13}$
        & $-0.871_{\pm 0.141}$
        & $130.13_{\pm 0.63}$\\
        RL
        & $97.88_{\pm 2.12}$
        & $99.88_{\pm 0.01}$
        & $99.01_{\pm 0.02}$
        & $0.0_{\pm 0.0}$
        & $0.74_{\pm 0.04}$
        & $1.052_{\pm 0.011}$
        & $65.45_{\pm 0.12}$\\
        GA
        & $\textbf{100.0}_{\pm 0.0}$
        & $99.80_{\pm 0.03}$
        & $98.49_{\pm 0.12}$
        & $\textbf{4.82}_{\pm 0.98}$
        & $0.39_{\pm 0.05}$
        & $0.498_{\pm 0.025}$
        & $68.32_{\pm 0.80}$\\ 
        Bad-T
        & $\textbf{100.0}_{\pm 0.0}$
        & $\textbf{99.55}_{\pm 0.03}$
        & $\underline{98.40}_{\pm 0.20}$
        & $0.0_{\pm 0.0}$
        & $0.84_{\pm 0.06}$
        & $0.997_{\pm 0.016}$
        & $100.90_{\pm 1.02}$\\ 
        EU-5
        & $\textbf{100.0}_{\pm 0.0}$
        & $99.76_{\pm 0.01}$
        & $98.80_{\pm 0.01}$
        & $0.30_{\pm 0.01}$
        & $0.28_{\pm 0.03}$
        & $0.901_{\pm 0.006}$
        & $\underline{29.89}_{\pm 0.09}$\\ 
        CF-5
        & $\textbf{100.0}_{\pm 0.0}$
        & $99.76_{\pm 0.0}$
        & $98.86_{\pm 0.02}$
        & $0.35_{\pm 0.03}$
        & $0.26_{\pm 0.01}$
        & $0.941_{\pm 0.001}$
        & $34.12_{\pm 0.09}$\\ 
        EU-10
        & $\textbf{100.0}_{\pm 0.0}$
        & $99.72_{\pm 0.02}$
        & $98.63_{\pm 0.04}$
        & $0.64_{\pm 0.02}$
        & $\underline{0.23}_{\pm 0.03}$
        & $\underline{0.268}_{\pm 0.016}$
        & $32.74_{\pm 0.19}$\\ 
        CF-10
        & $\textbf{100.0}_{\pm 0.0}$
        & $99.77_{\pm 0.01}$
        & $98.75_{\pm 0.02}$
        & $0.64_{\pm 0.04}$
        & $\textbf{0.21}_{\pm 0.02}$
        & $0.377_{\pm 0.039}$
        & $36.79_{\pm 0.15}$\\ 
        SCRUB
        & $\textbf{100.0}_{\pm 0.0}$ 
        & $\underline{99.66}_{\pm 0.0}$
        & $98.57_{\pm 0.01}$
        & $94.74_{\pm 0.26}$
        & $3.87_{\pm 0.07}$
        & $0.907_{\pm 0.027}$
        & $\textbf{22.99}_{\pm 0.24}$\\ 
        SALUN
        & $\textbf{100.0}_{\pm 0.0}$
        & $99.78_{\pm 0.02}$
        & $98.89_{\pm 0.02}$
        & $0.01_{\pm 0.01}$
        & $0.39_{\pm 0.05}$
        & $1.066_{\pm 0.041}$
        & $61.37_{\pm 0.10}$\\ 
        \(\mathbf{\ell_1}\)-sparse
        & $\textbf{100.0}_{\pm 0.0}$
        & $97.48_{\pm 0.27}$
        & $95.78_{\pm 0.16}$
        & $\underline{3.89}_{\pm 0.79}$
        & $0.41_{\pm 0.03}$
        & $-0.573_{\pm 0.290}$
        & $51.44_{\pm 0.04}$\\ 
        \midrule
        \textbf{COLA}
        & $\underline{99.44}_{\pm 0.02}$
        & $100.0_{\pm 0.0}$
        & $98.82_{\pm 0.06}$
        & $11.90_{\pm 1.36}$
        & $0.63_{\pm 0.11}$
        & $\textbf{-0.067}_{\pm 0.010}$
        & \textbf{$116.01_{\pm 0.96}$}\\ 
        \midrule
        \bottomrule[1pt]
    \end{tabular}
    }
\caption{Single-class forgetting result on CIFAR-10 dataset across different model architectures. A better performance of an MU method corresponds to a smaller performance gap with Retrain (except RTE), with the top method in \textbf{bold} and the second best \underline{underlined}. The \(\star\) symbol indicated in RTE of Original and Retrain means that models are pretrained on ImageNet-21K and then finetuned on CIFAR-10, with the reported time reflecting only the finetuning process. In contrast, Original and Retrain without \(\star\) are trained from scratch on CIFAR-10.}
\label{tab:single_class_cifar10}
\end{table*}
\newpage
%%%%%%%%%%%%%%%%%%%%%%%%%%%%%%%%%%%%%%%%%%%
\begin{table*}[t!]
    
    \centering
    \resizebox{0.82\textwidth}{!}{%
    \begin{tabular}{lcccccccccccc}
        \toprule[1pt]
        \midrule
        \multicolumn{8}{c}{\textbf{CIFAR-100 - ResNet-18}} \\
        \midrule
         Methods & UA & RA & TA & MIA & JSD & \textbf{IDI} & RTE (min)\\
        \midrule
        Original
        & $0.0$
        & $99.98$
        & $78.18$
        & $92.80$
        & $2.91$
        & $1.000$
        & $175.08$
        \\ 
        Retrain
        & $100.0$
        & $99.96$
        & $79.48$
        & $2.00$
        & $0.0$
        & $0.0$
        & $171.27$
        \\ \midrule
        FT
        & $\textbf{100.00}_{\pm 0.0}$
        & $\textbf{99.97}_{\pm 0.0}$
        & $77.49_{\pm 0.14}$
        & $0.07_{\pm 0.09}$
        & $0.37_{\pm 0.01}$
        & $0.610_{\pm 0.022}$
        & $9.50_{\pm 0.03}$\\
        RL
        & $93.80_{\pm 0.75}$
        & $\underline{99.98}_{\pm 0.0}$
        & $\underline{77.94}_{\pm 0.10}$
        & $0.0_{\pm 0.0}$
        & $0.52_{\pm 0.01}$
        & $0.467_{\pm 0.010}$
        & $3.52_{\pm 0.0}$\\
        GA
        & $\underline{99.93}_{\pm 0.09}$
        & $96.87_{\pm 0.52}$
        & $69.87_{\pm 0.78}$
        & $21.40_{\pm 2.04}$
        & $1.18_{\pm 0.02}$
        & $0.392_{\pm 0.021}$
        & $5.32_{\pm 0.01}$\\ 
        Bad-T
        & $\textbf{100.0}_{\pm 0.0}$
        & $\underline{99.98}_{\pm 0.0}$
        & $77.66_{\pm 0.26}$
        & $40.87_{\pm 36.87}$
        & $2.53_{\pm 0.44}$
        & $1.079_{\pm 0.024}$
        & $5.78_{\pm 0.02}$\\ 
        EU-5
        & $\textbf{100.0}_{\pm 0.0}$ 
        & $99.78_{\pm 0.01}$
        & $75.01_{\pm 0.04}$
        & $9.33_{\pm 0.75}$
        & $0.66_{\pm 0.01}$
        & $\underline{0.064}_{\pm 0.037}$
        & ${2.14}_{\pm 0.0}$\\ 
        CF-5
        & $\textbf{100.0}_{\pm 0.0}$
        & $\textbf{99.97}_{\pm 0.0}$
        & $77.30_{\pm 0.28}$
        & $\underline{2.87}_{\pm 0.66}$
        & $0.40_{\pm 0.03}$
        & $0.388_{\pm 0.010}$
        & ${2.14}_{\pm 0.01}$\\ 
        EU-10
        & $\textbf{100.0}_{\pm 0.0}$
        & $91.94_{\pm 0.08}$
        & $72.84_{\pm 0.04}$
        & $12.67_{\pm 0.47}$
        & $0.53_{\pm 0.02}$
        & $-0.221_{\pm 0.009}$
        & $4.39_{\pm 0.02}$\\ 
        CF-10
        & $\textbf{100.0}_{\pm 0.0}$
        & $99.89_{\pm 0.02}$
        & $76.49_{\pm 0.02}$
        & $7.07_{\pm 0.84}$
        & $0.49_{\pm 0.01}$
        & $0.175_{\pm 0.040}$
        & $4.29_{\pm 0.04}$\\ 
        SCRUB
        & $\textbf{100.0}_{\pm 0.0}$
        & $\underline{99.98}_{\pm 0.0}$
        & $\textbf{78.17}_{\pm 0.04}$
        & $0.07_{\pm 0.09}$
        & $\underline{0.31}_{\pm 0.01}$
        & $0.339_{\pm 0.069}$
        & ${2.27}_{\pm 0.02}$\\ 
        SALUN
        & $95.73_{\pm 0.85}$
        & $99.22_{\pm 0.13}$
        & $74.20_{\pm 0.52}$
        & ${0.09}_{\pm 0.02}$
        & $0.65_{\pm 0.01}$
        & $0.529_{\pm 0.022}$
        & $4.63_{\pm 0.06}$\\ 
        \(\mathbf{\ell_1}\)-sparse
        & $96.93_{\pm 0.19}$
        & $98.90_{\pm 0.12}$
        & $74.69_{\pm 0.06}$
        & $6.60_{\pm 0.43}$
        & $0.34_{\pm 0.01}$
        & $0.334_{\pm 0.026}$
        & $4.55_{\pm 0.01}$\\ 
        \midrule
        \textbf{COLA}
        & $\textbf{100.0}_{\pm 0.0}$ 
        & $99.80_{\pm 0.00}$
        & $76.48_{\pm 0.11}$
        & $9.60_{\pm 1.31}$
        & $\textbf{0.26}_{\pm 0.01}$
        & $\textbf{-0.037}_{\pm 0.006}$
        & $7.51_{\pm 0.02}$\\ 
%%%%%%%%%%%%%%%%%%%%%%%%%%%%%%%%%%%%%%%%%%%
        \toprule[1pt]
        \midrule
        \multicolumn{8}{c}{\textbf{CIFAR-100 - ResNet-50}} \\
        \midrule
        Methods & UA & RA & TA & MIA & JSD & \textbf{IDI} & RTE (min)\\
        \midrule
        Original
        & $0.0$
        & $99.98$
        & $79.84$
        & $91.60$
        & $3.43$
        & $1.000$
        & $345.54$
        \\ 
        Retrain
        & $100.0$
        & $99.97$
        & $79.42$
        & $3.40$
        & $0.0$
        & $0.0$
        & $338.58$
        \\ \midrule
        FT
        & $99.33_{\pm 0.09}$
        & $99.93_{\pm 0.03}$
        & $77.71_{\pm 0.18}$
        & $0.40_{\pm 0.16}$
        & $0.57_{\pm 0.02}$
        & $0.618_{\pm 0.018}$
        & $16.34_{\pm 0.47}$\\
        RL
        & $\textbf{100.0}_{\pm 0.0}$
        & $99.95_{\pm 0.02}$
        & $\underline{79.56}_{\pm 0.04}$
        & $0.0_{\pm 0.0}$
        & $0.80_{\pm 0.0}$
        & $0.649_{\pm 0.013}$
        & $8.38_{\pm 0.14}$\\
        GA
        & $99.60_{\pm 0.43}$
        & $98.00_{\pm 0.72}$
        & $72.73_{\pm 1.16}$
        & $13.33_{\pm 4.43}$
        & $0.99_{\pm 0.04}$
        & $0.526_{\pm 0.009}$
        & $9.50_{\pm 0.54}$\\ 
        Bad-T
        & $\textbf{100.0}_{\pm 0.0}$
        & $99.90_{\pm 0.10}$
        & $77.53_{\pm 1.21}$
        & $94.80_{\pm 2.75}$
        & $3.98_{\pm 0.25}$
        & $0.990_{\pm 0.033}$
        & $12.69_{\pm 1.54}$\\ 
        EU-5
        & $\textbf{100.0}_{\pm 0.0}$
        & $\textbf{99.97}_{\pm 0.01}$
        & $78.31_{\pm 0.21}$
        & $\underline{1.20}_{\pm 0.99}$
        & $0.61_{\pm 0.04}$
        & $0.520_{\pm 0.023}$
        & $\underline{6.81}_{\pm 0.01}$\\ 
        CF-5
        & $\textbf{100.0}_{\pm 0.0}$
        & $\textbf{99.97}_{\pm 0.01}$
        & $78.98_{\pm 0.16}$
        & $0.27_{\pm 0.09}$
        & $0.50_{\pm 0.02}$
        & $0.575_{\pm 0.016}$
        & $6.82_{\pm 0.01}$\\ 
        EU-10
        & $\textbf{100.0}_{\pm 0.0}$
        & $98.52_{\pm 0.14}$
        & $75.66_{\pm 0.03}$
        & $15.00_{\pm 1.45}$
        & $0.69_{\pm 0.01}$
        & $\underline{0.050}_{\pm 0.004}$
        & $7.81_{\pm 0.01}$\\ 
        CF-10
        & $\textbf{100.0}_{\pm 0.0}$
        & $99.95_{\pm 0.01}$
        & $78.47_{\pm 0.10}$
        & $5.87_{\pm 0.09}$
        & $0.50_{\pm 0.02}$
        & $0.302_{\pm 0.035}$
        & $7.82_{\pm 0.02}$\\ 
        SCRUB
        & $\textbf{100.0}_{\pm 0.0}$
        & $\textbf{99.97}_{\pm 0.0}$
        & $79.61_{\pm 0.09}$
        & $0.20_{\pm 0.16}$
        & $\underline{0.43}_{\pm 0.02}$
        & $0.620_{\pm 0.034}$
        & $\textbf{4.59}_{\pm 0.13}$\\ 
        SALUN
        & $\underline{99.73}_{\pm 0.38}$
        & $\underline{99.98}_{\pm 0.0}$
        & $\textbf{79.51}_{\pm 0.15}$
        & $0.0_{\pm 0.0}$
        & $0.80_{\pm 0.01}$
        & $0.679_{\pm 0.010}$
        & $12.83_{\pm 0.87}$\\ 
        \(\mathbf{\ell_1}\)-sparse
        & $96.20_{\pm 0.16}$
        & $99.42_{\pm 0.06}$
        & $76.16_{\pm 0.31}$
        & $\textbf{2.60}_{\pm 0.33}$
        & $\underline{0.43}_{\pm 0.01}$
        & $0.325_{\pm 0.018}$
        & $15.78_{\pm 0.05}$\\ 
        \midrule
        \textbf{COLA}
        & $\textbf{100.0}_{\pm 0.0}$ 
        & $99.90_{\pm 0.01}$
        & $78.59_{\pm 0.28}$
        & $10.27_{\pm 0.90}$
        & $\textbf{0.42}_{\pm 0.02}$
        & $\textbf{0.016}_{\pm 0.031}$
        & $16.25_{\pm 0.10}$\\ 
%%%%%%%%%%%%%%%%%%%%%%%%%%%%%%%%%%%%%%%%%%%
        \toprule[1pt]
        \midrule
        \multicolumn{8}{c}{\textbf{CIFAR-100 - ViT}} \\
        \midrule
        Methods & UA & RA & TA & MIA & JSD & \textbf{IDI} & RTE (min)\\
        \midrule
        Original
        & $7.00$
        & $95.85$
        & $90.78$
        & $69.20$
        & $2.71$
        & $1.000$
        & $102.45^\star$
        \\ 
        Retrain
        & $100.0$
        & $95.79$
        & $90.58$
        & $10.00$
        & $0.0$
        & $0.0$ 
        & $94.29^\star$
        \\ \midrule
        FT
        & $\textbf{100.0}_{\pm 0.0}$
        & $99.79_{\pm 0.04}$
        & $88.69_{\pm 0.11}$
        & $14.80_{\pm 2.40}$
        & $0.57_{\pm 0.03}$
        & $-0.934_{\pm 0.011}$
        & $140.61_{\pm 0.25}$\\
        RL
        & $99.19_{\pm 0.23}$
        & $97.11_{\pm 0.02}$
        & $92.28_{\pm 0.06}$
        & $0.31_{\pm 0.01}$
        & $0.82_{\pm 0.01}$
        & $1.091_{\pm 0.031}$
        & $73.12_{\pm 0.18}$\\
        GA
        & $\textbf{100.0}_{\pm 0.0}$
        & $98.19_{\pm 0.20}$
        & $\textbf{90.59}_{\pm 0.21}$
        & $17.60_{\pm 4.78}$
        & $0.31_{\pm 0.01}$
        & $0.587_{\pm 0.011}$
        & $75.22_{\pm 0.61}$\\
        Bad-T
        & $95.80_{\pm 0.08}$
        & $\textbf{95.88}_{\pm 0.12}$
        & $90.15_{\pm 0.02}$
        & $0.0_{\pm 0.0}$
        & $1.11_{\pm 0.14}$
        & $1.213_{\pm 0.002}$
        & $96.43_{\pm 0.01}$\\ 
        EU-5
        & $\textbf{100.0}_{\pm 0.0}$
        & $97.59_{\pm 0.04}$
        & $92.04_{\pm 0.02}$
        & $\underline{7.10}_{\pm 0.70}$
        & $\underline{0.27}_{\pm 0.01}$
        & $1.143_{\pm 0.008}$
        & $\underline{32.17}_{\pm 0.02}$\\ 
        CF-5
        & $\textbf{100.0}_{\pm 0.0}$
        & $97.81_{\pm 0.01}$
        & $91.98_{\pm 0.05}$
        & $6.93_{\pm 0.32}$
        & $\underline{0.27}_{\pm 0.01}$
        & $1.087_{\pm 0.050}$
        & $36.73_{\pm 0.03}$\\ 
        EU-10
        & $\textbf{100.0}_{\pm 0.0}$
        & $97.87_{\pm 0.01}$
        & $91.45_{\pm 0.07}$
        & $13.30_{\pm 1.97}$
        & $0.36_{\pm 0.02}$
        & $0.849_{\pm 0.012}$
        & $34.23_{\pm 0.02}$\\ 
        CF-10
        & $\textbf{100.0}_{\pm 0.0}$
        & $97.87_{\pm 0.01}$
        & $91.61_{\pm 0.05}$
        & $15.80_{\pm 0.80}$
        & $0.32_{\pm 0.02}$
        & $0.734_{\pm 0.011}$
        & $39.12_{\pm 0.0}$\\ 
        SCRUB
        & $\textbf{100.0}_{\pm 0.00}$
        & $96.95_{\pm 0.03}$
        & $92.12_{\pm 0.06}$
        & $17.00_{\pm 1.21}$
        & $\underline{0.27}_{\pm 0.02}$
        & $\underline{0.037}_{\pm 0.036}$
        & $\textbf{17.84}_{\pm 0.13}$\\ 
        SALUN
        & $\underline{99.73}_{\pm 0.31}$
        & $98.32_{\pm 0.04}$
        & $92.23_{\pm 0.05}$
        & $0.47_{\pm 0.06}$
        & $0.78_{\pm 0.02}$
        & $1.123_{\pm 0.043}$
        & $203.12_{\pm 0.51}$\\ 
        \(\mathbf{\ell_1}\)-sparse
        & $\textbf{100.0}_{\pm 0.0}$
        & $\underline{96.37}_{\pm 0.06}$
        & $\underline{90.92}_{\pm 0.07}$
        & $3.80_{\pm 1.62}$
        & $\textbf{0.23}_{\pm 0.01}$
        & $1.144_{\pm 0.002}$
        & $56.93_{\pm 0.32}$\\ 
        \midrule
        \textbf{COLA}
        & $\textbf{100.0}_{\pm 0.0}$
        & $99.76_{\pm 0.02}$
        & $90.23_{\pm 0.04}$
        & $\textbf{12.00}_{\pm 2.20}$
        & $0.54_{\pm 0.01}$
        & $\textbf{-0.022}_{\pm 0.016}$
        & $112.58_{\pm 0.82}$\\ 
        \midrule
        \bottomrule[1pt]
    \end{tabular}
    }
\caption{Single-class forgetting result on CIFAR-100 dataset across different model architectures. A better performance of an MU method corresponds to a smaller performance gap with Retrain (except RTE), with the top method in \textbf{bold} and the second best \underline{underlined}. The \(\star\) symbol indicated in RTE of Original and Retrain means that models are pretrained on ImageNet-21K and then finetuned on CIFAR-100, with the reported time reflecting only the finetuning process. In contrast, Original and Retrain without are \(\star\) trained from scratch on CIFAR-100.
}
\label{tab:single_class_cifar100}
\end{table*}
\newpage
\begin{table*}[t!]
    
    \centering
    \resizebox{0.82\textwidth}{!}{%
    \begin{tabular}{lcccccccccccc}
        \toprule[1pt]
        \midrule
        \multicolumn{8}{c}{\textbf{CIFAR-10 - 2-class forgetting}} \\
        \midrule
        Methods & UA & RA & TA & MIA & JSD & \textbf{IDI} & RTE (min) \\
        \midrule
        Original
        & $0.0$
        & $100.0$
        & $95.76$
        & $91.10$
        & $3.55$
        & $1.000$
        & $170.32$
        \\ 
        Retrain
        & $100.0$
        & $100.0$
        & $96.38$
        & $29.58$
        & $0.0$
        & $0.0$
        & $135.23$
        \\ \midrule
        FT
        & $\underline{99.98}_{\pm 0.01}$ 
        & $\textbf{100.0}_{\pm 0.0}$
        & $\underline{96.36}_{\pm 0.09}$
        & $0.96_{\pm 0.53}$
        & $0.58_{\pm 0.08}$
        & $0.750_{\pm 0.009}$
        & $5.92_{\pm 0.09}$\\
        RL
        & $99.70_{\pm 0.02}$
        & $\textbf{100.0}_{\pm 0.0}$
        & $\textbf{96.39}_{\pm 0.01}$
        & $0.0_{\pm 0.0}$
        & $1.07_{\pm 0.01}$
        & $0.863_{\pm 0.001}$
        & $2.79_{\pm 0.02}$\\
        GA
        & $99.07_{\pm 0.38}$
        & $99.43_{\pm 0.13}$
        & $94.83_{\pm 0.22}$
        & $\underline{26.71}_{\pm 3.68}$
        & $0.42_{\pm 0.02}$
        & $0.612_{\pm 0.001}$
        & $3.72_{\pm 0.13}$\\ 
        Bad-T
        & $99.96_{\pm 0.05}$
        & $\textbf{100.0}_{\pm 0.0}$
        & $95.33_{\pm 0.09}$
        & $67.47_{\pm 34.59}$
        & $3.98_{\pm 1.08}$
        & $1.010_{\pm 0.005}$
        & $4.40_{\pm 0.20}$\\ 
        EU-5
        & $\textbf{100.0}_{\pm 0.0}$
        & $\textbf{100.0}_{\pm 0.0}$
        & $96.48_{\pm 0.06}$
        & $0.06_{\pm 0.03}$
        & $0.57_{\pm 0.05}$
        & $0.624_{\pm 0.001}$
        & $\textbf{1.39}_{\pm 0.02}$\\
        CF-5
        & $80.06_{\pm 8.26}$
        & $\textbf{100.0}_{\pm 0.0}$
        & $96.70_{\pm 0.04}$
        & $0.0_{\pm 0.0}$
        & $0.80_{\pm 0.02}$
        & $0.781_{\pm 0.006}$
        & $\underline{1.41}_{\pm 0.05}$\\ 
        EU-10
        & $\textbf{100.0}_{\pm 0.0}$
        & $99.67_{\pm 0.02}$
        & $94.94_{\pm 0.17}$
        & $25.92_{\pm 0.79}$
        & $\underline{0.35}_{\pm 0.01}$
        & $\textbf{-0.011}_{\pm 0.011}$
        & $2.20_{\pm 0.17}$\\ 
        CF-10
        & $\textbf{100.0}_{\pm 0.0}$
        & $99.67_{\pm 0.02}$
        & $94.94_{\pm 0.17}$
        & $21.20_{\pm 1.43}$
        & $\underline{0.35}_{\pm 0.01}$
        & $\underline{0.221}_{\pm 0.007}$
        & $2.19_{\pm 0.14}$\\ 
        SCRUB
        & $\underline{99.98}_{\pm 0.0}$
        & $\underline{99.99}_{\pm 0.0}$
        & $96.31_{\pm 0.08}$
        & $46.74_{\pm 5.31}$
        & $1.47_{\pm 0.10}$
        & $0.374_{\pm 0.005}$
        & $3.27_{\pm 0.01}$\\ 
        SALUN
        & $95.86_{\pm 4.18}$
        & $\underline{99.99}_{\pm 0.01}$
        & $96.27_{\pm 0.11}$
        & $0.04_{\pm 0.01}$
        & $0.89_{\pm 0.05}$
        & $0.951_{\pm 0.019}$
        & $3.17_{\pm 0.02}$\\ 
        \(\mathbf{\ell_1}\)-sparse
        & $99.91_{\pm 0.05}$
        & $99.98_{\pm 0.0}$
        & $96.47_{\pm 0.09}$
        & $1.57_{\pm 0.11}$
        & $0.50_{\pm 0.02}$
        & $0.560_{\pm 0.004}$
        & $2.62_{\pm 0.06}$\\ 
        \midrule
        \textbf{COLA}
        & $\textbf{100.0}_{\pm 0.0}$ 
        & $99.92_{\pm 0.0}$
        & $96.41_{\pm 0.15}$
        & $\textbf{31.40}_{\pm 2.98}$
        & $\textbf{0.26}_{\pm 0.01}$
        & $\textbf{0.011}_{\pm 0.029}$
        & $4.59_{\pm 0.02}$\\ 
%%%%%%%%%%%%%%%%%%%%%%%%%%%%%%%%%%%%%%%%%%%
\toprule[1pt]
        \midrule
        \multicolumn{8}{c}{\textbf{CIFAR-100 - 5-class forgetting}} \\
        \midrule
        Methods & UA & RA & TA & MIA & JSD & \textbf{IDI} & RTE (min)\\
        \midrule
        Original
        & $0.0$
        & $99.98$
        & $77.95$
        & $95.00$
        & $3.18$
        & $1.000$
        & $175.08$
        \\ 
        Retrain
        & $100.0$
        & $99.98$
        & $78.45$
        & $7.12$
        & $0.0$
        & $0.0$
        & $165.92$
        \\ \midrule
        FT
        & $\textbf{100.00}_{\pm 0.0}$ 
        & $99.93_{\pm 0.06}$
        & $77.43_{\pm 0.20}$
        & $0.20_{\pm 0.06}$
        & $0.38_{\pm 0.01}$
        & $0.596_{\pm 0.009}$
        & $9.21_{\pm 0.06}$\\
        RL
        & $98.61_{\pm 0.22}$
        & $\textbf{99.98}_{\pm 0.0}$
        & $\textbf{77.78}_{\pm 0.19}$
        & \textbf{$0.0_{\pm 0.0}$}
        & $0.71_{\pm 0.01}$
        & $0.613_{\pm 0.008}$
        & $3.39_{\pm 0.09}$\\
        GA
        & $79.99_{\pm 4.75}$
        & $95.18_{\pm 0.40}$
        & $68.68_{\pm 0.52}$
        & $32.25_{\pm 2.02}$
        & $1.36_{\pm 0.06}$
        & $0.236_{\pm 0.010}$
        & $4.99_{\pm 0.04}$\\ 
        Bad-T
        & $\textbf{100.0}_{\pm 0.0}$
        & $\textbf{99.98}_{\pm 0.0}$
        & $75.93_{\pm 0.57}$
        & $44.60_{\pm 31.96}$
        & $2.86_{\pm 0.25}$
        & $1.021_{\pm 0.031}$
        & $5.51_{\pm 0.11}$\\ 
        EU-5
        & $\textbf{100.0}_{\pm 0.0}$
        & $99.75_{\pm 0.02}$
        & $75.14_{\pm 0.12}$
        & $12.40_{\pm 0.26}$
        & $0.54_{\pm 0.01}$
        & $\underline{0.054}_{\pm 0.010}$
        & $\textbf{2.01}_{\pm 0.0}$\\ 
        CF-5
        & $\textbf{100.0}_{\pm 0.0}$
        & $\underline{99.97}_{\pm 0.0}$
        & $77.36_{\pm 0.06}$
        & $\underline{3.37}_{\pm 0.52}$
        & $\underline{0.36}_{\pm 0.02}$
        & $0.319_{\pm 0.011}$
        & $\underline{2.10}_{\pm 0.0}$\\ 
        EU-10
        & $\textbf{100.0}_{\pm 0.0}$ 
        & $91.76_{\pm 0.12}$
        & $73.24_{\pm 0.11}$
        & $21.96_{\pm 0.49}$
        & $0.48_{\pm 0.01}$
        & $-0.155_{\pm 0.008}$
        & $4.25_{\pm 0.0}$\\ 
        CF-10
        & $\textbf{100.0}_{\pm 0.0}$
        & $99.88_{\pm 0.01}$
        & $76.59_{\pm 0.24}$
        & $\textbf{10.69}_{\pm 1.29}$
        & $0.40_{\pm 0.01}$
        & $0.087_{\pm 0.019}$
        & $4.29_{\pm 0.01}$\\ 
        SCRUB
        & $\textbf{100.0}_{\pm 0.0}$
        & $\underline{99.97}_{\pm 0.0}$
        & $\underline{77.64}_{\pm 0.11}$
        & $0.95_{\pm 0.35}$
        & $0.56_{\pm 0.03}$
        & $0.289_{\pm 0.015}$
        & $2.27_{\pm 0.03}$\\ 
        SALUN
        & $\textbf{100.0}_{\pm 0.0}$
        & $99.96_{\pm 0.01}$
        & $77.18_{\pm 0.14}$
        & $0.13_{\pm 0.09}$
        & $0.55_{\pm 0.01}$
        & $0.597_{\pm 0.029}$
        & $4.46_{\pm 0.04}$\\ 
        \(\mathbf{\ell_1}\)-sparse
        & $\underline{98.63}_{\pm 0.37}$
        & $97.50_{\pm 0.14}$
        & $73.46_{\pm 0.25}$
        & $12.35_{\pm 0.82}$
        & $0.38_{\pm 0.01}$
        & $0.196_{\pm 0.011}$
        & $4.19_{\pm 0.01}$\\ 
        \midrule
        \textbf{COLA}
        & $\textbf{100.0}_{\pm 0.0}$ 
        & $99.82_{\pm 0.0}$
        & $77.47_{\pm 0.26}$
        & $11.16_{\pm 0.54}$
        & $\textbf{0.29}_{\pm 0.01}$
        & $\textbf{0.044}_{\pm 0.010}$
        & $7.31_{\pm 0.02}$\\ 
%%%%%%%%%%%%%%%%%%%%%%%%%%%%%%%%%%%%%%%%%%%
        \toprule[1pt]
        \midrule
        \multicolumn{8}{c}{\textbf{CIFAR-100 - 20-class forgetting}} \\
        \midrule
        Methods & UA & RA & TA & MIA & JSD & \textbf{IDI} & RTE (min)\\
        \midrule
        Original
        & $0.0$
        & $99.97$
        & $78.03$
        & $95.04$
        & $3.15$
        & $1.000$
        & $175.08$
        \\ 
        Retrain
        & $100.0$
        & $99.98$
        & $80.01$
        & $7.55$
        & $0.0$
        & $0.0$
        & $139.93$
        \\ \midrule
        FT
        & $\underline{99.81}_{\pm 0.04}$
        & $\underline{99.97}_{\pm 0.00}$
        & $\textbf{79.11}_{\pm 0.35}$
        & $0.22_{\pm 0.05}$
        & $0.37_{\pm 0.01}$
        & $\underline{0.474}_{\pm 0.007}$
        & $7.43_{\pm 0.07}$\\
        RL
        & $95.77_{\pm 0.09}$
        & $\textbf{99.98}_{\pm 0.01}$
        & $78.42_{\pm 0.05}$
        & \textbf{$0.0_{\pm 0.0}$}
        & $0.63_{\pm 0.01}$
        & $1.207_{\pm 0.004}$
        & $2.94_{\pm 0.01}$\\
        GA
        & $67.06_{\pm 2.58}$
        & $96.65_{\pm 0.47}$
        & $70.80_{\pm 0.65}$
        & $30.16_{\pm 1.42}$
        & $1.46_{\pm 0.11}$
        & $1.027_{\pm 0.006}$
        & $4.11_{\pm 0.02}$\\ 
        Bad-T
        & $95.54_{\pm 0.61}$
        & $\textbf{99.98}_{\pm 0.01}$
        & $69.71_{\pm 0.32}$
        & $32.07_{\pm 35.23}$
        & $2.83_{\pm 0.26}$
        & $1.211_{\pm 0.011}$
        & $5.17_{\pm 0.11}$\\ 
        EU-5
        & $\textbf{100.0}_{\pm 0.0}$
        & $99.82_{\pm 0.02}$
        & $76.89_{\pm 0.03}$
        & $14.50_{\pm 0.54}$
        & $0.52_{\pm 0.01}$
        & $0.807_{\pm 0.003}$
        & $\underline{1.83}_{\pm 0.04}$\\ 
        CF-5
        & $\textbf{100.0}_{\pm 0.0}$
        & $99.96_{\pm 0.02}$
        & $\underline{78.82}_{\pm 0.06}$
        & $2.68_{\pm 0.21}$
        & $\underline{0.33}_{\pm 0.01}$
        & $1.060_{\pm 0.008}$
        & $\textbf{1.80}_{\pm 0.03}$\\ 
        EU-10
        & $\textbf{100.0}_{\pm 0.0}$ 
        & $93.25_{\pm 0.32}$
        & $74.79_{\pm 0.39}$
        & $25.63_{\pm 0.38}$
        & $0.47_{\pm 0.01}$
        & $0.617_{\pm 0.005}$
        & $3.61_{\pm 0.51}$\\ 
        CF-10
        & $\textbf{100.0}_{\pm 0.0}$
        & $99.91_{\pm 0.01}$
        & $78.39_{\pm 0.24}$
        & $13.57_{\pm 0.32}$
        & $0.39_{\pm 0.01}$
        & $0.889_{\pm 0.005}$
        & $3.68_{\pm 0.16}$\\ 
        SCRUB
        & $95.03_{\pm 0.75}$
        & $99.90_{\pm 0.00}$
        & $77.61_{\pm 0.07}$
        & $0.93_{\pm 0.13}$
        & $0.38_{\pm 0.01}$
        & $0.997_{\pm 0.007}$
        & $2.14_{\pm 0.02}$\\ 
        SALUN
        & $90.69_{\pm 0.76}$
        & $98.97_{\pm 0.14}$
        & $74.72_{\pm 0.54}$
        & $0.17_{\pm 0.03}$
        & $0.60_{\pm 0.01}$
        & $1.113_{\pm 0.008}$
        & $3.85_{\pm 0.0}$\\ 
        \(\mathbf{\ell_1}\)-sparse
        & $83.49_{\pm 0.46}$
        & $99.52_{\pm 0.03}$
        & $76.79_{\pm 0.20}$
        & $\textbf{6.36}_{\pm 0.59}$
        & $0.38_{\pm 0.01}$
        & $1.035_{\pm 0.007}$
        & $3.08_{\pm 0.07}$\\ 
        \midrule
        \textbf{COLA}
        & $\textbf{100.0}_{\pm 0.0}$
        & $99.92_{\pm 0.0}$
        & $78.59_{\pm 0.32}$
        & $\underline{11.52}_{\pm 0.39}$
        & $\textbf{0.24}_{\pm 0.01}$
        & $\textbf{0.007}_{\pm 0.010}$
        & $6.97_{\pm 0.01}$\\ 
        \midrule
        \bottomrule[1pt]
    \end{tabular}
}
\caption{Multi-class forgetting on CIFAR-10 and CIFAR-100 datasets on ResNet-18 model. A better performance of an MU method corresponds to a smaller performance gap with Retrain (except RTE), with the top method in \textbf{bold} and the second best \underline{underlined}.}
\label{tab:multi_class_unlearning}
\end{table*}
\newpage
%%%%%%%%%%%%%%%%%%%%%%%%%%%%%%%%%%%%%%%%%%%
\begin{table*}[t!]
    
    \centering
    \resizebox{\textwidth}{!}{%
    \begin{tabular}{lcccccccccccc}
        \toprule[1pt]
        \midrule
        \multicolumn{8}{c}{\textbf{ImageNet-1K - ResNet-50}} \\
        \midrule
        Methods & UA & RA & TA & MIA & JSD & \textbf{IDI} & RTE (min)\\
        \midrule
        Original
        & $11.72$
        & $87.45$
        & $76.11$
        & $61.69$
        & $3.73$
        & $1.000$
        & $2680.15$
        \\ 
        Retrain
        & $100.0$
        & $88.80$
        & $75.88$
        & $9.41$
        & $0.0$
        & $0.0$
        & $2661.90$
        \\ \midrule
        FT
        & $\textbf{100.0}_{\pm 0.0}$
        & $\textbf{88.52}_{\pm 0.0}$
        & $\underline{76.16}_{\pm 0.01}$
        & $8.24_{\pm 1.23}$
        & $\underline{0.24}_{\pm 0.01}$
        & $0.102_{\pm 0.026}$
        & $140.04_{\pm 1.42}$\\
        RL
        & $\underline{99.96}_{\pm 0.03}$
        & $86.46_{\pm 0.07}$
        & $75.23_{\pm 0.01}$
        & $0.23_{\pm 0.01}$
        & $1.57_{\pm 0.03}$
        & $1.002_{\pm 0.007}$
        & $200.73_{\pm 1.87}$\\
        GA
        & $\textbf{100.0}_{\pm 0.0}$
        & $80.77_{\pm 0.22}$
        & $71.49_{\pm 0.10}$
        & $4.20_{\pm 0.46}$
        & $0.42_{\pm 0.03}$
        & $0.328_{\pm 0.023}$
        & $212.14_{\pm 2.61}$\\ 
        Bad-T
        & $98.01_{\pm 0.02}$
        & $84.03_{\pm 0.03}$
        & $73.42_{\pm 0.03}$
        & $69.13_{\pm 12.57}$
        & $3.51_{\pm 0.41}$
        & $1.152_{\pm 0.072}$
        & $211.52_{\pm 0.96}$\\ 

        EU-5
        & $\textbf{100.0}_{\pm 0.0}$
        & $79.62_{\pm 0.0}$
        & $71.22_{\pm 0.13}$
        & $13.33_{\pm 1.53}$
        & $0.26_{\pm 0.01}$
        & $0.183_{\pm 0.028}$
        & $193.38_{\pm 0.78}$\\ 
        CF-5
        & $\textbf{100.0}_{\pm 0.0}$
        & $84.31_{\pm 0.08}$
        & $74.16_{\pm 0.06}$
        & $10.21_{\pm 5.33}$
        & $\textbf{0.23}_{\pm 0.01}$
        & $0.701_{\pm 0.014}$
        & ${81.53}_{\pm 0.56}$\\ 
        EU-10
        & $\textbf{100.0}_{\pm 0.0}$
        & $71.84_{\pm 0.03}$
        & $65.78_{\pm 0.02}$
        & $16.65_{\pm 1.91}$
        & $0.35_{\pm 0.04}$
        & $\underline{-0.051}_{\pm 0.021}$
        & $193.79_{\pm 0.47}$\\ 
        CF-10
        & $\textbf{100.0}_{\pm 0.0}$
        & $80.87_{\pm 0.04}$
        & $72.34_{\pm 0.08}$
        & $13.99_{\pm 5.41}$
        & $0.25_{\pm 0.01}$
        & $0.608_{\pm 0.012}$
        & ${82.29}_{\pm 0.34}$\\ 
        SCRUB
        & $99.28_{\pm 0.07}$
        & $\underline{88.39}_{\pm 0.04}$
        & $76.51_{\pm 0.03}$
        & $7.42_{\pm 0.51}$
        & $0.25_{\pm 0.01}$
        & $0.517_{\pm 0.011}$
        & $426.04_{\pm 2.98}$\\ 
        SALUN
        & $89.67_{\pm 0.27}$
        & $86.25_{\pm 0.15}$
        & $75.54_{\pm 0.10}$
        & $0.50_{\pm 0.09}$
        & $0.88_{\pm 0.01}$
        & $0.343_{\pm 0.017}$
        & $793.82_{\pm 3.32}$\\
        \(\mathbf{\ell_1}\)-sparse
        & $97.57_{\pm 0.61}$
        & $85.33_{\pm 0.07}$
        & $74.77_{\pm 0.03}$
        & $\underline{8.84}_{\pm 1.39}$
        & $0.32_{\pm 0.02}$
        & $0.239_{\pm 0.031}$
        & $226.74_{\pm 1.35}$\\ 
        \midrule
        \textbf{COLA}
        & $\textbf{100.0}_{\pm 0.0}$
        & $87.93_{\pm 0.05}$
        & $\textbf{76.15}_{\pm 0.04}$
        & $\textbf{9.95}_{\pm 1.21}$
        & $\underline{0.24}_{\pm 0.01}$
        & $\textbf{0.040}_{\pm 0.042}$
        & $171.44_{\pm 0.75}$\\ 
%%%%%%%%%%%%%%%%%%%%%%%%%%%%%%%%%%%%%%%%%%%
        \toprule[1pt]
        \midrule
        \multicolumn{8}{c}{\textbf{ImageNet-1K - ViT}} \\
        \midrule
        Methods & UA & RA & TA & MIA & JSD & \textbf{IDI} & RTE (min)\\
        \midrule
        Original
        & $2.48$
        & $98.18$
        & $80.59$
        & $71.00$
        & $4.45$
        & $1.000$
        & $1943.69^\star$
        \\ 
        Retrain
        & $100.0$
        & $98.33$
        & $80.42$
        & $8.09$
        & $0.0$
        & $0.0$ 
        & $1920.77^\star$
        \\ \midrule
        FT
        & $96.39_{\pm 0.01}$
        & $98.85_{\pm 0.03}$
        & $80.93_{\pm 0.06}$
        & $3.88_{\pm 0.33}$
        & $0.65_{\pm 0.02}$
        & $0.937_{\pm 0.009}$
        & $281.73_{\pm 2.30}$\\
        RL
        & $98.33_{\pm 0.02}$
        & $98.99_{\pm 0.07}$
        & $81.65_{\pm 0.07}$
        & $0.0_{\pm 0.0}$
        & $2.13_{\pm 0.15}$
        & $1.152_{\pm 0.033}$
        & $150.32_{\pm 4.31}$\\
        GA
        & $\textbf{100.0}_{\pm 0.0}$
        & $97.04_{\pm 0.01}$
        & $\underline{80.17}_{\pm 0.04}$
        & $8.26_{\pm 2.14}$
        & $0.52_{\pm 0.23}$
        & $0.674_{\pm 0.021}$
        & $193.73_{\pm 2.23}$\\
        Bad-T
        & $98.21_{\pm 0.03}$
        & $\underline{97.85}_{\pm 0.07}$
        & $\textbf{80.58}_{\pm 0.03}$
        & $0.0_{\pm 0.0}$
        & $2.62_{\pm 0.06}$
        & $1.312_{\pm 0.015}$
        & $721.15_{\pm 5.23}$\\ 
        EU-5
        & $\textbf{100.0}_{\pm 0.0}$
        & $93.82_{\pm 0.02}$
        & $80.00_{\pm 0.01}$
        & $4.74_{\pm 1.33}$
        & $0.63_{\pm 0.02}$
        & $0.519_{\pm 0.008}$
        & $300.55_{\pm 0.76}$\\ 
        CF-5
        & $98.75_{\pm 0.0}$
        & $96.57_{\pm 0.01}$
        & $80.09_{\pm 0.04}$
        & $4.49_{\pm 0.34}$
        & $0.64_{\pm 0.01}$
        & $0.731_{\pm 0.024}$
        & $\textbf{122.39}_{\pm 0.53}$\\ 
        EU-10
        & $\textbf{100.0}_{\pm 0.0}$
        & $87.33_{\pm 0.10}$
        & $76.26_{\pm 0.13}$
        & $\textbf{8.09}_{\pm 0.20}$
        & $\textbf{0.36}_{\pm 0.02}$
        & $-2.662_{\pm 0.231}$
        & $345.37_{\pm 0.70}$\\ 
        CF-10
        & $\underline{99.95}_{\pm 0.01}$
        & $93.86_{\pm 0.02}$
        & $78.69_{\pm 0.01}$
        & $7.68_{\pm 1.11}$
        & $0.72_{\pm 0.03}$
        & $\underline{0.009}_{\pm 0.021}$
        & $\underline{140.11}_{\pm 0.49}$\\ 
        SCRUB
        & $\textbf{100.0}_{\pm 0.00}$
        & $98.84_{\pm 0.02}$
        & $81.62_{\pm 0.01}$
        & $3.19_{\pm 0.91}$
        & $1.062_{\pm 0.03}$
        & $-0.846_{\pm 0.032}$
        & $404.02_{\pm 2.96}$\\ 
        SALUN
        & $94.64_{\pm 0.76}$
        & $\textbf{98.13}_{\pm 0.21}$
        & $80.74_{\pm 0.05}$
        & $0.13_{\pm 0.01}$
        & $1.83_{\pm 0.09}$
        & $0.980_{\pm 0.065}$
        & $321.13_{\pm 2.75}$\\ 
        \(\mathbf{\ell_1}\)-sparse
        & $93.55_{\pm 0.62}$
        & $94.69_{\pm 0.37}$
        & $78.84_{\pm 0.10}$
        & $2.98_{\pm 0.33}$
        & $\underline{0.49}_{\pm 0.01}$
        & $0.831_{\pm 0.022}$
        & $717.42_{\pm 3.21}$\\
        \midrule
        \textbf{COLA}
        & $\textbf{100.0}_{\pm 0.0}$
        & $96.42_{\pm 0.03}$
        & $79.28_{\pm 0.21}$
        & $\underline{8.02}_{\pm 1.36}$
        & $0.59_{\pm 0.02}$
        & $\textbf{0.006}_{\pm 0.007}$
        & $501.12_{\pm 2.17}$\\
        \midrule
        \bottomrule[1pt]

\end{tabular}
    }
\caption{5-class forgetting results on ImageNet-1K dataset across different model architectures. A better performance of an MU method corresponds to a smaller performance gap with Retrain (except RTE), with the top method in \textbf{bold} and the second best \underline{underlined}. The \(\star\) symbol indicated in RTE of Original and Retrain means that models are pretrained on ImageNet-21K and then finetuned on ImageNet-1K, with the reported time reflecting only the finetuning process. In contrast, Original and Retrain without \(\star\) are trained from scratch on ImageNet-1K.
} 
\label{tab:5_class_unlearning_imagenet}
\end{table*}
\newpage
%%%%%%%%%%%%%%%%%%%%%%%%%%%%%%%%%%%%%%%%%%%
\begin{table*}[t!]
    
    \centering
    \resizebox{\textwidth}{!}{%
    \begin{tabular}{lcccccccccccc}
        \toprule[1pt]
        \midrule
        \multicolumn{8}{c}{\textbf{CIFAR-10 - ResNet-18}} \\
        \midrule
        Methods & UA & RA & TA & MIA & JSD & \textbf{IDI} & RTE (min)\\
        \midrule
        Original
        & $0.0$
        & $100.0$
        & $95.54$
        & $92.90$
        & $0.09$
        & $1.000$
        & $170.32$
        \\ 
        Retrain
        & $3.94$
        & $100.0$
        & $95.26$
        & $75.12$
        & $0.0$
        & $0.0$
        & $152.87$
        \\ \midrule
        FT
        & $5.03_{\pm 0.40}$
        & $98.95_{\pm 0.21}$
        & $92.94_{\pm 0.26}$
        & $83.52_{\pm 0.58}$
        & $0.07_{\pm 0.11}$
        & $\underline{-0.069}_{\pm 0.013}$
        & $8.11_{\pm 0.03}$\\
        RL
        & $4.77_{\pm 0.27}$
        & $\textbf{99.92}_{\pm 0.0}$
        & $\textbf{93.54}_{\pm 0.04}$
        & $22.47_{\pm 1.19}$
        & $0.38_{\pm 0.02}$
        & $0.084_{\pm 0.030}$
        & $2.75_{\pm 0.01}$\\
        GA
        & $2.86_{\pm 0.76}$
        & $98.37_{\pm 0.71}$
        & $91.90_{\pm 0.70}$
        & $85.49_{\pm 2.17}$
        & $0.09_{\pm 0.01}$
        & $0.924_{\pm 0.028}$
        & $4.31_{\pm 0.03}$\\
        Bad-T
        & $5.47_{\pm 1.05}$
        & $\underline{99.87}_{\pm 0.05}$
        & $91.51_{\pm 0.61}$
        & $39.53_{\pm 3.43}$
        & $0.27_{\pm 0.03}$
        & $0.939_{\pm 0.053}$
        & $4.78_{\pm 0.09}$\\
        EU-10
        & $3.16_{\pm 0.19}$
        & $98.68_{\pm 0.08}$
        & $93.07_{\pm 0.12}$
        & $\underline{83.40}_{\pm 0.21}$
        & $\underline{0.06}_{\pm 0.01}$
        & $-0.110_{\pm 0.013}$
        & $\underline{2.13}_{\pm 0.05}$\\ 
        CF-10
        & $2.71_{\pm 0.24}$
        & $99.11_{\pm 0.06}$
        & $\underline{93.47}_{\pm 0.15}$
        & $84.33_{\pm 0.05}$
        & $\textbf{0.05}_{\pm 0.01}$
        & $0.219_{\pm 0.029}$
        & $\textbf{2.10}_{\pm 0.06}$\\ 
        SCRUB
        & $\underline{4.31}_{\pm 1.50}$
        & $96.21_{\pm 1.70}$
        & $88.83_{\pm 1.86}$
        & $37.88_{\pm 7.65}$
        & $0.56_{\pm 0.09}$
        & $0.322_{\pm 0.016}$
        & $3.37_{\pm 0.05}$\\ 
        SALUN
        & $2.74_{\pm 0.30}$
        & $97.77_{\pm 0.04}$
        & $91.68_{\pm 0.44}$
        & $83.52_{\pm 2.20}$
        & $0.10_{\pm 0.03}$
        & $0.861_{\pm 0.012}$
        & $5.69_{\pm 0.04}$\\
        \(\mathbf{\ell_1}\)-sparse
        & $5.47_{\pm 0.22}$
        & $96.66_{\pm 0.07}$
        & $91.31_{\pm 0.25}$
        & $\textbf{77.12}_{\pm 0.21}$
        & $0.09_{\pm 0.01}$
        & $-0.157_{\pm 0.026}$
        & $3.03_{\pm 0.04}$\\ 
        \midrule
        % \textbf{COLA}
        % & $1.22_{\pm 0.07}$
        % & $\textbf{99.99}_{\pm 0.01}$
        % & $94.71_{\pm 0.05}$
        % & $88.95_{\pm 0.31}$
        % & $0.07_{\pm 0.01}$
        % & $0.849_{\pm 0.076}$
        % & $6.95_{\pm 0.03}$\\ 
        \textbf{COLA+}
        & $\textbf{3.90}_{\pm 0.08}$
        & $99.24_{\pm 0.17}$
        & $93.23_{\pm 0.09}$
        & $83.48_{\pm 0.10}$
        & $\underline{0.06}_{\pm 0.01}$
        & $\textbf{0.024}_{\pm 0.010}$
        & $7.80_{\pm 0.02}$\\ 
%%%%%%%%%%%%%%%%%%%%%%%%%%%%%%%%%%%%%%%%%%%
        \toprule[1pt]
        \midrule
        \multicolumn{8}{c}{\textbf{CIFAR-100 - ResNet-18}} \\
        \midrule
        Methods & UA & RA & TA & MIA & JSD & \textbf{IDI} & RTE (min)\\
        \midrule
        Original
        & $0.0$
        & $99.98$
        & $78.09$
        & $95.82$
        & $0.56$
        & $1.000$
        & $175.08$
        \\ 
        Retrain
        & $23.10$
        & $99.98$
        & $77.78$
        & $39.72$
        & $0.0$
        & $0.0$
        & $170.31$
        \\ \midrule
        FT
        & $17.44_{\pm 1.12}$
        & $98.46_{\pm 0.24}$
        & $70.99_{\pm 0.45}$
        & $67.35_{\pm 0.53}$
        & $0.46_{\pm 0.02}$
        & $0.311_{\pm 0.034}$
        & $8.40_{\pm 0.13}$\\
        RL
        & $24.67_{\pm 0.42}$
        & $\underline{99.66}_{\pm 0.0}$
        & $73.10_{\pm 0.49}$
        & $2.13_{\pm 0.17}$
        & $0.84_{\pm 0.02}$
        & $-0.246_{\pm 0.056}$
        & $2.95_{\pm 0.03}$\\
        GA
        & $11.73_{\pm 1.43}$
        & $95.21_{\pm 0.78}$
        & $68.38_{\pm 1.03}$
        & $74.97_{\pm 1.10}$
        & $0.65_{\pm 0.01}$
        & $0.704_{\pm 0.039}$
        & $4.66_{\pm 0.03}$\\
        Bad-T
        & $64.35_{\pm 7.44}$
        & $99.07_{\pm 0.56}$
        & $53.05_{\pm 2.53}$
        & $11.85_{\pm 5.93}$
        & $1.51_{\pm 0.16}$
        & $1.003_{\pm 0.006}$
        & $5.04_{\pm 0.05}$\\
        EU-10
        & $24.15_{\pm 0.09}$
        & $90.15_{\pm 0.08}$
        & $72.25_{\pm 0.36}$
        & $\textbf{59.47}_{\pm 0.39}$
        & $0.27_{\pm 0.01}$
        & $0.404_{\pm 0.085}$
        & $\underline{2.31}_{\pm 0.02}$\\ 
        CF-10
        & $20.40_{\pm 0.20}$
        & $95.06_{\pm 0.24}$
        & $\textbf{74.44}_{\pm 0.23}$
        & $62.18_{\pm 0.27}$
        & $\underline{0.25}_{\pm 0.01}$
        & $0.464_{\pm 0.061}$
        & $\textbf{2.30}_{\pm 0.02}$\\ 
        SCRUB
        & $3.47_{\pm 2.85}$
        & $97.77_{\pm 2.31}$
        & $71.89_{\pm 2.87}$
        & $71.49_{\pm 4.15}$
        & $0.37_{\pm 0.02}$
        & $0.528_{\pm 0.013}$
        & $3.59_{\pm 0.05}$\\ 
        SALUN
        & $32.77_{\pm 1.20}$
        & $\textbf{99.87}_{\pm 0.02}$
        & $71.97_{\pm 0.37}$
        & $3.32_{\pm 0.28}$
        & $0.81_{\pm 0.02}$
        & $\underline{-0.226}_{\pm 0.078}$
        & $5.99_{\pm 0.09}$\\ 
        \(\mathbf{\ell_1}\)-sparse
        & $\textbf{22.83}_{\pm 0.15}$
        & $88.94_{\pm 0.41}$
        & $69.54_{\pm 0.73}$
        & $62.36_{\pm 0.37}$
        & $0.26_{\pm 0.01}$
        & $0.634_{\pm 0.072}$
        & $3.37_{\pm 0.03}$\\ 
        \midrule
        % \textbf{COLA}
        % & $15.44_{\pm 0.23}$
        % & $96.29_{\pm 0.01}$
        % & $\textbf{75.61}_{\pm 0.11}$
        % & $64.62_{\pm 0.42}$
        % & $0.18_{\pm 0.01}$
        % & $0.724_{\pm 0.072}$
        % & $6.95_{\pm 0.03}$\\ 
        \textbf{COLA+}
        & $\underline{23.50}_{\pm 0.16}$
        & $93.78_{\pm 0.07}$
        & $\underline{73.15}_{\pm 0.59}$
        & $\underline{59.58}_{\pm 0.24}$
        & $\textbf{0.24}_{\pm 0.01}$
        & $\textbf{0.078}_{\pm 0.013}$
        & $10.2_{\pm 0.16}$\\
        
        \midrule
        \bottomrule[1pt]

\end{tabular}
    }
\caption{Random data forgetting on CIFAR-10 and CIFAR-100 datasets on ResNet-18 model. A better performance of an MU method corresponds to a smaller performance gap with Retrain (except RTE), with the top method in \textbf{bold} and the second best \underline{underlined}.}
\label{tab:random_sample_unlearning}
\end{table*}
\clearpage
\begin{table*}[h]
\centering

\begin{tabular}{lcccccc}
\toprule
Method     &  Block 1     & Block 2     & Block 3     & Block 4     & Block 5     & IDI   \\
\midrule
Original   & 0.002 & 0.002 & 0.003 & 0.006 & 0.006 & 0.005 \\
Retrain    & 0.001 & 0.003 & 0.003 & 0.006 & 0.007 & 0.007 \\
FT         & 0.001 & 0.002 & 0.004 & 0.010 & 0.008 & 0.007 \\
RL         & 0.002 & 0.004 & 0.005 & 0.003 & 0.005 & 0.004 \\
GA         & 0.001 & 0.001 & 0.004 & 0.006 & 0.011 & 0.013 \\
l1-sparse  & 0.001 & 0.000 & 0.002 & 0.007 & 0.007 & 0.011 \\
SCRUB      & 0.001 & 0.005 & 0.003 & 0.004 & 0.005 & 0.007 \\
SALUN      & 0.002 & 0.001 & 0.003 & 0.005 & 0.012 & 0.011 \\
\bottomrule
\end{tabular}
\caption{Standard Deviation of \cref{fig:mi__resnet18_resnet50_cifar10_class} - (CIFAR-10, ResNet-18)}
\label{tab:resnet18_cifar10_std}
\end{table*}

\begin{table*}[h]
\centering

\begin{tabular}{lccccccc}
\toprule
Method     & Block 1     & Block 2     & Block 3     & Block 4     & Block 5     & Block 6     & IDI   \\
\midrule
Original   & 0.003 & 0.002 & 0.009 & 0.007 & 0.013 & 0.008 & 0.011 \\
Retrain    & 0.001 & 0.003 & 0.001 & 0.008 & 0.005 & 0.007 & 0.009 \\
FT         & 0.003 & 0.005 & 0.008 & 0.015 & 0.011 & 0.012 & 0.019 \\
RL         & 0.005 & 0.005 & 0.007 & 0.003 & 0.008 & 0.004 & 0.009 \\
GA         & 0.002 & 0.001 & 0.003 & 0.011 & 0.010 & 0.013 & 0.018 \\
l1-sparse  & 0.002 & 0.004 & 0.002 & 0.004 & 0.021 & 0.015 & 0.023 \\
SCRUB      & 0.000 & 0.004 & 0.004 & 0.023 & 0.028 & 0.031 & 0.060 \\
SALUN      & 0.001 & 0.000 & 0.006 & 0.005 & 0.020 & 0.011 & 0.019 \\
\bottomrule
\end{tabular}
\caption{Standard Deviation of \cref{fig:mi__resnet18_resnet50_cifar10_class} - (CIFAR-10, ResNet-50)}
\label{tab:resnet50_cifar10_std}
\end{table*}

\begin{table*}[h]
\centering

\begin{tabular}{lccc}
\toprule
Method & Ratio 1:5   & Ratio 1:20  & Ratio 1:99  \\
\midrule
FT     & 0.013 & 0.008 & 0.019 \\
RL     & 0.016 & 0.007 & 0.009 \\
GA     & 0.020 & 0.008 & 0.018 \\
CF-10  & 0.007 & 0.016 & 0.035 \\
SALUN  & 0.023 & 0.025 & 0.019 \\
SCRUB  & 0.006 & 0.013 & 0.023 \\
\bottomrule
\end{tabular}
\caption{Standard Deviation of \cref{fig:exp__ratio}}
\label{tab:ratio_std}
\end{table*}
\newpage

\newpage

\newpage

\newpage

\newpage

%%%%%%%%%%%%%%%%%%%%%%%%%%%%%%%%%%%%%%%%%%%%%%%%%%%%%%%%%%%%%%%%%%%%%%%%%%%%%%%
%%%%%%%%%%%%%%%%%%%%%%%%%%%%%%%%%%%%%%%%%%%%%%%%%%%%%%%%%%%%%%%%%%%%%%%%%%%%%%%
\clearpage
\end{document}